\documentclass[11pt]{article}

\usepackage[preprint]{acl}

\usepackage{times}
\usepackage{latexsym}

\usepackage[T1]{fontenc}
\usepackage[utf8]{inputenc}

\usepackage{microtype}

\usepackage{inconsolata}

\usepackage{graphicx}

\usepackage{microtype}
\usepackage{graphicx}
\usepackage{subcaption}
\usepackage{booktabs} % for professional tables

\usepackage{hyperref}

\usepackage{amsmath,amsfonts,bm}

\def\eqref#1{equation~\ref{#1}}
\def\1{\bm{1}}

\DeclareMathAlphabet{\mathsfit}{\encodingdefault}{\sfdefault}{m}{sl}
\SetMathAlphabet{\mathsfit}{bold}{\encodingdefault}{\sfdefault}{bx}{n}

\usepackage[utf8]{inputenc} % allow utf-8 input
\usepackage[T1]{fontenc}    % use 8-bit T1 fonts
\usepackage{url}            % simple URL typesetting
\usepackage{booktabs}       % professional-quality tables
\usepackage{amsfonts}       % blackboard math symbols
\usepackage{nicefrac}       % compact symbols for 1/2, etc.
\usepackage{microtype}      % microtypography
\usepackage{xcolor}         % colors
\usepackage{booktabs}       % table
\usepackage{graphicx}       % also table
\usepackage[most]{tcolorbox}
\tcbuselibrary{listings, breakable}
\usepackage{booktabs}
\usepackage{multirow}
\usepackage{colortbl}
\usepackage{xcolor}
\usepackage{booktabs}
\usepackage{multirow}
\usepackage{colortbl}
\usepackage{booktabs}       % 专业表格线
\usepackage{multirow}       % 多行单元格支持
\usepackage[table,xcdraw]{xcolor}  % 同时支持表格背景和颜色宏
\usepackage{amsmath}
\usepackage{amssymb}
\usepackage{mathtools}
\usepackage{amsthm}
\usepackage{algorithm}
\usepackage{algpseudocode}
\usepackage{enumitem}
\usepackage{multirow}
\usepackage{multicol}
\usepackage{graphicx}
\usepackage{tabularx}
\usepackage{makecell}
\theoremstyle{plain}

\usepackage{amsmath}
\usepackage{amssymb}
\usepackage{mathtools}
\usepackage{amsthm}

\usepackage[capitalize,noabbrev]{cleveref}

\theoremstyle{plain}

\theoremstyle{definition}

\theoremstyle{remark}

\usepackage[textsize=tiny]{todonotes}

\title{Aplaud: Adaptive Personalized Low-Rank Decomposition for User-Specific LLM}

\author{
  Xinyu Li\textsuperscript{1},
  Ruoming Jin\textsuperscript{1},
  Jianfeng Zhu\textsuperscript{1},
  Ruixin Guo\textsuperscript{1},
  Zhi Liu\textsuperscript{2} \\
  \textsuperscript{1}Department of Computer Science, Kent State University \\
  \textsuperscript{2}iLambda Inc. \\
  \texttt{\{xli74,rjin1,jzhu10,rguo5\}@kent.edu} \\
  \texttt{zliu@ilambda.ai}
}

\begin{document}
\maketitle
\begin{abstract}
In this paper, we study the problem of \textit{personalized survey response prediction} using fine-tuned large language models (LLMs). This task poses unique challenges: limited per-user training data, scalability of model storage, and the need to exploit shared structure across survey questions. To address these issues, we propose \textbf{Aplaud} (Adaptive Personalized Low-rank and User-specific Nested Decomposition), a lightweight and scalable framework for LLM personalization. Aplaud extends the LoRA paradigm by separating adaptation into a frozen, shared low-rank basis and a compact user-specific correction, augmented with a rank-one residual for finer personalization. To further reduce per-user parameter cost and mitigate overfitting, the correction matrix can be factorized into an even lower-rank form. Empirical results demonstrate that Aplaud achieves efficient, scalable personalization across users while outperforming state-of-the-art LoRA-based personalized LLM approaches in both generalization and inference efficiency.
\end{abstract}

\section{Introduction}
Surveys and polls such as the Pew Research Survey~\cite{pewresearch2024}, the General Social Survey (GSS)~\cite{gss2022}, the Gallup World Poll~\cite{gallup}, and the American National Election Studies (ANES)~\cite{anes} have long been central to public policy, social science, and marketing research~\cite{graham2023polling,waldner2018unwelcome,malhotra2019marketing,churchill2010marketing,dillman2014tailored}. Beyond these canonical instruments, surveys and interviews remain foundational methods across domains including product design, political science, biomedicine, psychology, and education. However, traditional survey methodologies are increasingly strained by rising costs, declining response rates, and persistent concerns about accuracy and representativeness~\cite{keeter2017low,clinton2021taskforce,kennedy2018evaluation}.

Driven by growing demand in the multi-billion-dollar market research sector, researchers and practitioners have begun exploring \emph{synthetic participants}—LLM-generated respondents—as a scalable alternative to human data collection~\citep{argyle2023out,jiang2023social,aher_using_2023,horton_large_2023,demszky2023using,hamalainen2023evaluating,ravi2023large,louie2024roleplay,prpa2024challenges,hu2024psycollm,kim_ai-augmented_2024,bisbee_synthetic_2024,sun2024random,zhang2024simulating,suh2025languagemodelfinetuningscaled,hao2025multi,anthis2025llmsocialsimulationspromising}. Industry adoption has accelerated rapidly: the Qualtrics 2025 marketing trend report~\cite{qualtrics2025market} explicitly positions synthetic responses as substitutes for human respondents, while organizations such as YouGov and Kantar and startups including SyntheticUsers~\cite{syntheticuser2023}, OpinioAI~\cite{opinioai2023}, Delve.ai~\cite{delveai2023}, and PersonaLive.ai~\cite{personalive2024} now offer synthetic survey responses at scale.

Most existing work, however, focuses on conditioning LLM outputs on coarse demographic attributes—a setting we refer to as \textbf{persona-level (subpopulation) prediction}~\citep{argyle2023out,aher_using_2023,horton_large_2023,sanders2023ai,lee2024can,anthis2025llmsocialsimulationspromising,santurkar2023whose,wang_not_2024,kapania2024simulacrum,gao_take_2024,giorgi2024modelinghumansubjectivityllms,wang2025largelanguagemodelsreplace}. Empirical studies consistently show that such approaches produce homogenized and biased responses, failing to capture individual-level variation and opinion diversity~\citep{santurkar2023whose,wang_not_2024,lee2024can,kapania2024simulacrum,gao_take_2024,giorgi2024modelinghumansubjectivityllms,wang2025largelanguagemodelsreplace,neumann2025usellmssimulateopinions}. Recent efforts improve subpopulation alignment via fine-tuning or RLHF~\citep{jang2023personalized,zhang2024personalization,suh2025languagemodelfinetuningscaled}, but remain fundamentally limited to demographic aggregation.

\noindent\textbf{The Personalized Survey Response Prediction Problem.}
To move beyond subpopulation-level modeling and enable personalization at the level of individual users, we introduce the \emph{personalized survey response prediction problem}: \emph{Can a fine-tuned (personalized) LLM replicate, predict, or simulate an individual’s responses to unseen survey questions, given their answers to a prior set of questions in an existing survey?}
Unlike demographic conditioning or persona-level simulation, this setting targets individualized behavioral and preference modeling, requiring the model to capture stable yet nuanced response patterns for each user.

This problem has immediate real-world relevance. Organizations in both public and private sectors routinely conduct large-scale surveys and retain the resulting data. When new or follow-up questions arise, it is often desirable to re-query the original participants to preserve continuity and comparability. In practice, however, re-contacting respondents can be costly or infeasible due to attrition, survey fatigue, and escalating incentive costs. Consequently, stakeholders are increasingly turning to LLMs to generate synthetic responses as a preliminary step before committing resources to additional data collection~\cite{qualtrics2025market}.

%The task is closely related to \emph{digital twins}, which aim to simulate individual behavior or preferences rather than aggregate subpopulation statistics~\citep{syntheticuser2023,personalive2024}. It also connects to recent advances in recommendation and personalization, where \emph{personalized LLMs}, or \emph{personalized alignment}, have become central research goals~\citep{zhang2024personalization,guan2025surveypersonalizedalignment}.

%Yet, to the best of our knowledge, personalized LLMs have not been systematically developed or evaluated for individual-level survey response prediction, leaving an important and largely unexplored gap that this work addresses.

\subsection{Research Challenges and Our Approach}
The personalized survey response prediction problem presents three core challenges.
First, the number of survey questions per user is typically limited, often ranging from tens to, at most, a few thousand, resulting in sparse personalized data. Na\"{\i}vely fine-tuning user-specific parameters under such constraints can easily lead to severe overfitting.
Second, real-world deployments may involve thousands to tens of millions of users. Even with parameter-efficient fine-tuning (PEFT) methods such as LoRA, maintaining a separate adapter for each user incurs prohibitive storage and deployment costs at scale, rendering per-user fine-tuning impractical.
Third, surveys typically ask the same set of questions across users, inducing shared semantics and correlations in responses. An effective personalization strategy should exploit this structure rather than treating users as independent learning problems.

\begin{figure}
    \centering
    \includegraphics[scale=0.25]{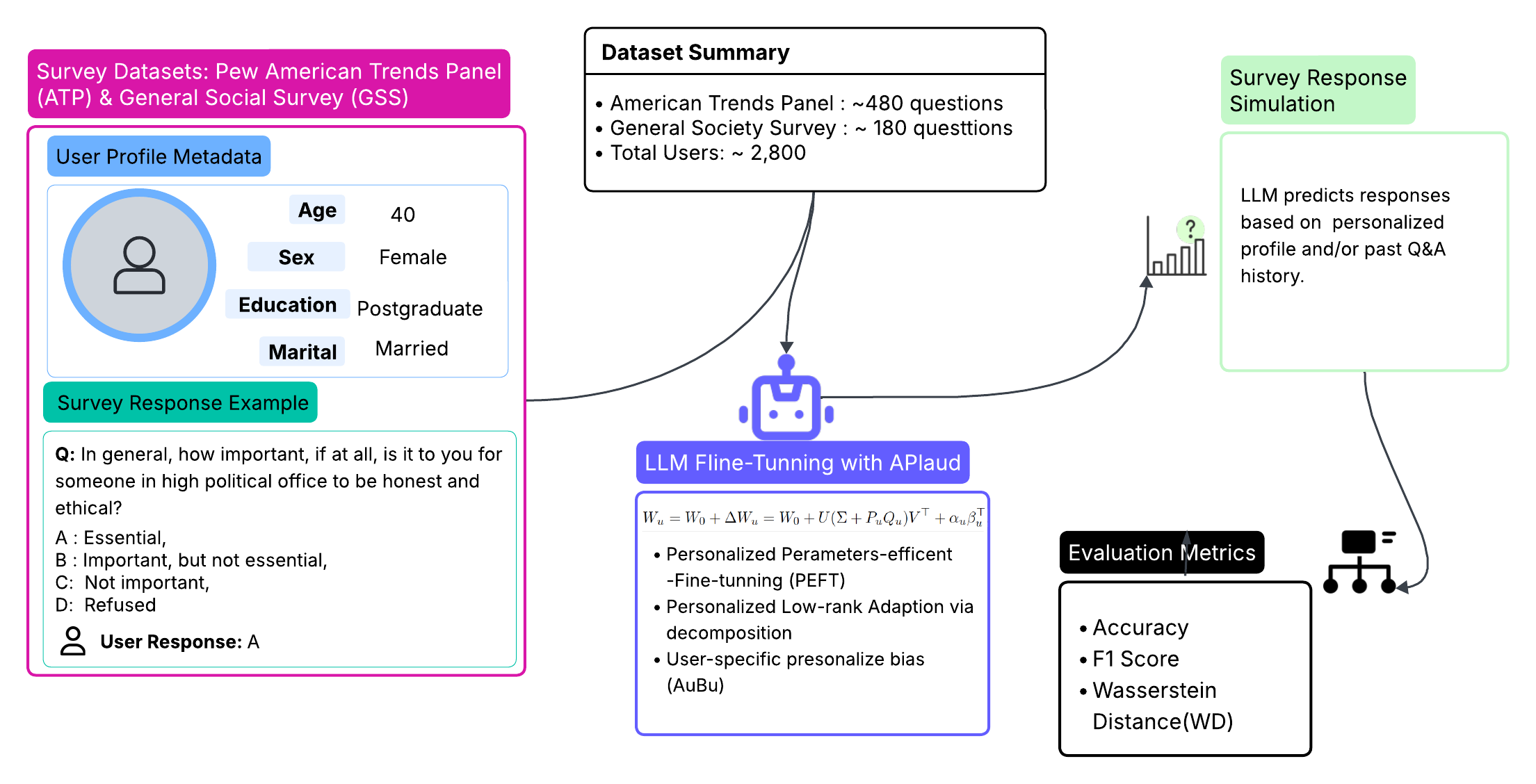}
    \caption{Aplaud Overview. By leveraging shared subspace, Aplaud is able to align individual preferences for structured survey data with limited parameter budget.}
    \label{fig:overview}
\end{figure}

To address these challenges, we propose \textbf{Aplaud}, a scalable and data-efficient framework for individualized survey response prediction that leverages shared structure across survey questions while maintaining compact personalization. An overview of the proposed framework is shown in Figure~\ref{fig:overview}.

The key idea of Aplaud is to begin from the shared low-rank space learned via standard LoRA training, represented by a low-rank update $AB$, and apply singular value decomposition (SVD) to obtain an orthogonal decomposition $U \Sigma V^\top$. We hypothesize that the resulting subspaces $U$ and $V$ capture stable, semantically meaningful directions that generalize across users. Personalization is introduced through a small user-specific matrix $C_u$ that modulates the shared singular values. Concretely, the adapted weight is given by
\[
    W = W_0 + U(\Sigma + C_u)V^\top ,
\]
where $U$ and $V$ are shared across users, while $C_u$ encodes individual-specific preferences.

To further enhance expressiveness without sacrificing efficiency, Aplaud augments this formulation with a lightweight user-specific low-rank residual term. Finally, to reduce parameter overhead and mitigate overfitting, the personalization matrix $C_u$ is factorized into a lower-rank form $P_u Q_u$.

This design directly addresses the three challenges above:
(1) compact user-specific parameters mitigate overfitting when per-user data is limited;
(2) low-rank factorization substantially reduces per-user storage and deployment costs; and
(3) shared subspaces maximize the reuse of semantic information common across users.

\noindent\textbf{Our Contributions.}
This work makes the following contributions:
\begin{itemize}
\item We introduce the \emph{personalized survey response prediction problem}, a novel and practically motivated task that bridges survey research and personalized LLM development, and propose it as a new benchmark for individual-level personalization.
\item We present \textbf{Aplaud}, a scalable and parameter-efficient framework that combines shared LoRA-derived subspaces with lightweight, user-specific corrections via SVD-based decomposition and residual adaptation.
\item Through extensive empirical evaluation, we show that Aplaud reduces per-user parameter cost by orders of magnitude compared to state-of-the-art personalized LLM methods (e.g., OPPU), while achieving comparable or superior predictive accuracy.
\end{itemize}
\section{Preliminary:  LoRA and Personalized LLM}
\label{sec:preliminary}

\subsection{LoRA and its Variants}
LoRA~\citep{hulora} is one of the most widely adopted parameter-efficient fine-tuning (PEFT) techniques. 
% It is motivated by the low intrinsic dimensionality hypothesis~\citep{angwin_machine_2016}, which suggests that fine-tuning can often be effectively performed in a lower-dimensional subspace. LoRA achieves this by introducing low-rank updates to the dense layers of a pre-trained neural network, instead of modifying the full parameter matrix.
Formally, let $W_0 \in \mathbb{R}^{d \times k}$ denote the original weight matrix of a dense layer. LoRA introduces a trainable update $\Delta W \in \mathbb{R}^{d \times k}$ such that the updated layer is parameterized by:
\[
W = W_0 + \Delta W.
\]
Rather than learning $\Delta W$ directly, LoRA factorizes it as the product of two low-rank matrices:
\[W = W_0 + s AB,
\]
where $A \in \mathbb{R}^{d \times r}$, $B \in \mathbb{R}^{r \times k}$, and $r \ll \min\{d, k\}$. This factorization significantly reduces the number of trainable parameters from $d \times k$ (in full fine-tuning) to $r \times (d + k)$. For simplicity of exposition, we assume in the remainder of this paper that $W_0$ is a square matrix (i.e., $d = k$), which is commonly the case in transformer-based architectures. Nonetheless, our method generalizes naturally to non-square matrices.

% We note that several recent approaches have explored the use of Singular Value Decomposition (SVD) to enhance LoRA training. For example, given the SVD of a pre-trained weight matrix \( W_0 = U \Sigma V^\top \), one can initialize the LoRA adapters using the factorized components—setting \( A = U \Sigma^{1/2} \) and \( B = \Sigma^{1/2} V^\top \)—to provide a data-informed starting point for fine-tuning.

Some methods adopt SVD-inspired parameterizations directly during training. For instance, AdaLoRA~\citep{zhang2023adaloraa} approximates the weight update matrix as \( \Delta W = P \Sigma Q^\top \), where \( P \) and \( Q \) are constrained to be approximately orthogonal via regularization terms \( \|P^\top P - I\| \) and \( \|Q^\top Q - I\| \), and \( \Sigma \) is a learnable diagonal matrix.
PiSSA (Principal Singular Values and Singular Vectors Adaptation)~\citep{meng2024pissa} initializes LoRA adapters using the top singular components of the pre-trained weights via SVD while freezing the remaining components. 
In contrast, MiLoRA~\citep{wang2024miloraharnessingminorsingular} and KASA~\citep{wang2024kasaknowledgeawaresingularvalueadaptation} propose to freeze the top singular components and instead fine-tune the minor singular directions, emphasizing complementary subspaces for adaptation.
To the best of our knowledge, our approach (Aplaud) is the \textit{first} to leverage SVD-based decomposition strategies to support personalized LLM adaptation and personalized alignment.

\subsection{Personalized LLM}
\noindent{\bf Prompt-based Personalization.}
User information—such as demographics, preferences, behavioral signals, and historical activity -- derived from user-generated content or contextual background, is typically encoded into prompts~\citep{xu2022long,aher_using_2023,argyle2023out,li2023text,bao2023tallrec,dong2023steerlm,lee2024aligning,yang2024rewards,li20251000000usersuserscaling}. When user history is extensive, techniques such as prompt refinement~\citep{li2024learning} and retrieval-augmented generation (RAG)~\citep{wang2024unims} can be employed to construct more informative and scalable prompts.

% In the context of personalized survey response prediction, the number of available questions and responses per user typically ranges from a few dozen to a few thousand--sufficient to fit within the context window of commercial LLMs such as ChatGPT. However, when using open-source models with more limited context capacity, it may be necessary to summarize prior interactions or apply RAG-based mechanisms to generate compact, user-specific prompts.

\noindent{\bf Encoding-based Personalization.} In this class of approaches, user data and preferences are compressed into vector representations or embeddings~\citep{ning2024user,li2024learning,shenfeld2025language}, which are then integrated into the model to modulate token-level processing and output generation for personalization. Similarly, user-specific latent variables and reward models have been developed to enable personalization through reinforcement learning~\citep{poddar2024personalizing,gong2025latent,chen2025pal}.

While these methods allow for individualized conditioning, they typically rely on a shared transformer backbone, resulting in a uniform inference architecture across all users. 
% This shared structure implicitly assumes a common "thinking process" for all individuals, which may be too restrictive to accurately capture the full range of human variability in preferences, reasoning patterns, and response styles.

\noindent{\bf Parameter-based Personalization.} 
In this category, the first class of methods encodes user preferences directly into model parameters via full-parameter personalization, where a separate model is trained for each user by fine-tuning~\citep{kang2023llms,li2024learning,wang2023rolellm} or optimizing via reinforcement learning~\citep{jang2023personalized,wu2024fine} all model weights. While offering maximal flexibility, this approach is often prohibitively expensive in both storage and computation.
The second class of methods leverages parameter-efficient fine-tuning (PEFT), which introduces per-user adaptation modules—such as LoRA, while keeping the base model frozen~\citep{tan2024personalized,dan2024p,huang2024selective}.  
Next, we  will introduce OPPU (One PEFT Per User)~\citep{tan2024democratizing}, which is the current SOTA per-user LLM framework and provides a straightforward way to the personalized survey response prediction problem. 

\noindent{\bf OPPU for Personalized Survey Response Prediction.}
OPPU builds an independent parameter-efficient fine-tuning (PEFT) model for each user.
In practice, this often entails assigning each individual their own LoRA (Low-Rank Adaptation) module~\citep{hulora} as:
\[
   W_u = W_0 + s_1 AB + s_2 A_u B_u,
\]
where $W_0$ denotes the pretrained model weights, $AB$ is a shared low-rank adaptation trained on the entire dataset using standard LoRA (first stage, with $s_1$ to its scaling factor), and $A_u B_u$ represents the user-specific low-rank parameters. 
% In this setup, the shared component $AB$ captures global adaptation across all users, while personalization is introduced in a second stage by training the individual-specific parameters $(A_u, B_u)$ on top of the updated weights $W_0 + AB$ (with $s_2$ as its scaling factor). 

Note that a follow-up study~\citep{tan2024personalized} extends OPPU by allowing target users to select and assemble personalized PEFT modules from a shared pool using their historical data. This reduces storage costs by avoiding one fully independent PEFT per user, but it sacrifices accuracy relative to fully personalized models. Other approaches have proposed reinforcement learning to incorporate user-specific preferences via reward models~\citep{zhang2024personalization,guan2025surveypersonalizedalignment}. However, even with these advances, a foundational challenge remains: {\em how to represent each user within the LLM architecture in a way that is parameter-efficient, storage-optimized, structure-aware, and resistant to overfitting given the limited data available for each individual?}

\vspace{-3mm}
\section{Aplaud Approach}
\label{sec:method}
\vspace{-3mm}

\begin{figure*}[htbp]
  \centering
\includegraphics[scale=0.3]{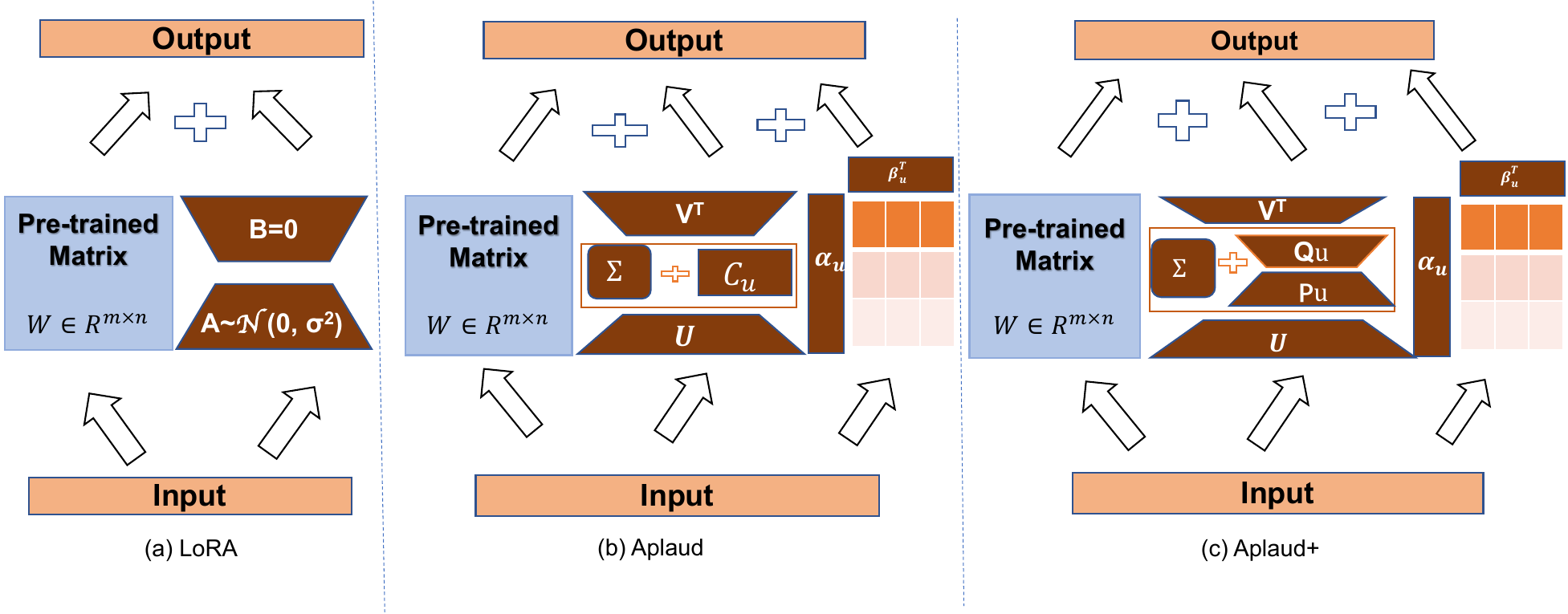}
  % \vspace*{-3.0ex}
  \caption{LoRA vs. Aplaud Overview}
  \label{fig:method}
  \vspace*{-3.0ex}
\end{figure*}

\textbf{Aplaud} (\textit{Adaptive Personalized Low-rank and User-specific Nested Decomposition}) is a novel parameter-efficient fine-tuning method designed to personalize large language models (LLMs) at the per-user level. It extends the standard LoRA framework by enabling scalable and expressive user-specific adaptation while minimizing the per-user parameter footprint. Figure~\ref{fig:method} illustrates the Aplaud method in comparison with LoRA, which provides only global (non-personalized) adaptation.

\subsection{Model Details: Compact Corrections and Residual Refinement}
To address the limitations of existing personalized LLMs such as OPPU~\cite{tan2024personalized}, which assigns each user an independent pair of matrices $(A_u, B_u)$, Aplaud instead reuses the shared low-rank update $AB$ learned in the first stage of a standard LoRA adaptation. Specifically, we apply singular value decomposition (SVD) to the global LoRA update:
\[
\Delta W = AB = U \Sigma V^\top,
\]
where \( U \in \mathbb{R}^{d\times r} \) and \( V \in \mathbb{R}^{d\times r} \) are orthogonal matrices capturing dominant update directions, and \( \Sigma \in \mathbb{R}^{r \times r} \) is a diagonal matrix of singular values \( (\sigma_1, \ldots, \sigma_r) \).

Intuitively, each singular vector \(V_i\) in \(V\) corresponds to a semantic direction against which the input $x$ is measured. The projection $V_i^\top x$ is scaled by the singular value $\sigma_i$ and then mapped to the corresponding output direction $U_i$, yielding a transformed coordinate $\sigma_i (V_i^\top x) U_i$. Because user-specific datasets are small and the semantic structure of survey questions is relatively stable across users, we hypothesize that the shared subspaces $U$ and $V$ capture most of the semantic directions needed for personalization. We further justify the shared-subspace assumption in Appendix~\ref{app:shared_subspace_assumption}, where we interpret ApLaud as a shared-subspace estimator and validate the learned \(V\)-subspace through probing experiments.

Building on this, instead of training new $(A_u, B_u)$ for each user, we inject a compact, user-specific correction matrix \( C_u \in \mathbb{R}^{r \times r} \) into the singular spectrum:
\[
\boxed{W_u = W_0 + \Delta W_u = W_0 + s\, U(\Sigma + C_u)V^\top,}
\]
where \( W_0 \) is the base model weight and $s$ is a scaling factor. The shared $U$ and $V$ are fixed across all users, while $C_u$ provides personalized adjustments. Importantly, $C_u$ is extremely lightweight: for rank $r=8$, it requires only $64$ parameters, and even for rank $r=64$, only $4096$, on the order of a single singular vector.

Compared with $U\Sigma V^\top$, the modified form $U(\Sigma+C_u)V^\top$ remains in the subspace spanned by $U$, i.e., 
\[
U(\Sigma+C_u)V^\top x \in \mathrm{Span}(\mathrm{Col}(U)).
\] 
However, $C_u$ enables each user to reweight and mix semantic directions in $V$, thereby reflecting their individual preferences and importance weights.

To capture fine-grained, idiosyncratic nuances beyond the shared subspace, Aplaud augments the representation with a lightweight personalized residual:
\[
\boxed{W_u = W_0 + s\, U(\Sigma + C_u)V^\top + \alpha_u \beta_u^\top,}
\]
where \( \alpha_u, \beta_u \in \mathbb{R}^d \) are learned per-user vectors. This rank-one residual enables Aplaud to adjust beyond the shared low-rank structure, modeling distinctive patterns that cannot be expressed solely within the subspace spanned by \(U\) and \(V\). 
In principle, the residual can be extended to higher rank, but we find that a rank-one correction is generally sufficient in our experimental settings (see Appendix for detailed results). 

\noindent{\bf Aplaud+: Nested Low-Rank Factorization.} 
To further compress the user-specific component and improve regularization, we factorize the correction matrix as
\[
C_u \approx P_u Q_u,
\]
yielding the personalized update
\[
\boxed{W_u = W_0+ sU (\Sigma + P_u Q_u) V^\top + \alpha_u \beta_u^\top,}
\]
where \(P_u, Q_u \in \mathbb{R}^{r \times k}\). 

Aplaud and Aplaud+ drastically reduce the per-user parameter footprint. 
For Aplaud, with rank \(r=64\), the correction matrix \(C_u\) requires only \(64 \times 64 = 4096\) parameters. 
For Aplaud+, using a nested inner rank of \(k=16\), the factorization \(P_u Q_u\) requires only \(2rk = 2048\) parameters per user for each weight matrix \(W\). 
The residual vectors \(\alpha_u, \beta_u \in \mathbb{R}^d\) introduce an additional \(2d\) parameters. 
By contrast, OPPU requires \(2dr = 524{,}288\) parameters when \(d=4096\). 
Thus, Aplaud and Aplaud+ achieve approximately \(128\times\) and \(256\times\) parameter reduction, respectively, without residual terms, and still over \(42\times\) and \(50\times\) reduction, respectively, when including the residual terms -- all while preserving expressive capacity.

\subsection{Training Procedure for Aplaud}
The training procedure for \textbf{Aplaud} consists of two stages, similar to other personalized LLM frameworks such as OPPU~\cite{tan2024democratizing}. 

\noindent\textbf{Stage 1: Global Adaptation with LoRA-style Training.} \\
We begin with a standard parameter-efficient fine-tuning (PEFT) procedure such as LoRA, applied across the full training dataset comprising all users’ responses. Specifically, we learn a global low-rank update:
\[
W = W_0 + \Delta W = W_0 + AB,
\]
where \( W_0 \in \mathbb{R}^{d \times d} \) is the pre-trained weight matrix, and \( A \in \mathbb{R}^{d \times r}, \, B \in \mathbb{R}^{r \times d} \) are trainable low-rank matrices. Following LoRA convention, \( A \) is initialized from a standard Gaussian distribution and \( B \) is initialized to zero, ensuring the pretrained model behavior is preserved at initialization. This stage captures population-level adaptation trends.  

After training, we compute the singular value decomposition (SVD):
\[
AB = U \Sigma V^\top,
\]
where \( U, V \in \mathbb{R}^{d \times r} \) and \( \Sigma \in \mathbb{R}^{r \times r} \). These components define a shared low-rank subspace, which remains fixed for all users in personalization stage.  

\noindent\textbf{Stage 2: Personalized Fine-tuning of \( C_u \) and Residual Terms.} \\
For each user \(u\), we fine-tune a compact correction matrix \(C_u\) together with residual vectors \(\alpha_u, \beta_u\). We initialize \(C_u = \mathbf{0}\), set \(\alpha_u \sim \mathcal{N}(0, I)\), and \(\beta_u = \mathbf{0}\). To stabilize training, we normalize \(\alpha_u\) and introduce a scaling factor \(m\), yielding:
\[
\boxed{W_u = W_0 + s\, U(\Sigma + C_u)V^\top + m \frac{\alpha_u}{\|\alpha_u\|}\,\beta_u^\top.}
\]
Here, $s$ controls the global scaling, while $m$ modulates the strength of the residual correction.  

\subsection{Aplaud+ Training Process}
In \textbf{Aplaud+}, the correction matrix \(C_u\) is further factorized into a nested low-rank form \(P_u Q_u\). Training proceeds in three substages:  

\noindent\textbf{Stage 2(a): Training $C_u$.}  
We first learn the full correction matrix \( C_u \in \mathbb{R}^{r \times r} \) using user \(u\)’s data:
\[
W_u = W_0 + s\, U(\Sigma + \gamma C_u)V^\top,
\]
where \(\gamma\) is a scaling factor similar to LoRA.  

\noindent\textbf{Stage 2(b): Low-Rank Factorization of \( C_u \).}  
Next, we compress \(C_u\) via SVD:
\[
C_u = U_C \Sigma_C V_C^\top.
\]
Truncating to a smaller rank \(k \ll r\), we initialize:
\[
P_u = U_C \Sigma_C^{1/2}, \quad Q_u = \Sigma_C^{1/2} V_C^\top,
\]
and re-train using user \(u\)’s data:
\[
W_u = W_0 + s\, U(\Sigma + P_u Q_u)V^\top.
\]

\noindent\textbf{Stage 2(c): Residual Learning.}  
Finally, we fine-tune residual vectors \(\alpha_u, \beta_u\):
\[
\boxed{W_u = W_0 + s\, U(\Sigma + P_u Q_u)V^\top + m \frac{\alpha_u}{\|\alpha_u\|}\,\beta_u^\top.}
\]

\noindent\textbf{Remarks.}  
These substages serve distinct purposes: Stage 2(a) help initialize $P_u, Q_u$ from a learned $C_u$; Stages 2(b) and 2(c) then train the nested low-rank and residual components. While they could in principle be trained jointly, we find that separating them improves stability. Despite the per-user independence of these stages, the small size of survey datasets (20–40 responses) allows training each user’s model in just 1–2 minutes on a single A100 GPU.  In addition, interestingly, from the theoretical justification, our proposed approach can be considered an efficient way for \textit{2-way Tucker Tensor Factorization} by considering all W as a three way tensor. The details of the theoretical and empirical justification can be found in Appendix~\ref{app:shared_subspace_assumption}. 

% \noindent\textbf{Personalized Parameter Size Summary.}  
% - For APlaud, each user is represented by a correction \(C_u \in \mathbb{R}^{r \times r}\) plus an optional rank-one residual \(\alpha_u \beta_u^\top\), for a total of \(r^2 + 2d\) parameters per weight matrix.  
% - For Aplaud+, we use \(P_u \in \mathbb{R}^{r \times k}, Q_u \in \mathbb{R}^{k \times r}\) with residual vectors, for a total of \(2rk + 2d\) parameters per weight matrix.  

In conclusion, APlaud and Aplaud+ provide fine-grained personalization with dramatically reduced per-user memory footprint, leveraging shared \(U, V, \Sigma\) for structure while adapting lightweight corrections and residuals for individual flexibility.
\vspace{-3mm}

\section{Experiment}
\label{sec:experiments}
\vspace{-2mm}
\textbf{Datasets} For survey data, we utilize data from two prominent sources of US public opinion: the annual Pew American Trends Panel (ATP) and the General Society Survey (GSS) \cite{pewATP}\cite{gss2022} for LLM simulation of survey and opinions\citep{santurkar2023whose}\cite{suh2025languagemodelfinetuningscaled}. We further evaluate our methods on LAMP Movie-Tagging\cite{salemi2023lamp}, a publicly available personalized dataset from a non-survey domain to guarantee dataset diversity. 
We would clarify that APLaud is not intended to be a universal solution for all personalization scenarios, but rather a method tailored to a specific settings where (i) users share a common task structure and (ii) per-user data are limited. However, it can still be used to achieve personalization on other tasks. To support this, we also empirically evaluate our methods in text generation task where we include results on other datasets from LaMP~\cite{salemi2023lamp} and LongLaMP datasets~\cite{kumar2024longlamp}. We include more details for data, results for text generation tasks, and experiment settings in appdendix ~\ref{appendix: exp settings} and ~\ref{appdix:generation_task}.

\textbf{Baseline} We compare our proposed method, APlaud, with \textcolor{black}{several set of baseline approaches}, including: (1) Non-personalized  LoRA and its variants: LoRA\cite{hu_lora_2021}, PiSSA\cite{meng2024pissa}, MiLoRA\cite{wang2024miloraharnessingminorsingular}, AdaLoRA\cite{zhang2023adaloraa} and QLoRA\cite{dettmers_qLoRA_2023}; \textcolor{black}{(2) Stronger proprietary model such as GPT-5 with profile only as best zero shot baseline; (3) Retrieval-based approach, where we also use GPT-5 and augments the prompt with the five most relevant historical QA pairs, inserted as few-shot exemplars;} (4) Personalized per user LLM: such as OPPU\cite{tan2024democratizing}. We employ Mistral-7B-v0.2-Instruct \citep{jiang2023mistral7b}, LLaMA-2-7B \cite{touvron2023llama} and Qwen2.5-7B~\cite{qwen2.5} (Qwen2.5-7B results in appendix~\ref{tab:qwen_results}) as backbone models to verify the robustness of our approach under different base architectures. 

\textbf{Evaluation} We use Accuracy and Macro-F1 as the main evaluation metrics. We additionally report Wasserstein Distance (WD) results in the appendix ~\ref{appendix: wd distance} to assess the discrepancy between predicted and ground-truth response distributions.
%Together, these metrics provide a holistic view of both classification accuracy and distributional alignment in personalized response generation.

%$W = W_{0}+U(\Sigma+P_uQ_{u}^T)V^{T}+m_{u}(AuBu/\|Bu\|_{F})$

\begin{table*}[t]
\centering
\caption{Performance comparison across different datasets with llama2-7B backbone. Bold numbers indicate the best results within each dataset. Dataset names: G\&L = ATP Gender \& Leadership, TS = ATP Trust in Science, F\&R = ATP Family and Relationships, EI = ATP Economic Inequality, GSS = General Social Survey. LAMP MV = LAMP Movie Tagging}
\label{tab:main_results_llam2}
\resizebox{\textwidth}{!}{%
\begin{tabular}{lcccccccccccc}
\toprule
\multirow{2}{*}{Method} & \multicolumn{2}{c}{G\&L} & \multicolumn{2}{c}{TS} & \multicolumn{2}{c}{F\&R} & \multicolumn{2}{c}{EI} & \multicolumn{2}{c}{GSS} & \multicolumn{2}{c}{LAMP MV} \\
\cmidrule(lr){2-3} \cmidrule(lr){4-5} \cmidrule(lr){6-7} \cmidrule(lr){8-9} \cmidrule(lr){10-11} \cmidrule(lr){12-13}
 & ACC & Macro-F1 & ACC & Macro-F1 & ACC & Macro-F1 & ACC & Macro-F1 & ACC & Macro-F1 & ACC & Macro-F1 \\
\midrule
\multicolumn{13}{l}{\textbf{Non-Personalized}} \\
LoRA       & 0.6393 & 0.6140 & 0.7052 & 0.5343 & 0.5772 & 0.3348 & 0.4617 & 0.3308 & 0.3785 & 0.2479 & 0.6214 & 0.5076 \\
PiSSA      & 0.6296 & 0.6142 & 0.7386 & 0.5620 & 0.5473 & 0.3427 & 0.4783 & 0.3722 & 0.3559 & 0.2333 & 0.6201 & 0.5280 \\
MiLoRA     & 0.6714 & 0.6634 & 0.7406 & 0.5768 & 0.5551 & 0.3503 & 0.4618 & 0.3522 & 0.3836 & 0.2681 & 0.6308 & 0.5341 \\
AdaLoRA    & 0.6700 & \textbf{0.6675} & 0.7779 & 0.5965 & 0.5564 & 0.3602 & 0.4716 & 0.3689 & 0.3907 & 0.2834 & 0.6146 & 0.5058 \\
QLoRA      & 0.6618 & 0.6475 & 0.7467 & 0.5316 & 0.5408 & 0.3411 & 0.4683 & 0.3296 & 0.3738 & 0.2531 & 0.6302 & 0.5253 \\
\midrule
\multicolumn{13}{l}{\textbf{Personalized}} \\
\textcolor{black}{GPT5-profile} &
\textcolor{black}{0.5394} &
\textcolor{black}{0.5340} &
\textcolor{black}{0.6170} &
\textcolor{black}{0.3448} &
\textcolor{black}{0.4954} &
\textcolor{black}{0.2879} &
\textcolor{black}{0.4566} &
\textcolor{black}{0.3491} &
\textcolor{black}{0.4689} &
\textcolor{black}{0.2570} &
\textcolor{black}{0.5478} &
\textcolor{black}{0.4507} \\
\textcolor{black}{GPT5-RAG} &
\textcolor{black}{0.6377} &
\textcolor{black}{0.6306} &
\textcolor{black}{0.7001} &
\textcolor{black}{0.6169} &
\textcolor{black}{0.6174} &
\textcolor{black}{0.3542} &
\textcolor{black}{\textbf{0.5213}} &
\textcolor{black}{0.4117} &
\textcolor{black}{\textbf{0.5106}} &
\textcolor{black}{\textbf{0.3520}} &
\textcolor{black}{-} &
\textcolor{black}{-} \\
OPPU       & 0.6651 & 0.6559 & 0.7548 & 0.6275 & 0.6096 & 0.3612 & 0.5008 & 0.4091 & 0.3701 & 0.2588 & 0.6336 & 0.5147 \\
Cu         & 0.6651 & 0.6453 & 0.7497 & 0.6423 & 0.5994 & 0.3475 & 0.4975 & 0.3611 & 0.3870 & 0.2590 & 0.6414 & 0.5358 \\
Aplaud    & 0.6731 & 0.6581 & 0.7761 & 0.6637 & 0.6151 & 0.3669 & 0.5042 & 0.3945 & 0.3912 & 0.2715 & 0.6442 & 0.5366 \\
Aplaud+ & \textbf{0.6828} & 0.6642 & \textbf{0.7974} & \textbf{0.6824} & \textbf{0.6381} & \textbf{0.3704} & 0.5183 & \textbf{0.4180} & 0.3969 & 0.2719 & \textbf{0.6593} & \textbf{0.5529} \\
\bottomrule
\end{tabular}}
\end{table*}

% mistral-7B result===============================================================================

\begin{table*}[t]
\centering
\caption{\textcolor{black}{Performance comparison across different datasets with Mistral-7B backbone. Bold numbers indicate the best results within each dataset. Dataset names: G\&L = ATP Gender \& Leadership, TS = ATP Trust in Science, F\&R = ATP Family and Relationships, EI = ATP Economic Inequality, GSS = General Social Survey. LAMP MV = LAMP Movie Tagging}}
\label{tab:main_results_mistral7B}
\resizebox{\textwidth}{!}{%
\begin{tabular}{lcccccccccccc}
\toprule
\multirow{2}{*}{\textcolor{black}{Method}} & \multicolumn{2}{c}{\textcolor{black}{G\&L}} & \multicolumn{2}{c}{\textcolor{black}{TS}} & \multicolumn{2}{c}{\textcolor{black}{F\&R}} & \multicolumn{2}{c}{\textcolor{black}{EI}} & \multicolumn{2}{c}{\textcolor{black}{GSS}} & \multicolumn{2}{c}{\textcolor{black}{LAMP MV}} \\
\cmidrule(lr){2-3} \cmidrule(lr){4-5} \cmidrule(lr){6-7} \cmidrule(lr){8-9} \cmidrule(lr){10-11} \cmidrule(lr){12-13}
 & \textcolor{black}{ACC} & \textcolor{black}{Macro-F1} & \textcolor{black}{ACC} & \textcolor{black}{Macro-F1} & \textcolor{black}{ACC} & \textcolor{black}{Macro-F1} & \textcolor{black}{ACC} & \textcolor{black}{Macro-F1} & \textcolor{black}{ACC} & \textcolor{black}{Macro-F1} & \textcolor{black}{ACC} & \textcolor{black}{Macro-F1} \\
\midrule
\multicolumn{13}{l}{\textcolor{black}{\textbf{Non-Personalized}}} \\
\textcolor{black}{Lora}      & \textcolor{black}{0.6554} & \textcolor{black}{0.6342} & \textcolor{black}{0.7072} & \textcolor{black}{0.4186} & \textcolor{black}{0.5681} & \textcolor{black}{0.2372} & \textcolor{black}{0.4708} & \textcolor{black}{0.3356} & \textcolor{black}{0.4336} & \textcolor{black}{0.2624} & \textcolor{black}{0.6669} & \textcolor{black}{0.5035} \\
\textcolor{black}{PiSSA}     & \textcolor{black}{0.6521} & \textcolor{black}{0.6466} & \textcolor{black}{0.7375} & \textcolor{black}{0.5169} & \textcolor{black}{0.5408} & \textcolor{black}{0.2387} & \textcolor{black}{0.4525} & \textcolor{black}{0.3119} & \textcolor{black}{0.4110} & \textcolor{black}{0.2832} & \textcolor{black}{0.6801} & \textcolor{black}{0.5114} \\
\textcolor{black}{MiLoRA}    & \textcolor{black}{0.6586} & \textcolor{black}{0.6499} & \textcolor{black}{0.7446} & \textcolor{black}{0.5039} & \textcolor{black}{0.5539} & \textcolor{black}{0.2572} & \textcolor{black}{0.4633} & \textcolor{black}{0.3297} & \textcolor{black}{0.4532} & \textcolor{black}{0.3146} & \textcolor{black}{0.6823} & \textcolor{black}{0.5014} \\
\textcolor{black}{AdaLoRA}   & \textcolor{black}{0.6425} & \textcolor{black}{0.6318} & \textcolor{black}{0.7071} & \textcolor{black}{0.4768} & \textcolor{black}{0.6005} & \textcolor{black}{0.3653} & \textcolor{black}{0.4708} & \textcolor{black}{0.3114} & \textcolor{black}{\textbf{0.5649}} & \textcolor{black}{\textbf{0.3573}} & \textcolor{black}{0.6635} & \textcolor{black}{0.5349} \\
\textcolor{black}{QLoRA}     & \textcolor{black}{0.6505} & \textcolor{black}{0.6368} & \textcolor{black}{0.7183} & \textcolor{black}{0.5127} & \textcolor{black}{0.5422} & \textcolor{black}{0.2486} & \textcolor{black}{0.4700} & \textcolor{black}{0.3307} & \textcolor{black}{0.4435} & \textcolor{black}{0.2903} & \textcolor{black}{0.7047} & \textcolor{black}{\textbf{0.5517}} \\
\midrule
\multicolumn{13}{l}{\textcolor{black}{\textbf{Personalized}}} \\
\textcolor{black}{GPT5-profile} &
\textcolor{black}{0.5394} &
\textcolor{black}{0.5340} &
\textcolor{black}{0.6170} &
\textcolor{black}{0.3448} &
\textcolor{black}{0.4954} &
\textcolor{black}{0.2879} &
\textcolor{black}{0.4566} &
\textcolor{black}{0.3491} &
\textcolor{black}{0.4689} &
\textcolor{black}{0.2570} &
\textcolor{black}{0.5478} &
\textcolor{black}{0.4507} \\
\textcolor{black}{GPT5-RAG} &
\textcolor{black}{0.6377} &
\textcolor{black}{0.6306} &
\textcolor{black}{0.7001} &
\textcolor{black}{0.6169} &
\textcolor{black}{0.6174} &
\textcolor{black}{0.3542} &
\textcolor{black}{\textbf{0.5213}} &
\textcolor{black}{\textbf{0.4117}} &
\textcolor{black}{0.5106} &
\textcolor{black}{0.3520} &
\textcolor{black}{-} &
\textcolor{black}{-} \\
\textcolor{black}{OPPU}      & \textcolor{black}{0.6731} & \textcolor{black}{0.6560} & \textcolor{black}{0.7852} & \textcolor{black}{0.6962} & \textcolor{black}{0.6368} & \textcolor{black}{0.3654} & \textcolor{black}{0.4925} & \textcolor{black}{0.3936} & \textcolor{black}{0.4322} & \textcolor{black}{0.2906} & \textcolor{black}{0.6917} & \textcolor{black}{0.4401} \\
\textcolor{black}{Cu}        & \textcolor{black}{0.6828} & \textcolor{black}{0.6691} & \textcolor{black}{0.7781} & \textcolor{black}{0.6338} & \textcolor{black}{0.5746} & \textcolor{black}{0.2796} & \textcolor{black}{0.5008} & \textcolor{black}{0.3868} & \textcolor{black}{0.4548} & \textcolor{black}{0.2917} & \textcolor{black}{0.6982} & \textcolor{black}{0.5014} \\
\textcolor{black}{Aplaud}    & \textcolor{black}{0.6828} & \textcolor{black}{0.6707} & \textcolor{black}{0.7852} & \textcolor{black}{0.7127} & \textcolor{black}{0.6278} & \textcolor{black}{0.3842} & \textcolor{black}{0.5042} & \textcolor{black}{0.3921} & \textcolor{black}{0.4506} & \textcolor{black}{0.3385} & \textcolor{black}{\textbf{0.7159}} & \textcolor{black}{0.5116} \\
\textcolor{black}{Aplaud+}   & \textcolor{black}{\textbf{0.6876}} & \textcolor{black}{\textbf{0.6742}} & \textcolor{black}{\textbf{0.7862}} & \textcolor{black}{\textbf{0.7204}} & \textcolor{black}{\textbf{0.6537}} & \textcolor{black}{\textbf{0.4369}} & \textcolor{black}{0.4992} & \textcolor{black}{0.3859} & \textcolor{black}{0.4605} & \textcolor{black}{0.3459} & \textcolor{black}{0.7081} & \textcolor{black}{0.5204} \\
\bottomrule
\end{tabular}}
\end{table*}

% param count====================================
% \begin{table*}[t]
% \centering
% \caption{Parameter count comparison per user per layer. We assume using Mistral-7B as foundation model and all LoRA-based PEFT with rank 64. For the SVD step, we retain the top 16 dimensions and bias term with rank 1.} \label{table: mem}

% \begin{tabular}{l l r r}
% \toprule
%  & Per-user per layer private parameter Calculation & \#params & Percentage \\
% \midrule
% OPPU & \texttt{4096×64×2×2 + (4096×64 + 1024×64)×2} &  &  \\
%      & \texttt{+(4096×64 + 14336×64)×3}  & 5,242,880 & 100\% \\
% SVD  & \texttt{(64×16×2)×7} & 14,336 & 0.27\% \\
% Aplaud & \texttt{(64×64)×7 + (4096×1×2×2 + (4096×1 + 1024×1)×2 } &  &  \\
%  & \texttt{ + (4096×1 + 14336×1)×3} & 110,599 & 2.11\% \\
% Aplaud+ & \texttt{(64×16×2)×7 + 4096×1×2×2  } &  &  \\
%  & \texttt{ + (4096×1 + 14336×1)×3} & 96,263 & 1.84\% \\
% \bottomrule
% \end{tabular}
% \end{table*}

\begin{table*}[t]
\centering
\caption{Parameter count comparison per user per layer. We assume using Mistral-7B as foundation model and all LoRA-based PEFT with rank 64. For the SVD step, we retain the top 16 dimensions and bias term with rank 1.} \label{table: mem}
\scriptsize
\setlength{\tabcolsep}{4pt}
\renewcommand{\arraystretch}{1.05}
\resizebox{0.92\textwidth}{!}{%
\begin{tabular}{llrr}
\toprule
Method & Calculation & \# Params & Percentage \\
\midrule
OPPU
&
\begin{tabular}[c]{@{}l@{}}
\(4096{\times}64{\times}2{\times}2 + (4096{\times}64 + 1024{\times}64){\times}2\) \\
\({}+ (4096{\times}64 + 14336{\times}64){\times}3\)
\end{tabular}
& 5,242,880 & 100.00\% \\

SVD
&
\((64{\times}16{\times}2){\times}7\)
& 14,336 & 0.27\% \\

APLaud
&
\begin{tabular}[c]{@{}l@{}}
\((64{\times}64){\times}7 + 4096{\times}1{\times}2{\times}2\) \\
\({}+ (4096{\times}1 + 1024{\times}1){\times}2 + (4096{\times}1 + 14336{\times}1){\times}3\)
\end{tabular}
& 110,599 & 2.11\% \\

APLaud+
&
\begin{tabular}[c]{@{}l@{}}
\((64{\times}16{\times}2){\times}7 + 4096{\times}1{\times}2{\times}2\) \\
\({}+ (4096{\times}1 + 14336{\times}1){\times}3\)
\end{tabular}
& 96,263 & 1.84\% \\
\bottomrule
\end{tabular}}
\vspace{-3mm}
\end{table*}

\textbf{Main Results}
% As shown in Table~\ref{tab:main_results_llam2} \textcolor{black}{and Table~\ref{tab:main_results_mistral7B}}, personalized approaches consistently surpass their non-personalized counterparts in terms of both accuracy (ACC) and macro-F1, highlighting the benefits of modeling user-specific adaptation. While adaptive non-personalized methods such as AdaLoRA and MiLoRA demonstrate relatively competitive performance, they remain inferior to personalized strategies on most datasets. 
\textcolor{black}{We first report our main results in Table~\ref{tab:main_results_llam2} and Table~\ref{tab:main_results_mistral7B}. We also report the Relative of Improvement Results using one of our methods(Aplaud+) in Table~\ref{tab:main_results_llam2_roi} and Table~\ref{tab:main_mistral7B_roi} in Appendix~\ref{sec: roi}. Aplaud families demonstrates markedly superior personalization capability across different datasets. Typically, on the Llama2-7B backbone, Aplaud+ delivers consistent gains of 6–10\% in ACC and 11–15\% in Macro-F1 over state-of-the-art non-personalized PEFT methods, and outperforms even SOTA general-purpose model (GPT5-profile) by 17.2\% ACC and 33.2\% Macro-F1 on average. Compared with the strong personalized adapter-based baseline OPPU, Aplaud also outperforms with a significant margin, improving 4.6\% in ACC and 4.5\% in Macro-F1. On the Mistral-7B backbone, the improvement is also pronounced, reaching 8–10\% ACC and 24–36\% Macro-F1 over non-personalized methods; surpassing retrieval-based approach (GPT5-RAG) 2.4\% ACC and 7.8\% Macro-F1 and outperform OPPU by 2.5\% ACC and 10.2\% Macro-F1.}

\textcolor{black}{Firstly, personalized approaches on average consistently surpass their non-personalized counterparts in terms of both accuracy (ACC) and macro-F1, highlighting the benefits of modeling user-specific adaptation. While adaptive non-personalized methods such as AdaLoRA and MiLoRA demonstrate relatively competitive performance, they remain inferior to personalized strategies on most datasets. }

\textcolor{black}{When comparing with stronger models and retrieval-based approaches (GPT-5 profile / GPT-5 RAG), our methods also demonstrate clear superiority. For example, in F\&R data and with Mistral-7B as backbone, Aplaud+ achieves as large as 51.8\% on Macro-F1, compared with zero-shot GPT and also 23.3\% compared with GPT-RAG. This disparity reflects a fundamental limitation of retrieval-based personalization: Although free of training, the quality of each generated response relies on a narrow subsample of history records, which inevitably cannot capture a respondent’s full behavioral signature, thus lacking personalization expressiveness. Especially in survey prediction tasks, a few retrieved samples rarely reflect the full spectrum of a user’s attitudes and easily omit key signals. In contrast, our lightweight adapter architecture accumulates user-specific signals across the full interaction history and encodes them as persistent parametric memory, forming a holistic and stable representation of user preferences.}

Compared with the representative personalized benchmark OPPU, our proposed Aplaud framework achieves consistent improvements across most datasets, yielding more balanced gains in both ACC and macro-F1. Notably, the enhanced variant Aplaud+ establishes new state-of-the-art performance on the majority of datasets, with \textcolor{black}{average ACC improvements 4.6\%, 2.5\% and Macro-F1 improvements 4.5\%, and 10.2\% on Llama2-7B and Mistrial-7B, respectively.} These results underscore that Aplaud not only advances beyond non-personalized tuning but also surpasses existing personalized approaches such as OPPU, thereby demonstrating the effectiveness of incorporating user-specific signals to enhance robustness and generalization across diverse domains. We defer additional experimental results to the appendix.

\textbf{Complexity Analysis and Parameter Efficiency.} In general, let $L$ denote the number of layers, $d$ the hidden dimension, and $r$ the LoRA rank. 
In APlaud, each user is represented by a correction matrix $C_u^{(l)} \in \mathbb{R}^{r \times r}$ together with a rank-1 residual $(\alpha_u^{(l)}, \beta_u^{(l)})$, resulting in a per-user complexity of $\mathcal{O}(L r^2 + L d)$. 
For Aplaud+, we further factorize $C_u^{(l)} \approx P_u^{(l)} Q_u^{(l)}$ with $P_u^{(l)} \in \mathbb{R}^{r \times k}$ and $Q_u^{(l)} \in \mathbb{R}^{k \times r}$ ($k \ll r$), reducing the complexity to $\mathcal{O}(L r k + L d)$. 
In contrast, OPPU requires $\mathcal{O}(L d r)$ parameters per user. 
Since $r, k \ll d$, our method operates primarily in the low-rank space and is significantly more scalable for large models. 

Table~\ref{table: mem} compares the per-user parameter count per layer across methods. 
A standard OPPU design requires over 5M parameters per user per layer, since it places independent adapters on all \emph{seven weight matrices} in each transformer block: four in the attention module ($W_q, W_k, W_v, W_o$) and three in the feed-forward module ($W_{\text{up}}, W_{\text{gate}}, W_{\text{down}}$). 
This heavy footprint makes OPPU impractical to scale across large user populations. 

By contrast, our methods dramatically reduce this overhead. 
The pure SVD ($C_u \approx P_u Q_u$ ) variant compresses the personalization into a compact shared subspace, achieving a $99.7\%$ reduction in parameter size. 
APlaud introduces lightweight user-specific corrections and a rank-one residual, requiring only about $2\%$ of the OPPU footprint, while Aplaud+ further factorizes the corrections to reduce usage to under $2\%$. 
These results highlight the strong parameter efficiency of our framework, which balances compactness with sufficient expressive capacity to yield substantial performance improvements.

To further highlight the significance of APLaud's parameter efficiency, we conduct a performance--parameter tradeoff analysis by varying OPPU's LoRA rank to control its per-user parameter budget. As shown in Appendix~\ref{app:performance_tradeoff}, APLaud/APLaud+ achieve better performance with substantially fewer user-specific trainable parameters.

%Finally, we note that the ablation study and additional experiments can be found in the Appendix. 

% Table \ref{table: mem} compares the per-user parameter count per layer. While OPPU requires over 5M parameters, our methods dramatically reduce the overhead. The SVD variant achieves  99.7\% reduction, and even adding a bias term (SVD + bias) still keeps the size under 2\% of OPPU. This demonstrates the strong parameter efficiency of our approach while providing substantial performance improvements.

\vspace*{-3.0mm}

\section{Conclusion}
\label{sec:conclusion}
\vspace*{-1.4ex}
In this work, we introduced \textbf{APlaud}, a scalable and lightweight framework for personalizing large language models (LLMs) at the individual user level in the context of survey response prediction. APlaud leverages a shared low-rank subspace obtained through global LoRA fine-tuning, while enabling user-specific adaptation via a nested low-rank correction and an optional rank-one residual. This design achieves strong personalization with minimal per-user parameter overhead. 
%Our approach addresses key challenges in user modeling—such as data sparsity and scalability -- while consistently outperforming existing LoRA-based personalized methods in both accuracy and parameter efficiency. These results highlight the promise of APlaud for simulating individualized behavior in large-scale settings, offering a principled bridge between global adaptation and fine-grained user representation.

%In future work, we plan to extend APlaud to a broader range of personalization tasks, including recommendation systems, writing  assistants, and other applications where lightweight and expressive user modeling is critical. We also plan to integrate it with the quantization approach to further reduce parameter space. 

\section*{Limitations}

\paragraph{Mitigations and Scope.}
APlaud is designed to model individual response patterns \emph{conditional on existing survey data} and is not intended to replace real human respondents in high-stakes decision-making, policy formation, or sensitive social inference. We emphasize that personalized synthetic responses should be used as a complementary analytical tool rather than a substitute for direct human engagement. Moreover, the method operates within the constraints of parameter-efficient fine-tuning and shared representations, which naturally limit memorization and reduce the risk of exposing sensitive information. Responsible deployment should include transparency about synthetic data usage, appropriate human oversight, and adherence to existing ethical guidelines for survey research and data governance.

\paragraph{Inference Effiency} Additionally, we want to mention that although per-user adapter-based personalization can introduce serving challenges, in Modern LLM serving frameworks already support efficient cache management, and our compact user modules integrate naturally with such architecture and reducing LoRA parameter to 1\% is already a big step forwar.

Overall, we believe that APlaud represents a measured and responsible advance toward scalable personalization in machine learning. While it introduces new capabilities for modeling individual-level preferences, the framework is designed to support exploratory analysis and system development, rather than to replace human judgment or participation in real world data collection.

% \section*{Acknowledgments}

% This document has been adapted
% by Steven Bethard, Ryan Cotterell and Rui Yan
% from the instructions for earlier ACL and NAACL proceedings, including those for
% ACL 2019 by Douwe Kiela and Ivan Vuli\'{c},
% NAACL 2019 by Stephanie Lukin and Alla Roskovskaya,
% ACL 2018 by Shay Cohen, Kevin Gimpel, and Wei Lu,
% NAACL 2018 by Margaret Mitchell and Stephanie Lukin,
% Bib\TeX{} suggestions for (NA)ACL 2017/2018 from Jason Eisner,
% ACL 2017 by Dan Gildea and Min-Yen Kan,
% NAACL 2017 by Margaret Mitchell,
% ACL 2012 by Maggie Li and Michael White,
% ACL 2010 by Jing-Shin Chang and Philipp Koehn,
% ACL 2008 by Johanna D. Moore, Simone Teufel, James Allan, and Sadaoki Furui,
% ACL 2005 by Hwee Tou Ng and Kemal Oflazer,
% ACL 2002 by Eugene Charniak and Dekang Lin,
% and earlier ACL and EACL formats written by several people, including
% John Chen, Henry S. Thompson and Donald Walker.
% Additional elements were taken from the formatting instructions of the \emph{International Joint Conference on Artificial Intelligence} and the \emph{Conference on Computer Vision and Pattern Recognition}.

% Bibliography entries for the entire Anthology, followed by custom entries
%\bibliography{custom,anthology-overleaf-1,anthology-overleaf-2}

\newpage
\bibliography{lora,LLMsocialSim,references-perLLM,references-survey,custom}

@misc{hu_lora_2021,
	title = {{LoRA}: {Low}-{Rank} {Adaptation} of {Large} {Language} {Models}},
	shorttitle = {{LoRA}},
	url = {https://openreview.net/forum?id=nZeVKeeFYf9},
	language = {en},
	urldate = {2025-02-03},
	publisher = {International Conference on Learning Representations},
	author = {Hu, Edward J. and Shen, Yelong and Wallis, Phillip and Allen-Zhu, Zeyuan and Li, Yuanzhi and Wang, Shean and Wang, Lu and Chen, Weizhu},
	month = oct,
	year = {2021},
}

@inproceedings{dettmers_qlora_2023,
	address = {Red Hook, NY, USA},
	series = {{NIPS} '23},
	title = {{QLORA}: efficient finetuning of quantized {LLMs}},
	shorttitle = {{QLORA}},
	urldate = {2025-02-02},
	booktitle = {Proceedings of the 37th {International} {Conference} on {Neural} {Information} {Processing} {Systems}},
	publisher = {Curran Associates Inc.},
	author = {Dettmers, Tim and Pagnoni, Artidoro and Holtzman, Ari and Zettlemoyer, Luke},
	month = dec,
	year = {2023},
	pages = {10088--10115},
}

@misc{gao_take_2024,
	title = {Take {Caution} in {Using} {LLMs} as {Human} {Surrogates}: {Scylla} {Ex} {Machina}},
	shorttitle = {Take {Caution} in {Using} {LLMs} as {Human} {Surrogates}},
	url = {http://arxiv.org/abs/2410.19599},
	doi = {10.48550/arXiv.2410.19599},
	urldate = {2024-12-20},
	publisher = {arXiv},
	author = {Gao, Yuan and Lee, Dokyun and Burtch, Gordon and Fazelpour, Sina},
	month = nov,
	year = {2024},
	note = {arXiv:2410.19599 [econ]},
}

@inproceedings{aher_using_2023,
	address = {Honolulu, Hawaii, USA},
	series = {{ICML}'23},
	title = {Using large language models to simulate multiple humans and replicate human subject studies},
	volume = {202},
	urldate = {2024-12-01},
	booktitle = {Proceedings of the 40th {International} {Conference} on {Machine} {Learning}},
	publisher = {JMLR.org},
	author = {Aher, Gati and Arriaga, Rosa I. and Kalai, Adam Tauman},
	month = jul,
	year = {2023},
	pages = {337--371},
}

@techreport{horton_large_2023,
	address = {Cambridge, MA},
	title = {Large {Language} {Models} as {Simulated} {Economic} {Agents}: {What} {Can} {We} {Learn} from {Homo} {Silicus}?},
	shorttitle = {Large {Language} {Models} as {Simulated} {Economic} {Agents}},
	language = {english},
	number = {w31122},
	urldate = {2023-10-21},
	institution = {National Bureau of Economic Research},
	author = {Horton, John},
	month = apr,
	year = {2023},
	doi = {10.3386/w31122},
	pages = {w31122},
}

@misc{kim_ai-augmented_2024,
	title = {{AI}-{Augmented} {Surveys}: {Leveraging} {Large} {Language} {Models} and {Surveys} for {Opinion} {Prediction}},
	shorttitle = {{AI}-{Augmented} {Surveys}},
	urldate = {2024-08-15},
	publisher = {arXiv},
	author = {Kim, Junsol and Lee, Byungkyu},
	month = apr,
	year = {2024},
	note = {arXiv: 2305.09620 [cs]
Number: arXiv:2305.09620},
}

@article{bisbee_synthetic_2024,
	title = {Synthetic {Replacements} for {Human} {Survey} {Data}? {The} {Perils} of {Large} {Language} {Models}},
	volume = {32},
	issn = {1047-1987, 1476-4989},
	shorttitle = {Synthetic {Replacements} for {Human} {Survey} {Data}?},
	doi = {10.1017/pan.2024.5},
	language = {english},
	number = {4},
	urldate = {2024-12-01},
	journal = {Political Analysis},
	author = {Bisbee, James and Clinton, Joshua D. and Dorff, Cassy and Kenkel, Brenton and Larson, Jennifer M.},
	month = oct,
	year = {2024},
	pages = {401--416},
}

@misc{wang_not_2024,
	title = {Not {Yet}: {Large} {Language} {Models} {Cannot} {Replace} {Human} {Respondents} for {Psychometric} {Research}},
	shorttitle = {Not {Yet}},
	urldate = {2024-12-01},
	author = {Wang, Pengda and Zou, Huiqi and Yan, Zihan and Guo, Feng and Sun, Tianjun and Xiao, Ziang and Zhang, Bo},
	month = sep,
	year = {2024},
	doi = {10.31219/osf.io/rwy9b},
}

@misc{anthis2025llmsocialsimulationspromising,
      title={LLM Social Simulations Are a Promising Research Method}, 
      author={Jacy Reese Anthis and Ryan Liu and Sean M. Richardson and Austin C. Kozlowski and Bernard Koch and James Evans and Erik Brynjolfsson and Michael Bernstein},
      year={2025},
      eprint={2504.02234},
      archivePrefix={arXiv},
      primaryClass={cs.HC},
      url={https://arxiv.org/abs/2504.02234}, 
}

@misc{guan2025surveypersonalizedalignment,
      title={A Survey on Personalized Alignment -- The Missing Piece for Large Language Models in Real-World Applications}, 
      author={Jian Guan and Junfei Wu and Jia-Nan Li and Chuanqi Cheng and Wei Wu},
      year={2025},
      eprint={2503.17003},
      archivePrefix={arXiv},
      primaryClass={cs.CL},
      url={https://arxiv.org/abs/2503.17003}, 
}

@misc{suh2025languagemodelfinetuningscaled,
      title={Language Model Fine-Tuning on Scaled Survey Data for Predicting Distributions of Public Opinions}, 
      author={Joseph Suh and Erfan Jahanparast and Suhong Moon and Minwoo Kang and Serina Chang},
      year={2025},
      eprint={2502.16761},
      archivePrefix={arXiv},
      primaryClass={cs.CL},
      url={https://arxiv.org/abs/2502.16761}, 
}

@article{yang2024rewards,
  title={Rewards-in-context: Multi-objective alignment of foundation models with dynamic preference adjustment},
  author={Yang, Rui and Pan, Xiaoman and Luo, Feng and Qiu, Shuang and Zhong, Han and Yu, Dong and Chen, Jianshu},
  journal={arXiv preprint arXiv:2402.10207},
  year={2024}
}

@article{wu2024fine,
  title={Fine-grained human feedback gives better rewards for language model training},
  author={Wu, Zeqiu and Hu, Yushi and Shi, Weijia and Dziri, Nouha and Suhr, Alane and Ammanabrolu, Prithviraj and Smith, Noah A and Ostendorf, Mari and Hajishirzi, Hannaneh},
  journal={Advances in Neural Information Processing Systems},
  volume={36},
  year={2024}
}

@inproceedings{dong2023steerlm,
    title = "{S}teer{LM}: Attribute Conditioned {SFT} as an (User-Steerable) Alternative to {RLHF}",
    author = "Dong, Yi  and
      Wang, Zhilin  and
      Sreedhar, Makesh  and
      Wu, Xianchao  and
      Kuchaiev, Oleksii",
    editor = "Bouamor, Houda  and
      Pino, Juan  and
      Bali, Kalika",
    booktitle = "Findings of the Association for Computational Linguistics: EMNLP 2023",
    month = dec,
    year = "2023",
    address = "Singapore",
    publisher = "Association for Computational Linguistics",
    url = "https://aclanthology.org/2023.findings-emnlp.754",
    doi = "10.18653/v1/2023.findings-emnlp.754",
    pages = "11275--11288",
}

@misc{qwen2.5,
    title = {Qwen2.5: A Party of Foundation Models},
    url = {https://qwenlm.github.io/blog/qwen2.5/},
    author = {Qwen Team},
    month = {September},
    year = {2024}
}

@techreport{qualtrics2025market,
  title        = {2025 Global Market Research Trends Report},
  author       = {{Qualtrics}},
  institution  = {Qualtrics},
  year         = {2025},
  note         = {Retrieved from \url{https://www.qualtrics.com/ebooks-guides/market-research-trends/}},
  url          = {https://www.qualtrics.com/ebooks-guides/market-research-trends/}
}

@inproceedings{li2023text,
  title={Text is all you need: Learning language representations for sequential recommendation},
  author={Li, Jiacheng and Wang, Ming and Li, Jin and Fu, Jinmiao and Shen, Xin and Shang, Jingbo and McAuley, Julian},
  booktitle={Proceedings of the 29th ACM SIGKDD Conference on Knowledge Discovery and Data Mining},
  pages={1258--1267},
  year={2023}
}

@inproceedings{bao2023tallrec,
  title={Tallrec: An effective and efficient tuning framework to align large language model with recommendation},
  author={Bao, Keqin and Zhang, Jizhi and Zhang, Yang and Wang, Wenjie and Feng, Fuli and He, Xiangnan},
  booktitle={Proceedings of the 17th ACM Conference on Recommender Systems},
  pages={1007--1014},
  year={2023}
}

@article{huang2024selective,
  title={Selective Prompting Tuning for Personalized Conversations with LLMs},
  author={Huang, Qiushi and Liu, Xubo and Ko, Tom and Wu, Bo and Wang, Wenwu and Zhang, Yu and Tang, Lilian},
  journal={arXiv preprint arXiv:2406.18187},
  year={2024}
}

@article{gong2025latent,
  title={Latent Preference Coding: Aligning Large Language Models via Discrete Latent Codes},
  author={Gong, Zhuocheng and Guan, Jian and Wu, Wei and Zhang, Huishuai and Zhao, Dongyan and Yan, Rui},
  year={2024}
}

@article{kang2023llms,
  title={Do llms understand user preferences? evaluating llms on user rating prediction},
  author={Kang, Wang-Cheng and Ni, Jianmo and Mehta, Nikhil and Sathiamoorthy, Maheswaran and Hong, Lichan and Chi, Ed and Cheng, Derek Zhiyuan},
  journal={arXiv preprint arXiv:2305.06474},
  year={2023}
}

@article{wang2023rolellm,
  title={Rolellm: Benchmarking, eliciting, and enhancing role-playing abilities of large language models},
  author={Wang, Zekun Moore and Peng, Zhongyuan and Que, Haoran and Liu, Jiaheng and Zhou, Wangchunshu and Wu, Yuhan and Guo, Hongcheng and Gan, Ruitong and Ni, Zehao and Yang, Jian and others},
  journal={arXiv preprint arXiv:2310.00746},
  year={2023}
}

@article{tan2024democratizing,
  title={Democratizing large language models via personalized parameter-efficient fine-tuning},
  author={Tan, Zhaoxuan and Zeng, Qingkai and Tian, Yijun and Liu, Zheyuan and Yin, Bing and Jiang, Meng},
  journal={arXiv preprint arXiv:2402.04401},
  year={2024}
}

@article{dan2024p,
  title={P-Tailor: Customizing Personality Traits for Language Models via Mixture of Specialized LoRA Experts},
  author={Dan, Yuhao and Zhou, Jie and Chen, Qin and Tian, Junfeng and He, Liang},
  journal={arXiv preprint arXiv:2406.12548},
  year={2024}
}

@inproceedings{xu2022long,
  title={Long Time No See! Open-Domain Conversation with Long-Term Persona Memory},
  author={Xu, Xinchao and Gou, Zhibin and Wu, Wenquan and Niu, Zheng-Yu and Wu, Hua and Wang, Haifeng and Wang, Shihang},
  booktitle={Findings of the Association for Computational Linguistics: ACL 2022},
  pages={2639--2650},
  year={2022}
}

@article{wang2024unims,
  title={Unims-rag: A unified multi-source retrieval-augmented generation for personalized dialogue systems},
  author={Wang, Hongru and Huang, Wenyu and Deng, Yang and Wang, Rui and Wang, Zezhong and Wang, Yufei and Mi, Fei and Pan, Jeff Z and Wong, Kam-Fai},
  journal={arXiv preprint arXiv:2401.13256},
  year={2024}
}

@misc{li20251000000usersuserscaling,
      title={From 1,000,000 Users to Every User: Scaling Up Personalized Preference for User-level Alignment}, 
      author={Jia-Nan Li and Jian Guan and Songhao Wu and Wei Wu and Rui Yan},
      year={2025},
      eprint={2503.15463},
      archivePrefix={arXiv},
      primaryClass={cs.CL},
      url={https://arxiv.org/abs/2503.15463}, 
}

@article{shenfeld2025language,
  title={Language Model Personalization via Reward Factorization},
  author={Shenfeld, Idan and Faltings, Felix and Agrawal, Pulkit and Pacchiano, Aldo},
  journal={arXiv preprint arXiv:2503.06358},
  year={2025}
}

@inproceedings{
chen2025pal,
title={{PAL}: Sample-Efficient Personalized Reward Modeling for Pluralistic Alignment},
author={Daiwei Chen and Yi Chen and Aniket Rege and Zhi Wang and Ramya Korlakai Vinayak},
booktitle={The Thirteenth International Conference on Learning Representations},
year={2025},
url={https://openreview.net/forum?id=1kFDrYCuSu}
}

@article{tucker1966some,
  title={Some mathematical notes on three-mode factor analysis},
  author={Tucker, Ledyard R},
  journal={Psychometrika},
  volume={31},
  number={3},
  pages={279--311},
  year={1966},
  publisher={Springer}
}

@article{kolda2009tensor,
  title={Tensor decompositions and applications},
  author={Kolda, Tamara G and Bader, Brett W},
  journal={SIAM review},
  volume={51},
  number={3},
  pages={455--500},
  year={2009},
  publisher={SIAM}
}

@inproceedings{hulora,
  title={LoRA: Low-Rank Adaptation of Large Language Models},
  author={Hu, Edward J and Wallis, Phillip and Allen-Zhu, Zeyuan and Li, Yuanzhi and Wang, Shean and Wang, Lu and Chen, Weizhu and others},
  booktitle={International Conference on Learning Representations},
  year={2021}
}

@misc{zhang2023adaloraa,
      title={AdaLoRA: Adaptive Budget Allocation for Parameter-Efficient Fine-Tuning}, 
      author={Qingru Zhang and Minshuo Chen and Alexander Bukharin and Nikos Karampatziakis and Pengcheng He and Yu Cheng and Weizhu Chen and Tuo Zhao},
      year={2023},
      eprint={2303.10512},
      archivePrefix={arXiv},
      primaryClass={cs.CL},
      url={https://arxiv.org/abs/2303.10512}, 
}

@misc{wang2024miloraharnessingminorsingular,
      title={MiLoRA: Harnessing Minor Singular Components for Parameter-Efficient LLM Finetuning}, 
      author={Hanqing Wang and Yixia Li and Shuo Wang and Guanhua Chen and Yun Chen},
      year={2024},
      eprint={2406.09044},
      archivePrefix={arXiv},
      primaryClass={cs.CL},
      url={https://arxiv.org/abs/2406.09044}, 
}

@inproceedings{
meng2024pissa,
title={Pi{SSA}: Principal Singular Values and Singular Vectors Adaptation of Large Language Models},
author={Fanxu Meng and Zhaohui Wang and Muhan Zhang},
booktitle={The Thirty-eighth Annual Conference on Neural Information Processing Systems},
year={2024},
url={https://openreview.net/forum?id=6ZBHIEtdP4}
}

@misc{wang2024kasaknowledgeawaresingularvalueadaptation,
      title={KaSA: Knowledge-Aware Singular-Value Adaptation of Large Language Models}, 
      author={Fan Wang and Juyong Jiang and Chansung Park and Sunghun Kim and Jing Tang},
      year={2024},
      eprint={2412.06071},
      archivePrefix={arXiv},
      primaryClass={cs.CL},
      url={https://arxiv.org/abs/2412.06071}, 
}

@article{kumar2024longlamp,
 author = {Kumar, Ishita and Viswanathan, Snigdha and Yerra, Sushrita and Salemi, Alireza and Rossi, Ryan A and Dernoncourt, Franck and Deilamsalehy, Hanieh and Chen, Xiang and Zhang, Ruiyi and Agarwal, Shubham and others},
 journal = {arXiv:2407.11016},
 title = {Longlamp: A benchmark for personalized long-form text generation},
 year = {2024}
}

@article{lee2024aligning,
 author = {Lee, Seongyun and Park, Sue Hyun and Kim, Seungone and Seo, Minjoon},
 journal = {arXiv:2405.17977},
 title = {Aligning to thousands of preferences via system message generalization},
 year = {2024}
}

@inproceedings{li2024learning,
 author = {Li, Cheng and Zhang, Mingyang and Mei, Qiaozhu and Kong, Weize and Bendersky, Michael},
 booktitle = {Proc. of Web Conference},
 title = {Learning to Rewrite Prompts for Personalized Text Generation},
 year = {2024}
}

@article{ning2024user,
 author = {Ning, Lin and Liu, Luyang and Wu, Jiaxing and Wu, Neo and Berlowitz, Devora and Prakash, Sushant and Green, Bradley and O'Banion, Shawn and Xie, Jun},
 journal = {arXiv:2402.13598},
 title = {User-LLM: Efficient LLM Contextualization with User Embeddings},
 year = {2024}
}

@article{poddar2024personalizing,
 author = {Poddar, Sriyash and Wan, Yanming and Ivison, Hamish and Gupta, Abhishek and Jaques, Natasha},
 journal = {arXiv:2408.10075},
 title = {Personalizing reinforcement learning from human feedback with variational preference learning},
 year = {2024}
}

@article{tan2024personalized,
 author = {Tan, Zhaoxuan and Liu, Zheyuan and Jiang, Meng},
 journal = {arXiv:2406.10471},
 title = {Personalized Pieces: Efficient Personalized Large Language Models through Collaborative Efforts},
 year = {2024}
}

@article{jang2023personalized,
  title={Personalized soups: Personalized large language model alignment via post-hoc parameter merging},
  author={Jang, Joel and Kim, Seungone and Lin, Bill Yuchen and Wang, Yizhong and Hessel, Jack and Zettlemoyer, Luke and Hajishirzi, Hannaneh and Choi, Yejin and Ammanabrolu, Prithviraj},
  journal={arXiv:2310.11564},
  year={2023}
}

@article{zhang2024personalization,
  title={Personalization of large language models: A survey},
  author={Zhang, Zhehao and Rossi, Ryan A and Kveton, Branislav and Shao, Yijia and Yang, Diyi and Zamani, Hamed and Dernoncourt, Franck and Barrow, Joe and Yu, Tong and Kim, Sungchul and others},
  journal={arXiv:2411.00027},
  year={2024}
}

@misc{jiang2023mistral7b,
      title={Mistral 7B}, 
      author={Albert Q. Jiang and Alexandre Sablayrolles and Arthur Mensch and Chris Bamford and Devendra Singh Chaplot and Diego de las Casas and Florian Bressand and Gianna Lengyel and Guillaume Lample and Lucile Saulnier and Lélio Renard Lavaud and Marie-Anne Lachaux and Pierre Stock and Teven Le Scao and Thibaut Lavril and Thomas Wang and Timothée Lacroix and William El Sayed},
      year={2023},
      eprint={2310.06825},
      archivePrefix={arXiv},
      primaryClass={cs.CL},
      url={https://arxiv.org/abs/2310.06825}, 
}

@misc{pewATP,
  author       = {Pew Research Center},
  title        = {American Trends Panel Datasets},
  year         = {2025},
  url          = {https://www.pewresearch.org/american-trends-panel/},
  note         = {Accessed: 2025-05-15}
}

@article{salemi2023lamp,
  title={Lamp: When large language models meet personalization},
  author={Salemi, Alireza and Mysore, Sheshera and Bendersky, Michael and Zamani, Hamed},
  journal={arXiv preprint arXiv:2304.11406},
  year={2023}
}

@article{touvron2023llama,
  title={Llama 2: Open foundation and fine-tuned chat models},
  author={Touvron, Hugo and Martin, Louis and Stone, Kevin and Albert, Peter and Almahairi, Amjad and Babaei, Yasmine and Bashlykov, Nikolay and Batra, Soumya and Bhargava, Prajjwal and Bhosale, Shruti and others},
  journal={arXiv preprint arXiv:2307.09288},
  year={2023}
}

@inproceedings{he2016fast,
  title={Fast matrix factorization for online recommendation with implicit feedback},
  author={He, Xiangnan and Zhang, Hanwang and Kan, Min-Yen and Chua, Tat-Seng},
  booktitle={Proceedings of the 39th International ACM SIGIR conference on Research and Development in Information Retrieval},
  pages={549--558},
  year={2016}
}

@inproceedings{he2017neural,
  title={Neural collaborative filtering},
  author={He, Xiangnan and Liao, Lizi and Zhang, Hanwang and Nie, Liqiang and Hu, Xia and Chua, Tat-Seng},
  booktitle={Proceedings of the 26th international conference on world wide web},
  pages={173--182},
  year={2017}
}

@inproceedings{he2020lightgcn,
  title={Lightgcn: Simplifying and powering graph convolution network for recommendation},
  author={He, Xiangnan and Deng, Kuan and Wang, Xiang and Li, Yan and Zhang, Yongdong and Wang, Meng},
  booktitle={Proceedings of the 43rd International ACM SIGIR conference on research and development in Information Retrieval},
  pages={639--648},
  year={2020}
}

@inproceedings{dai2023uncovering,
  title={Uncovering chatgpt’s capabilities in recommender systems},
  author={Dai, Sunhao and Shao, Ninglu and Zhao, Haiyuan and Yu, Weijie and Si, Zihua and Xu, Chen and Sun, Zhongxiang and Zhang, Xiao and Xu, Jun},
  booktitle={Proceedings of the 17th ACM Conference on Recommender Systems},
  pages={1126--1132},
  year={2023}
}

@article{peng2024ecellm,
  title={ecellm: Generalizing large language models for e-commerce from large-scale, high-quality instruction data},
  author={Peng, Bo and Ling, Xinyi and Chen, Ziru and Sun, Huan and Ning, Xia},
  journal={arXiv preprint arXiv:2402.08831},
  year={2024}
}

@article{wang2023rethinking,
  title={Rethinking the evaluation for conversational recommendation in the era of large language models},
  author={Wang, Xiaolei and Tang, Xinyu and Zhao, Wayne Xin and Wang, Jingyuan and Wen, Ji-Rong},
  journal={arXiv preprint arXiv:2305.13112},
  year={2023}
}

@inproceedings{geng2022recommendation,
  title={Recommendation as language processing (rlp): A unified pretrain, personalized prompt \& predict paradigm (p5)},
  author={Geng, Shijie and Liu, Shuchang and Fu, Zuohui and Ge, Yingqiang and Zhang, Yongfeng},
  booktitle={Proceedings of the 16th ACM conference on recommender systems},
  pages={299--315},
  year={2022}
}

@article{cui2022m6,
  title={M6-rec: Generative pretrained language models are open-ended recommender systems},
  author={Cui, Zeyu and Ma, Jianxin and Zhou, Chang and Zhou, Jingren and Yang, Hongxia},
  journal={arXiv preprint arXiv:2205.08084},
  year={2022}
}

@article{li2023gpt4rec,
  title={GPT4Rec: A generative framework for personalized recommendation and user interests interpretation},
  author={Li, Jinming and Zhang, Wentao and Wang, Tian and Xiong, Guanglei and Lu, Alan and Medioni, Gerard},
  journal={arXiv preprint arXiv:2304.03879},
  year={2023}
}

@article{valizadeh2025language,
  title={Language Models as Semantic Augmenters for Sequential Recommenders},
  author={Valizadeh, Mahsa and Dong, Xiangjue and Tuo, Rui and Caverlee, James},
  journal={arXiv preprint arXiv:2510.18046},
  year={2025}
}

@inproceedings{gao2025llm4rerank,
  title={Llm4rerank: Llm-based auto-reranking framework for recommendations},
  author={Gao, Jingtong and Chen, Bo and Zhao, Xiangyu and Liu, Weiwen and Li, Xiangyang and Wang, Yichao and Wang, Wanyu and Guo, Huifeng and Tang, Ruiming},
  booktitle={Proceedings of the ACM on Web Conference 2025},
  pages={228--239},
  year={2025}
}

@inproceedings{kornblith2019similarity,
  title={Similarity of neural network representations revisited},
  author={Kornblith, Simon and Norouzi, Mohammad and Lee, Honglak and Hinton, Geoffrey},
  booktitle={International conference on machine learning},
  pages={3519--3529},
  year={2019},
  organization={PMlR}
}

@article{sun2024random,
  title     = {Random Silicon Sampling: Simulating Human Sub-Population Opinion Using a Large Language Model Based on Group-Level Demographic Information},
  author    = {Sun, Seungjong and Lee, Eungu and Nan, Dongyan and Zhao, Xiangying and Lee, Wonbyung and Jansen, Bernard J. and Kim, Jang Hyun},
  journal   = {arXiv preprint arXiv:2402.18144},
  year      = {2024},
  url       = {https://arxiv.org/abs/2402.18144}
}

@book{malhotra2019marketing,
  title     = {Marketing Research: An Applied Orientation},
  author    = {Malhotra, Naresh K.},
  edition   = {7th},
  year      = {2019},
  publisher = {Pearson},
  address   = {Harlow, England},
  isbn      = {9781292265633}
}

@book{churchill2010marketing,
  title     = {Marketing Research: Methodological Foundations},
  author    = {Churchill, Gilbert A. and Iacobucci, Dawn},
  edition   = {10th},
  year      = {2010},
  publisher = {Cengage Learning},
  address   = {Mason, OH},
  isbn      = {9781439080672}
}

@book{dillman2014tailored,
  title     = {Internet, Phone, Mail, and Mixed-Mode Surveys: The Tailored Design Method},
  author    = {Dillman, Don A. and Smyth, Jolene D. and Christian, Leah Melani},
  edition   = {4th},
  year      = {2014},
  publisher = {Wiley},
  address   = {Hoboken, NJ},
  isbn      = {9781118456149}
}

@misc{wang2025largelanguagemodelsreplace,
      title={Large language models that replace human participants can harmfully misportray and flatten identity groups}, 
      author={Angelina Wang and Jamie Morgenstern and John P. Dickerson},
      year={2025},
      eprint={2402.01908},
      archivePrefix={arXiv},
      primaryClass={cs.CY},
      url={https://arxiv.org/abs/2402.01908}, 
}

@misc{giorgi2024modelinghumansubjectivityllms,
      title={Modeling Human Subjectivity in LLMs Using Explicit and Implicit Human Factors in Personas}, 
      author={Salvatore Giorgi and Tingting Liu and Ankit Aich and Kelsey Isman and Garrick Sherman and Zachary Fried and João Sedoc and Lyle H. Ungar and Brenda Curtis},
      year={2024},
      eprint={2406.14462},
      archivePrefix={arXiv},
      primaryClass={cs.CL},
      url={https://arxiv.org/abs/2406.14462}, 
}

@misc{neumann2025usellmssimulateopinions,
      title={Should you use LLMs to simulate opinions? Quality checks for early-stage deliberation}, 
      author={Terrence Neumann and Maria De-Arteaga and Sina Fazelpour},
      year={2025},
      eprint={2504.08954},
      archivePrefix={arXiv},
      primaryClass={cs.CY},
      url={https://arxiv.org/abs/2504.08954}, 
}

@misc{gss2022,
  title = {General Social Survey (GSS)},
  author = {{NORC at the University of Chicago}},
  year = {2022},
  howpublished = {\url{https://gss.norc.org}},
  note = {Accessed: 2025-05-15}
}

@misc{anes,
  title = {American National Election Studies (ANES)},
  author = {{American National Election Studies}},
  year = {2020},
  howpublished = {\url{https://electionstudies.org}},
  note = {Accessed: 2025-05-15}
}

@misc{gallup,
  title = {Gallup World Poll},
  author = {{Gallup Inc.}},
  year = {2022},
  howpublished = {\url{https://www.gallup.com/analytics}},
  note = {Accessed: 2025-05-15}
}

@article{lee2024can,
  title     = {Can large language models estimate public opinion about global warming? An empirical assessment of algorithmic fidelity and bias},
  author    = {Lee, Sanguk and Peng, Tai-Quan and Goldberg, Matthew H. and Rosenthal, Seth A. and Kotcher, John E. and Maibach, Edward W. and Leiserowitz, Anthony},
  journal   = {PLOS Climate},
  volume    = {3},
  number    = {8},
  pages     = {e0000429},
  year      = {2024},
  doi       = {10.1371/journal.pclm.0000429},
  url       = {https://journals.plos.org/climate/article?id=10.1371/journal.pclm.0000429},
  publisher = {Public Library of Science}
}

@article{sanders2023ai,
  title     = {Demonstrations of the Potential of AI-based Political Issue Polling},
  author    = {Sanders, Nathan E. and Ulinich, Alex and Schneier, Bruce},
  journal   = {Harvard Data Science Review},
  year      = {2023},
  volume    = {5},
  issue     = {4},
  doi       = {10.1162/99608f92.2d7c3c5b},
  url       = {https://hdsr.mitpress.mit.edu/pub/dm2hrtx0},
  publisher = {MIT Press}
}

@techreport{keeter2017low,
  author     = {Keeter, S. and Hatley, N. and Kennedy, C. and Lau, A.},
  title      = {What Low Response Rates Mean for Telephone Surveys},
  institution= {Pew Research Center},
  year       = {2017},
  url        = {https://www.pewresearch.org/methods/2017/05/15/what-low-response-rates-mean-for-telephone-surveys/},
}

@techreport{clinton2021taskforce,
  author     = {Clinton, J. D. and others},
  title      = {American Association of Public Opinion Research Task Force on Pre-Election Polling: An Evaluation of the 2020 General Election Polls},
  institution= {AAPOR},
  year       = {2021},
  url        = {https://aapor.org/wp-content/uploads/2022/11/Task-Force- on-2020-Pre-Election-Polling_Executive-Summary.pdf},
}

@article{kennedy2018evaluation,
  author    = {Kennedy, C. and others},
  title     = {An Evaluation of the 2016 Election Polls in the United States},
  journal   = {Public Opinion Quarterly},
  year      = {2018},
  volume    = {82},
  number    = {1},
  pages     = {1--33},
  doi       = {10.1093/poq/nfx047},
  url       = {https://academic.oup.com/poq/article-pdf/82/1/1/24265180/nfx047.pdf},
}

@misc{graham2023polling,
  author  = {Graham, D. A.},
  title   = {The Polling Crisis Is a Catastrophe for American Democracy},
  year    = {2023},
  note    = {The Atlantic},
  url     = {https://www.theatlantic.com/ideas/archive/2020/11/polling-catastrophe/616986/},
}

@article{waldner2018unwelcome,
  author    = {Waldner, D. and Lust, E.},
  title     = {Unwelcome Change: Coming to Terms with Democratic Backsliding},
  journal   = {Annual Review of Political Science},
  year      = {2018},
  volume    = {21},
  pages     = {93--113},
}

@misc{syntheticuser2023,
  title        = {Synthetic User},
  author       = {{SyntheticUser Team}},
  year         = {2023},
  note         = {Available at: \url{https://syntheticuser.com}},
  howpublished = {\url{https://syntheticuser.com}},
}

@misc{opinioai2023,
  title        = {OpinioAI},
  author       = {{OpinioAI Team}},
  year         = {2023},
  note         = {Available at: \url{https://www.opinioai.com}},
  howpublished = {\url{https://www.opinioai.com}},
}

@misc{delveai2023,
  title        = {Delve.ai},
  author       = {{Delve.ai Team}},
  year         = {2023},
  note         = {AI-Powered Customer Intelligence Platform. Available at: \url{https://delve.ai}},
  howpublished = {\url{https://delve.ai}},
}

@misc{personalive2024,
  title        = {PersonaLive.ai},
  author       = {{PersonaLive.ai Team}},
  year         = {2024},
  note         = {AI-driven psychometric and professional profiling and converstation platform. Available at: \url{https://www.personalive.ai}},
  howpublished = {\url{https://www.personalive.ai}},
}

@inproceedings{santurkar2023whose,
  title={Whose opinions do language models reflect?},
  author={Santurkar, Shibani and Durmus, Esin and Ladhak, Faisal and Lee, Cinoo and Liang, Percy and Hashimoto, Tatsunori},
  booktitle={International Conference on Machine Learning},
  pages={29971--30004},
  year={2023},
  organization={PMLR}
}

@article{argyle2023out,
  title={Out of one, many: Using language models to simulate human samples},
  author={Argyle, Lisa P and Busby, Ethan C and Fulda, Nancy and Gubler, Joshua R and Rytting, Christopher and Wingate, David},
  journal={Political Analysis},
  volume={31},
  number={3},
  pages={337--351},
  year={2023},
  publisher={Cambridge University Press}
}

@article{demszky2023using,
  title={Using large language models in psychology},
  author={Demszky, Dorottya and Yang, Diyi and Yeager, David S and Bryan, Christopher J and Clapper, Margarett and Chandhok, Susannah and Eichstaedt, Johannes C and Hecht, Cameron and Jamieson, Jeremy and Johnson, Meghann and others},
  journal={Nature Reviews Psychology},
  volume={2},
  number={11},
  pages={688--701},
  year={2023},
  publisher={Nature Publishing Group US New York}
}

@article{hu2024psycollm,
  title={PsycoLLM: Enhancing LLM for Psychological Understanding and Evaluation},
  author={Hu, Jinpeng and Dong, Tengteng and Ma, Hui and Zou, Peng and Sun, Xiao and Wang, Meng},
  journal={arXiv preprint arXiv:2407.05721},
  year={2024}
}

@article{jiang2023social,
  title={Social-LLM: Modeling User Behavior at Scale using Language Models and Social Network Data},
  author={Jiang, Julie and Ferrara, Emilio},
  journal={arXiv preprint arXiv:2401.00893},
  year={2023}
}

@inproceedings{hamalainen2023evaluating,
  title={Evaluating Large Language Models in Generating Synthetic HCI Research Data: a Case Study},
  author={H{\"a}m{\"a}l{\"a}inen, Perttu and Tavast, Mikke and Kunnari, Anton},
  booktitle={CHI '23: Proceedings of the 2023 CHI Conference on Human Factors in Computing Systems},
  pages={Article 433},
  year={2023},
  publisher={Association for Computing Machinery},
  doi={10.1145/3544548.3580688},
  url={https://dl.acm.org/doi/10.1145/3544548.3580688}
}

@inproceedings{prpa2024challenges,
  title={Challenges and Opportunities of LLM-Based Synthetic Personae and Data in HCI},
  author={Prpa, Mirjana and Troiano, Giovanni Maria and Wood, Matthew and Coady, Yvonne},
  booktitle={Proceedings of the CHI Conference on Human Factors in Computing Systems},
  year={2024},
  publisher={Association for Computing Machinery},
  doi={10.1145/3613905.3636293},
  url={https://dl.acm.org/doi/10.1145/3613905.3636293}
}

@article{hao2025multi,
  title={A Multi-LLM-Agent-Based Framework for Economic and Public Policy Analysis},
  author={Hao, Yuzhi and Xie, Danyang},
  journal={arXiv preprint arXiv:2502.16879},
  year={2025},
  url={https://arxiv.org/abs/2502.16879}
}

@article{zhang2024simulating,
  title={Simulating classroom education with llm-empowered agents},
  author={Zhang, Zheyuan and Zhang-Li, Daniel and Yu, Jifan and Gong, Linlu and Zhou, Jinchang and Liu, Zhiyuan and Hou, Lei and Li, Juanzi},
  journal={arXiv preprint arXiv:2406.19226},
  year={2024}
}

@article{ravi2023large,
  title={Large language models and medical education: Preparing for a rapid transformation in how trainees will learn to be doctors},
  author={Ravi, Akshay and Neinstein, Aaron and Murray, Sara G},
  journal={ATS scholar},
  volume={4},
  number={3},
  pages={282--292},
  year={2023},
  publisher={American Thoracic Society}
}

@article{louie2024roleplay,
  title={Roleplay-doh: Enabling Domain-Experts to Create LLM-simulated Patients via Eliciting and Adhering to Principles},
  author={Louie, Ryan and Nandi, Ananjan and Fang, William and Chang, Cheng and Brunskill, Emma and Yang, Diyi},
  journal={arXiv preprint arXiv:2407.00870},
  year={2024}
}

@article{kapania2024simulacrum,
  title={'Simulacrum of Stories': Examining Large Language Models as Qualitative Research Participants},
  author={Kapania, Shivani and Agnew, William and Eslami, Motahhare and Heidari, Hoda and Fox, Sarah},
  journal={arXiv preprint arXiv:2409.19430},
  year={2024}
}

@misc{pewresearch2024,
  author = "{Pew Research Center}",
  title = "Pew Research Center",
  year = "2024",
  url = "https://www.pewresearch.org/example-report",
  note = "Accessed: February 10, 2025"
}
\appendix
% \newpage
\clearpage
\appendix
\onecolumn

\section{Theoretical Explanation and Empirical Validation of the Shared-Subspace Assumption}
\label{app:shared_subspace_assumption}

This appendix provides both theoretical explanation and empirical validation for the shared-subspace assumption underlying APLaud. We first explain APLaud as a practical shared-subspace estimator for personalized adaptation, and then empirically validate whether the learned shared subspace captures task-relevant semantic structure.

\subsection{Theoretical Explanation}

\noindent
\textbf{APLaud as shared-subspace personalization.}  A useful way to interpret APLaud is through the lens of multilinear algebra, specifically a Tucker-2 style factorization of user-specific adaptation operators. 
Let \(\{\Delta W_u\}_{u=1}^n\) denote the personalized adaptation operators for \(n\) users, where each \(\Delta W_u \in \mathbb{R}^{d \times k}\). Stacking these user-specific updates along the user dimension forms a third-order tensor
\[
\mathcal{W} \in \mathbb{R}^{d \times k \times n},
\]
whose three modes correspond to the output dimension, input dimension, and user index, respectively. A natural way to model this collection of user-specific operators is through a Tucker-2-style factorization~\cite{tucker1966some,kolda2009tensor}:
\[
\Delta W_u \approx U C_u V^\top,
\]
where \(U \in \mathbb{R}^{d \times r}\) and \(V \in \mathbb{R}^{k \times r}\) are shared latent bases, and \(C_u \in \mathbb{R}^{r \times r}\) is a user-specific core matrix. Under this view, \(U\) and \(V\) define a shared low-dimensional adaptation subspace, while \(C_u\) specifies how user \(u\) modulates and combines these shared directions.

APLaud instantiates this structure in a practical way. Rather than explicitly fitting a full tensor decomposition over all user-specific updates, APLaud first learns a pooled LoRA update \(AB\) and then applies singular value decomposition:
\[
AB = U \Sigma V^\top.
\]
The resulting \(U\) and \(V\) estimate the dominant population-level row and column subspaces induced by the task. Personalization is then reduced to learning compact user-specific parameters within this shared basis:
\[
\Delta W_u
=
U(\Sigma + C_u)V^\top
+
\alpha_u \beta_u^\top.
\]
Here, \(U(\Sigma + C_u)V^\top\) captures user variation within the dominant shared adaptation subspace, while the rank-one residual \(\alpha_u \beta_u^\top\) captures idiosyncratic user behavior not well represented by the shared factors.

This perspective explains both the role of SVD and why APLaud is more data-efficient than per-user LoRA: instead of learning a full user-specific operator from sparse data, APLaud learns only a compact user-specific core \(C_u\) within a globally identified semantic subspace, plus a lightweight residual. Thus, APLaud is not merely ``SVD after LoRA,'' but a practical shared-subspace estimator for personalized adaptation.

\medskip
\noindent
\textbf{Meaning of \(U,V\) and validity of the shared-subspace assumption.}
The shared factors \(U\) and \(V\) have a natural interpretation in the adaptation process. The matrix \(V\) represents shared input-side semantic directions identified from the aggregated LoRA update, while \(U\) represents the corresponding output-side adaptation directions. The user-specific matrix \(C_u\) personalizes the model by reweighting or mixing these shared directions. Hence, APLaud assumes that users differ primarily in how they modulate common task-relevant directions, rather than requiring a completely new adaptation subspace for each user.

The shared-subspace assumption is most appropriate when users operate over a common task structure and personalization mainly reflects different mixtures or magnitudes of shared semantic directions. Survey response prediction is a particularly strong instance of this regime, since all users respond to the same or closely related questions, making the dominant adaptation directions naturally aligned across users.

The assumption may become weaker when personalization requires genuinely new directions outside the span of the aggregated update, i.e., when user-specific behavior is not well captured by the shared basis \(U,V\). In such cases, a fully independent per-user adapter may be more flexible. APLaud mitigates this limitation through the residual term \(\alpha_u \beta_u^\top\), which provides an additional path for capturing these idiosyncratic components when they arise.

\subsection{Empirical Validation}

To empirically examine whether such shared directions are meaningful, we conduct a \(V\)-space probing experiment across all five datasets. For each LoRA layer \(l\) and projection \(p\), we compute
\[
A_{l,p}B_{l,p}
=
U_{l,p}\Sigma_{l,p}V_{l,p}^\top.
\]
We then project question representations, without user profiles, onto the learned \(V\)-subspace and compare \textbf{V+LoRA} against \textbf{V+Base}. In \textbf{V+LoRA}, LoRA is enabled when extracting hidden representations. In \textbf{V+Base}, LoRA is disabled, while the same \(V\)-subspace is used with base-model hidden states.

To quantify semantic structure, we compare the representation distance with and without LoRA, with the clustering metric as
$$
\mathrm{Gap} = \overline{\cos}(\text{intra-topic}) - \overline{\cos}(\text{inter-topic}),
$$
where $\overline{\cos}$ denotes the mean pairwise cosine similarity computed over all question pairs within (or across) topic groups. We further define
$$
\Delta = \mathrm{Gap}_{\mathrm{LoRA}} - \mathrm{Gap}_{\mathrm{Base}},
$$
so $\Delta > 0$ indicates that stage-1 injects task-relevant semantic structure into the learned $V$-subspace

\begin{table}[t]
\centering
\caption{\(V\)-space probing results across five datasets. Positive \(\Delta\) indicates that the learned \(V\)-subspace captures stronger task-relevant semantic structure after LoRA adaptation.}
\label{tab:v_space_probing}
\small
\begin{tabular}{lcccccc}
\toprule
Dataset & \#Qs & \#Topics & V+LoRA Gap & V+Base Gap & \(\Delta\) & \(p\)-value \\
\midrule
G\&L & 51 & 5 & 0.0168 & 0.0124 & +0.0044 & 1.60e-51 \\
TS   & 63 & 6 & 0.0032 & 0.0010 & +0.0022 & 4.79e-06 \\
F\&R & 34 & 6 & 0.0240 & 0.0138 & +0.0101 & 3.54e-23 \\
EI   & 37 & 6 & 0.0265 & 0.0192 & +0.0073 & 5.73e-35 \\
GSS  & 18 & 5 & 0.0392 & 0.0253 & +0.0138 & 1.14e-11 \\
\bottomrule
\end{tabular}
\end{table}

As shown in Table~\ref{tab:v_space_probing}, the learned \(V\)-subspace is not an arbitrary algebraic artifact, but captures task-relevant semantic structure shared across questions by pushing apart unrelated topics and clustering related topics. This, in turn, supports the broader APLaud assumption that the dominant adaptation geometry for a task can be shared across users, while personalization is expressed through user-specific modulation within that shared geometry.

The results show that \(\Delta > 0\) on all five datasets, indicating that the learned \(V\)-subspace consistently captures task-relevant semantic structure introduced by LoRA training. For example, on TS, the base model shows near-zero topic discrimination (\(\mathrm{Gap}=0.0010\)), while LoRA makes the learned \(V\)-subspace more subtopic-discriminative. This provides empirical support for the interpretation in Theoretical Explanation that the shared subspace captures common task-adaptation directions, while personalization through \(C_u\) modulates how strongly each user responds along these directions.

At the same time, we do not claim that this assumption is universal. As discussed in the theoretical explanation, it is most appropriate when users share a common task geometry and differ mainly in how they weight shared semantic directions. It may weaken when personalization requires genuinely new directions outside the span of the aggregated update; in such cases, APLaud's residual term \(\alpha_u \beta_u^\top\) provides an additional path for capturing such idiosyncratic behavior.

Overall, both the theoretical explanation and empirical validation demonstrate that the shared-subspace assumption provides a meaningful and effective foundation for APLaud. Theoretically, APLaud decomposes personalized adaptation into shared task-relevant directions and compact user-specific modulation. Empirically, the \(V\)-space probing results show that the learned subspaces do capture the semantic structure shared across questions within the tasks, supporting the view that dominant adaptation geometry can be shared across users while preserving flexibility through lightweight user-specific components.

\section{Experiment Setting}
\label{appendix: exp settings}
\subsection{Environmental Setting}
All experiments were conducted on a single cluster node equipped with a Dell PowerEdge C6620 and NVIDIA H100 GPUs with 94 GB of memory.

\subsection{Data Processing}
\label{appx: data processing}
From ATP dataset, we selected four specific waves that cover a diverse range of public opinion topics, including gender and leadership, trust in science, family and relationships, and economic inequality. For the GSS, we focus on Panel 20, a longitudinal cohort that provides rich repeated measurement data on topics such as political trust, social norms, religiosity, and inequality. 

To explore user-level personalization and simulate LLM ``ownership'', we focus on the most engaged participants. Specifically, \textcolor{black}{we retain respondents with at least 10 valid answers and remove “Refused” responses. Since each wave contains over 100 ASK-ALL items, the filtered dataset provides roughly 50 answered questions per person on average—sufficient for robust personalization and learning stable user embeddings.  After this filtering,} we selected 200 users with the highest response rates and validated their responses across approximately 130 survey questions. To align the model output with the preferences and behavioral tendencies of the individual user, we first identify a subset comprising $30\%$ of questions that capture key aspects of user personality and behavioral traits to construct a user-specific profile through LLM-based prompting. Then we split the rest of the question into three sections, $80\%$ for training, $10\%$ for validation, $10\%$  for testing. This setup enables evaluation of each model’s ability to simulate personalized responses with minimal supervision. For LAMP Movie Tagging dataset, we follow the work of OPPU to choose the top 100 users and also split the data with 8:1:1 ratio. 
% In addition, we want to clarify that APLaud is not intended to be a universal solution for all personalization scenarios, but rather a method tailored to a specific settings where (i) users share a common task structure and (ii) per-user data are limited. However, it can still be used to achieve personalization on other tasks. To support this, we also empirically evaluate our methods in text generation task where we include results on the LaMP~\cite{salemi2023lamp} and LongLaMP datasets~\cite{kumar2024longlamp}. We include more results on text generation tasks in the appendix ~\ref{appdix:generation_task}.

\subsection{\textcolor{black}{Data Statistics}}\label{appdix: data statistics}
In this sections, we \textcolor{black}{first} provide all six data statistics we used in this paper in Table~\ref{tab:data_stats}
\begin{table*}[t]
\centering
\caption{Dataset statistics across six benchmarks. ``\# Qs'' denotes the number of questions, and ``Avg Q length'' is measured in tokens.}\label{tab:data_stats}
\begin{tabular}{lcccccc}
\toprule
 & G\&L & TS & F\&R & EI & GSS & LAMP Movie Tagging \\
\midrule
\# Users       & top 200 & top 200 & top 200 & top 200 & top 200 & top 100 \\
\# Qs          & 6104 & 7582 & 9036 & 9596 & 6161 & 8860 \\
Avg Q length   & 1343.2 & 1678.1 & 1384.5 & 1597.7 & 1195.6 & 572.3 \\
\bottomrule
\end{tabular}
\end{table*}

\textcolor{black}{We also disclose more details on the missing/incompleteness data statistics and analysis: Across the four ATP waves used in our study, the total number of survey items varies by design (W36: 139 questions, W42: 129, W50: 127, W54: 115). Consistent with ATP’s rotating-module structure, raw item-level missingness ranges from 28\% to 62\%, reflecting that different sub-samples receive different topical modules rather than indicating data quality issues. After applying our quality-control filter (retaining respondents with at least 10 valid answers), missingness decreases in every wave (e.g., W36: 62.1\% → 56.1\%; W42: 41.4\% → 39.5\%; W50: 54.8\% → 34.4\%; W54: 28.4\% → 24.4\%).}

\subsection{Training Details}
For a fair comparison, all experiments were trained for $5$ epochs.  In the Table\ref{tab:main_results_llam2}, we set the LoRA rank and dimension of $C_u$ to be 64. Subsequently, we selected the top 16 SVD dimension as the starting state for the second training phase, and set the residual module $A_u$ and $B_u$ to rank 1, respectively. Since the initialization of $m$ significantly influence the final result, we tuned its initialization from $\{50.0, 30.0, 20.0, 10.0,5.0, 2.0, 1.0, 0.1, 0.01, 0.01\}$ and reported the best result. These choices are based on our ablation study, where we explored the LoRA rank from $\{8, 16, 32, 64, 128\}$, SVD dimension from $\{4,8,16,32\}$, training epochs from $\{5,7,9,10\}$ and residual term rank from $\{1,2,4,8\}$. We found that reducing the rank of the pretrained LoRA may boosted performance, while changes in other hyperparameters had less impact.

\section{Ablation Study}
\begin{figure}[t]
  \centering
  \includegraphics[width=0.9\linewidth]{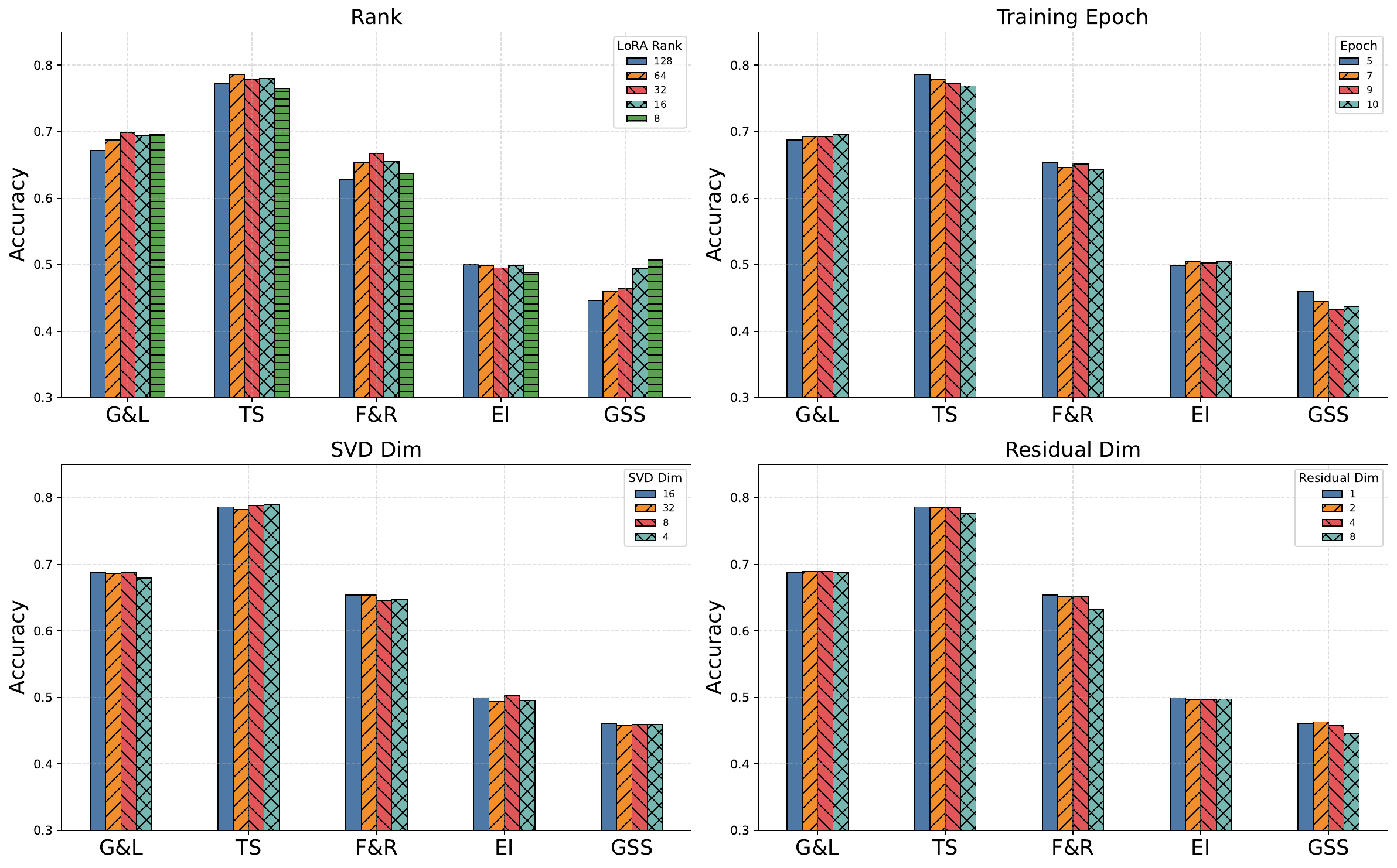}
  \caption{Ablation study on different hyperparameters.}
  \label{fig:ablation}
\end{figure}

In this section, we systematically explore the effect of 4 hyperparameters: LoRA Rank, SVD dim (i.e., rank of $P_u$ and $Q_u$), number of training epochs, and residual dimension (i.e., rank of $A_u$ and $B_u$) on our framework's performance. This analysis helps identify the optimal range for each setting and provides insight into the robustness of our approach. We conduct ablation study on our textbf{Aplaud+} model. The results of our ablation study are presented in Table \ref{table: ablation-studies} and Fig \ref{fig:ablation}. Specifically, in Table \ref{fig:ablation} (a), we fix the training epoch at 5 and set the residual dimension to 1. Given that the SVD dimension must remain smaller than $C_u$ (i.e., the LoRA rank), we experiment with the following (LoRA rank, SVD dim) pairs: (128, 16), (64, 16), (32, 16), (16, 8), and (8, 4). In Table \ref{table: ablation-studies} (b), we set LoRA rank to be 64, SVD dim 16, residual dimension to be 1 and experiment on training epochs from $\{ 5,7,9, 10\}$. In Table \ref{table: ablation-studies} (c), we set LoRA rank to be 64, training epoch to be 5, residual dimension to be 1
and experiment on SVD dimension from $\{32, 16, 8, 4\}$. In Table \ref{table: ablation-studies} (d), we set LoRA rank to be 64, training epoch to be 5, SVD dimension to be 16 and experiment on different residual dimension from $\{1,2,4,8\}$.

\begin{table*}[htbp]
\scriptsize
\centering
\caption{Ablation studies across four key hyperparameters. All reported values are test accuracy. G\&L = Gender \& Law, TS = \textcolor{black}{Twitter Stance}, F\&R = Finance \& Risk, EI = Emotional Intensity, GSS = General Social Survey.}
\label{table: ablation-studies}
\vspace{0.5em}
\begin{minipage}{0.48\textwidth}
\centering
\textbf{(a) LoRA rank}\par\vspace{0.3em}

\begin{tabular}{c ccccc}
\toprule
Rank & G\&L & TS & F\&R & EI & GSS \\
\midrule
128 & 0.6715 & 0.7730 & 0.6278 & 0.5000 & 0.4463 \\
64  & 0.6876 & 0.7862 & 0.6537 & 0.4992 & 0.4605 \\
32  & 0.6989 & 0.7781 & 0.6667 & 0.4950 & 0.4647 \\
16  & 0.6940 & 0.7801 & 0.6550 & 0.4983 & 0.4944 \\
8   & 0.6957 & 0.7649 & 0.6368 & 0.4883 & 0.5071 \\
\bottomrule
\end{tabular}
\end{minipage}
\hfill
\begin{minipage}{0.48\textwidth}
\centering
\textbf{(b) Training epoch}
\vspace{0.3em}
\begin{tabular}{c ccccc}
\toprule
Epoch & G\&L & TS & F\&R & EI & GSS \\
\midrule
5  & 0.6876 & 0.7862 & 0.6537 & 0.4992 & 0.4605 \\
7  & 0.6924 & 0.7781 & 0.6459 & 0.5042 & 0.4449 \\
9  & 0.6924 & 0.7730 & 0.6515 & 0.5025 & 0.4322 \\
10 & 0.6957 & 0.7690 & 0.6433 & 0.5042 & 0.4364 \\
\bottomrule
\end{tabular}
\end{minipage}

\vspace{1em}

\begin{minipage}{0.48\textwidth}
\centering
\textbf{(c) SVD dimension}
\vspace{0.3em}
\begin{tabular}{c ccccc}
\toprule
SVD dim & G\&L & TS & F\&R & EI & GSS \\
\midrule
32 & 0.6860 & 0.7822 & 0.6537 & 0.4933 & 0.4576 \\
16 & 0.6876 & 0.7862 & 0.6537 & 0.4992 & 0.4605 \\
8  & 0.6876 & 0.7882 & 0.6459 & 0.5025 & 0.4590 \\
4  & 0.6795 & 0.7893 & 0.6472 & 0.4950 & 0.4590 \\
\bottomrule
\end{tabular}
\end{minipage}
\hfill
\begin{minipage}{0.48\textwidth}
\centering
\textbf{(d) Residual dimension}
\vspace{0.3em}
\begin{tabular}{c ccccc}
\toprule
Res dim & G\&L & TS & F\&R & EI & GSS \\
\midrule
1 & 0.6876 & 0.7862 & 0.6537 & 0.4992 & 0.4605 \\
2 & 0.6892 & 0.7852 & 0.6511 & 0.4967 & 0.4633 \\
4 & 0.6892 & 0.7852 & 0.6519 & 0.4967 & 0.4576 \\
8 & 0.6876 & 0.7761 & 0.6329 & 0.4975 & 0.4449 \\
\bottomrule
\end{tabular}
\end{minipage}

\vspace{0.5em}
% \noindent\scriptsize\textit{All reported values are test accuracy. G\&L = Gender \& Law, TS = Twitter Stance, F\&R = Finance \& Risk, EI = Emotional Intensity, GSS = General Social Survey.}
\end{table*}

\subsection{LoRA Rank}
We observe that the impact of the pretrained LoRA rank on model performance is not consistent across datasets. For example, while accuracy on \textsc{G\&L} and \textsc{F\&R} improves as the rank decreases, the performance on \textsc{TS} and \textsc{EI} slightly drops when the rank is reduced to 8. This inconsistency suggests that the optimal LoRA rank may vary depending on the dataset characteristics. To ensure a fair comparison under a uniform setting, we additionally conduct experiments using pretrained LoRA with a fixed rank of 8 for all models. The results, summarized in Table~\ref{table: rank8 res}, show that our framework continues to outperform all baselines under this constrained low-rank setting.
\subsection{Training Epoch}
We evaluate the effect of training epochs by varying the number of fine-tuning epochs from 5 to 10. Our observations indicate that extending training beyond 5 epochs does not consistently improve performance. For example, on \textsc{TS}, accuracy decreased from 0.7862 (epoch 5) to 0.7690 (epoch 10), and on \textsc{F\&R}, it dropped from 0.6537 to 0.6433. This suggests that the model begins to overfit the limited personalized data after a few epochs. Overall, 5 epochs already yield strong performance, with additional training offering diminishing or even negative returns, thus emphasizing the importance of early stopping or lightweight adaptation.

\subsection{SVD Dimension}
To assess the role of the low-rank SVD projection dimension, we compared the SVD dimension (rank of $P_u$ and $Q_u$) of 4, 8, 16, and 32. The best overall performance was achieved using 16 dimensions, which served as our default setting. While increasing the SVD dimension to 32 maintained similar performance on \textsc{F\&R} and \textsc{G\&L}, it slightly degraded on \textsc{EI} and \textsc{GSS}. Conversely, reducing the dimension to 4 caused a more noticeable drop in performance, particularly on \textsc{G\&L} (0.6795). These trends indicate that although the method is robust to reasonable SVD compression, extremely low dimensions may under-represent user-specific variation.
\subsection{Residual Dimension}
We varied the dimension of the user-specific residual component (residual dimension, the rank of $A_u$ and $B_u$) from 1 to 8 to assess its effect. Performance remained relatively stable when increasing from 1 to 4 dimensions, showing minor fluctuations across datasets. For example, on \textsc{TS}, performance was nearly unchanged between 1 and 4 dimensions (0.7862 vs. 0.7852), suggesting that even a very low-rank residual term is sufficient to capture key personalization signals. However, setting the residual dimension to 8 lead to a performance drop on several datasets (e.g., \textsc{G\&L} and \textsc{F\&R}), likely due to overfitting. This result highlights the effectiveness of an extreme small residual dimension.

\subsection{Sparse-User Behavior}
\label{app:sparse_user_ablation}

To study sparse-user behavior, we conduct an ablation study on G\&L and TS by varying the amount of per-user training data from \(100\%\) to \(10\%\). The results are shown in Table~\ref{tab:sparse_user_ablation}.

\begin{table}[t]
\centering
\caption{Ablation study on sparse-user behavior. We vary the amount of per-user training data from \(100\%\) to \(10\%\) on G\&L and TS.}
\label{tab:sparse_user_ablation}
\small
\setlength{\tabcolsep}{5pt}
\begin{tabular}{lcccc}
\toprule
\multirow{2}{*}{Method / Ratio}
& \multicolumn{2}{c}{G\&L}
& \multicolumn{2}{c}{TS} \\
\cmidrule(lr){2-3}
\cmidrule(lr){4-5}
& ACC & Macro-F1 & ACC & Macro-F1 \\
\midrule
LoRA & 0.6522 & 0.6418 & 0.7214 & 0.4529 \\
\midrule
\(100\%\) & 0.6844 & 0.6764 & 0.7943 & 0.7253 \\
\(90\%\)  & 0.6812 & 0.6727 & 0.7872 & 0.7217 \\
\(80\%\)  & 0.6810 & 0.6711 & 0.7741 & 0.6559 \\
\(70\%\)  & 0.6763 & 0.6682 & 0.7649 & 0.6297 \\
\(60\%\)  & 0.6779 & 0.6707 & 0.7356 & 0.5000 \\
\(50\%\)  & 0.6763 & 0.6661 & 0.7244 & 0.4975 \\
\(40\%\)  & 0.6683 & 0.6618 & 0.7315 & 0.5010 \\
\(30\%\)  & 0.6586 & 0.6523 & 0.7295 & 0.5057 \\
\(20\%\)  & 0.6506 & 0.6426 & 0.7224 & 0.5018 \\
\(10\%\)  & 0.6329 & 0.6279 & 0.6910 & 0.4994 \\
\bottomrule
\end{tabular}
\end{table}

The results show a generally smooth degradation as the amount of user-specific training data decreases. On G\&L, ACC drops from \(0.6844\) with full user data to \(0.6329\) with only \(10\%\) user data. On TS, ACC drops from \(0.7943\) to \(0.6910\). The degradation becomes more noticeable when the user data ratio falls below roughly \(30\%\), suggesting that the supervision signal becomes too weak for stable personalization.

In the extremely low-data regime, personalization may even underperform the non-personalized LoRA baseline. For example, when using only \(10\%\) of user data, the model obtains lower ACC than LoRA on both G\&L and TS. This suggests that when user-specific data is very scarce, training personalized parameters on only a few examples can introduce additional bias and randomness. However, this issue is alleviated as more user data becomes available: once the user data ratio increases, the personalized model quickly recovers and consistently outperforms the LoRA baseline. These results indicate that APLaud is robust under moderate data sparsity, while also highlighting the expected limitation of personalization methods in extremely sparse-user settings.
\section{Additional Experimental Results}
\subsection{Rank-8 pretrained LoRA based Experimental Result}
As ablation studies indicated that rank of pretrained LoRA may influence our three-phase training performance across different datasets, for a fair and more comprehensive comparison, we also report experimental results (including baseline results) based on rank-8 pretrained LoRA in Table 
\ref{table: rank8 res} for better comparison. The experiment setting is as follow: LoRA rank is set to be 8, training epoch is set to be 5, SVD dimension to be 4, rank of residual term ($A_u$ and $B_u$) to be 1, we tune the initlization of $m$ from $\{50.0, 30.0, 20.0, 10.0,5.0, 2.0, 1.0, 0.1, 0.01, 0.01\}$ and reported the best result. Compared with Table \ref{tab:main_results_llam2}, the following conclusions can be drawn:
\begin{itemize}
    \item While all models are affected by the choice of LoRA rank, APlaud tends to benefit more from lower ranks compared to baseline methods, which sometimes exhibit marginal or negative effects, particularly on datasets like GSS.
    \item Our APlaud framework outperforms baseline models across all datasets, with the largest improvements reaching up to $12.8\%$ in accuracy and $6.17\%$ in macro-F1.
    \item These results further highlight the robustness of our model, as it consistently outperforms baselines in personalization tasks regardless of the pre-trained LoRA setup, and does so with substantially fewer parameters.

\end{itemize}
  
\begin{table*}[htbp]
\scriptsize
\centering
\caption{Performance comparison across different datasets based on pretrained rank-8 LoRA. Bold numbers indicate the best results within each dataset. Dataset names: G\&L = ATP Gender \& Leadership, TS = ATP Trust in Science, F\&R = ATP Family and Relationships, EI = ATP Economic Inequality, GSS = General Social Survey.}\label{table: rank8 res}
\vspace{0.3em}

\resizebox{\textwidth}{!}{%
\begin{tabular}{l cc cc cc cc cc}
\toprule
\multirow{2}{*}{Method} 
& \multicolumn{2}{c}{G\&L} 
& \multicolumn{2}{c}{TS} 
& \multicolumn{2}{c}{F\&R} 
& \multicolumn{2}{c}{EI} 
& \multicolumn{2}{c}{GSS} \\
\cmidrule(lr){2-11}
 & ACC & F1 & ACC & F1 & ACC & F1 & ACC & F1 & ACC & F1 \\
\midrule
LoRA         & 0.6312 & 0.6209 & 0.7123 & 0.4216 & 0.5681 & 0.2655 & 0.4117 & 0.3219 & 0.4901 & 0.3165 \\
OPPU         & 0.6828 & 0.6761 & 0.7822 & 0.7004 & 0.6174 & 0.3569 & 0.4950 & 0.4103 & 0.5014 & 0.3349 \\
Cu           & 0.6908 & 0.6824 & 0.7903 & 0.6598 & 0.5863 & 0.3100 & 0.4925 & 0.3982 & 0.5042 & 0.3269 \\
%PuQu       & 0.6957 & 0.6870 & 0.7974 & 0.5893 & 0.5785 & 0.2989 & 0.4900 & 0.3966 & \textbf{0.5282} & \textbf{0.3504} \\
Aplaud+  & \textbf{0.7005} & \textbf{0.6921} & \textbf{0.7994} & \textbf{0.7436} & \textbf{0.6498} & \textbf{0.4028} & \textbf{0.5008} & \textbf{0.4161} & 0.5240 & 0.3484 \\
\bottomrule
\end{tabular}
}

\vspace{0.3em}
\end{table*}

\label{appendix:additional_results}

\subsection{\textcolor{black}{Performance Result without Profile Input}}

\begin{table*}[t]
\centering
\caption{\textcolor{black}{Performance comparison across different datasets with LLaMA-2 backbone with profile being removed from prompts. Dataset names: G\&L = ATP Gender \& Leadership, TS =
ATP Trust in Science, F\&R = ATP Family and Relationships, EI = ATP Economic Inequality, GSS =
General Social Survey.}}
\label{tab:noprofile}
\small
\setlength{\tabcolsep}{4pt}

\begin{tabular}{lcccccccccc}
\toprule
\textcolor{black}{Method} &
\multicolumn{2}{c}{\textcolor{black}{G\&L}} &
\multicolumn{2}{c}{\textcolor{black}{TS}} &
\multicolumn{2}{c}{\textcolor{black}{F\&R}} &
\multicolumn{2}{c}{\textcolor{black}{EI}} &
\multicolumn{2}{c}{\textcolor{black}{GSS}} \\
\cmidrule(lr){2-3}
\cmidrule(lr){4-5}
\cmidrule(lr){6-7}
\cmidrule(lr){8-9}
\cmidrule(lr){10-11}
& \textcolor{black}{ACC} & \textcolor{black}{Macro-F1}
& \textcolor{black}{ACC} & \textcolor{black}{Macro-F1}
& \textcolor{black}{ACC} & \textcolor{black}{Macro-F1}
& \textcolor{black}{ACC} & \textcolor{black}{Macro-F1}
& \textcolor{black}{ACC} & \textcolor{black}{Macro-F1} \\
\midrule

\textcolor{black}{LoRA} &
\textcolor{black}{0.4576} & \textcolor{black}{0.3106} &
\textcolor{black}{0.6454} & \textcolor{black}{0.3249} &
\textcolor{black}{0.5486} & \textcolor{black}{0.2598} &
\textcolor{black}{0.4725} & \textcolor{black}{0.3028} &
\textcolor{black}{0.3583} & \textcolor{black}{0.2139} \\

\textcolor{black}{OPPU} &
\textcolor{black}{0.6271} & \textcolor{black}{0.6030} &
\textcolor{black}{0.7528} & \textcolor{black}{0.6115} &
\textcolor{black}{0.5970} & \textcolor{black}{0.3462} &
\textcolor{black}{0.4942} & \textcolor{black}{\textbf{0.4005}} &
\textcolor{black}{0.3686} & \textcolor{black}{0.2322} \\

\textcolor{black}{Cu} &
\textcolor{black}{0.6151} & \textcolor{black}{0.5798} &
\textcolor{black}{0.6991} & \textcolor{black}{0.5838} &
\textcolor{black}{0.5681} & \textcolor{black}{0.2975} &
\textcolor{black}{0.4942} & \textcolor{black}{0.3602} &
\textcolor{black}{0.3743} & \textcolor{black}{0.2545} \\

\textcolor{black}{Aplaud} &
\textcolor{black}{\textbf{0.6457}} & \textcolor{black}{\textbf{0.6225}} &
\textcolor{black}{0.7599} & \textcolor{black}{0.6165} &
\textcolor{black}{0.6148} & \textcolor{black}{0.3498} &
\textcolor{black}{0.4942} & \textcolor{black}{0.3614} &
\textcolor{black}{0.3743} & \textcolor{black}{0.2604} \\

\textcolor{black}{Aplaud+} &
\textcolor{black}{0.6329} & \textcolor{black}{0.6079} &
\textcolor{black}{\textbf{0.7639}} & \textcolor{black}{\textbf{0.6231}} &
\textcolor{black}{\textbf{0.6200}} & \textcolor{black}{\textbf{0.3599}} &
\textcolor{black}{\textbf{0.5083}} & \textcolor{black}{0.3918} &
\textcolor{black}{\textbf{0.3757}} & \textcolor{black}{\textbf{0.2617}} \\

\bottomrule
\end{tabular}

\end{table*}

\textcolor{black}{
Our user profile is constructed from survey-provided demographic metadata and the 10 survey questions most relevant to user characterization. These questions are selected from the user’s answered items and capture personal attitudes, preferences, and values. To avoid information leakage, we remove these 10 profile-related questions before forming the train/validation/test split. See \ref{appdix: profile generation} on the data and prompt we use for profile generation.}

\textcolor{black}{To verify the importance of profile information and to evaluate each model’s ability to learn user preferences without explicit profile signals, we conduct an additional ablation in which all models are trained and evaluated without any profile input. As shown in Table~\ref{tab:noprofile}, removing profile information leads to a consistent performance drop across all methods, demonstrating that user profiles provide valuable preference cues. Nevertheless, Aplaud and Aplaud+ remain highly competitive and often surpass OPPU across multiple datasets, despite using significantly fewer per-user parameters. These results indicate that our lightweight modules can effectively recover stable preference patterns even in the absence of explicit profile features, highlighting the robustness and parameter efficiency of our approach.}

\subsection{\textcolor{black}{Additional Evaluation on Generation Task}}\label{appdix:generation_task}

% \begin{table}[t]
% \centering
% \caption{\textcolor{black}{Generation Task Performance on LaMP News Headline Dataset.}}
% \label{tab: lamp_news_headline}

% \vspace{0.2cm}

% \begin{tabular}{lcc}
% \Xhline{2\arrayrulewidth}
% \textcolor{black}{Method} & \textcolor{black}{R-1} & \textcolor{black}{R-L} \\
% \hline
% \textcolor{black}{GPT5-profile} & \textcolor{black}{0.1312} & \textcolor{black}{0.1177} \\
% \textcolor{black}{OPPU}         & \textcolor{black}{0.1987} & \textcolor{black}{0.1832} \\
% \textcolor{black}{Cu}           & \textcolor{black}{0.1987} & \textcolor{black}{0.1825} \\
% \textcolor{black}{Aplaud}       & \textcolor{black}{0.2003} & \textcolor{black}{0.1838} \\
% \textcolor{black}{Aplaud+}      & \textcolor{black}{0.1987} & \textcolor{black}{0.1830} \\
% \Xhline{2\arrayrulewidth}
% \end{tabular}

% \end{table}

\begin{table}[t]
\centering
\caption{Performance on Generation Task}
\label{tab: generation task}
\begin{tabular}{lcccccc}
\toprule
 & \multicolumn{2}{c}{Product Review} 
 & \multicolumn{2}{c}{Abstract Generation} 
 & \multicolumn{2}{c}{News Headline} \\
\cmidrule(lr){2-3} \cmidrule(lr){4-5} \cmidrule(lr){6-7}
Method & R-1 & R-L & R-1 & R-L & R-1 & R-L \\
\midrule
LoRA      & 0.3652 & 0.1820 & 0.3675 & 0.2131 & 0.1982 & 0.1815 \\
OPPU      & 0.3671 & 0.1855 & 0.3712 & 0.2048 & 0.1987 & 0.1830 \\
Cu        & 0.3729 & 0.1948 & 0.3712 & 0.2146 & 0.1987 & 0.1825 \\
Aplaud    & 0.3760 & \textbf{0.2142} & 0.3801 & 0.2109 & \textbf{0.2003} & \textbf{0.1838} \\
Aplaud+   & \textbf{0.3816} & 0.2133 
           & \textbf{0.3896} & \textbf{0.2189} 
           & 0.1987 & 0.1830 \\
\bottomrule
\end{tabular}
\end{table}

We further evaluate our approach on the text generation task to assess whether the proposed personalization mechanism also benefits a text generation setting. As shown in Table~\ref{tab: generation task}, APLaud consistently achieves comparable or better performance than OPPU while using only ~1\% of the trainable parameters, demonstrating that its efficiency–performance tradeoff extends beyond the survey setting.
In summary, while we APLaud is not universally optimal for all personalization problems, it is demonstrated that it extends beyond survey prediction to general text generation tasks, with peak advantages appearing in structured, shared-task settings.

\subsection{\textcolor{black}{Additional Result where stage 1 does not use train/test user}}

\begin{table*}[t]
\caption{\textcolor{black}{Performance comparison across different datasets with LLaMA-2 backbone where users in Stage~1 pretraining does not overlap with users in Stage~2}}
\label{tab: no overlap stage 1}
\centering
\small
\resizebox{\linewidth}{!}{
\begin{tabular}{lcccccccccc}
\toprule
\textcolor{black}{Method} &
\multicolumn{2}{c}{\textcolor{black}{G\&L}} &
\multicolumn{2}{c}{\textcolor{black}{TS}} &
\multicolumn{2}{c}{\textcolor{black}{F\&R}} &
\multicolumn{2}{c}{\textcolor{black}{EI}} &
\multicolumn{2}{c}{\textcolor{black}{GSS}} \\
\cmidrule(lr){2-3}
\cmidrule(lr){4-5}
\cmidrule(lr){6-7}
\cmidrule(lr){8-9}
\cmidrule(lr){10-11}
& \textcolor{black}{ACC} & \textcolor{black}{Macro-F1}
& \textcolor{black}{ACC} & \textcolor{black}{Macro-F1}
& \textcolor{black}{ACC} & \textcolor{black}{Macro-F1}
& \textcolor{black}{ACC} & \textcolor{black}{Macro-F1}
& \textcolor{black}{ACC} & \textcolor{black}{Macro-F1} \\
\midrule

\textcolor{black}{LoRA} &
\textcolor{black}{0.5749} & \textcolor{black}{0.5344} &
\textcolor{black}{0.6667} & \textcolor{black}{0.3736} &
\textcolor{black}{0.4929} & \textcolor{black}{0.2416} &
\textcolor{black}{0.4642} & \textcolor{black}{0.3687} &
\textcolor{black}{0.3136} & \textcolor{black}{0.1707} \\

\textcolor{black}{OPPU} &
\textcolor{black}{0.6667} & \textcolor{black}{0.6529} &
\textcolor{black}{0.7305} & \textcolor{black}{0.5995} &
\textcolor{black}{0.6135} & \textcolor{black}{0.3471} &
\textcolor{black}{0.5042} & \textcolor{black}{0.3951} &
\textcolor{black}{0.3771} & \textcolor{black}{0.2909} \\

\textcolor{black}{Cu} &
\textcolor{black}{0.6441} & \textcolor{black}{0.6425} &
\textcolor{black}{0.7305} & \textcolor{black}{0.5987} &
\textcolor{black}{0.6196} & \textcolor{black}{0.3440} &
\textcolor{black}{0.5208} & \textcolor{black}{0.3955} &
\textcolor{black}{0.3775} & \textcolor{black}{0.2943} \\

\textcolor{black}{Aplaud} &
\textcolor{black}{0.6506} & \textcolor{black}{0.6506} &
\textcolor{black}{0.7599} & \textcolor{black}{0.6335} &
\textcolor{black}{0.6265} & \textcolor{black}{0.3695} &
\textcolor{black}{0.5167} & \textcolor{black}{0.3927} &
\textcolor{black}{\textbf{0.4124}} & \textcolor{black}{\textbf{0.3259}} \\

\textcolor{black}{Aplaud+} &
\textcolor{black}{\textbf{0.6860}} & \textcolor{black}{\textbf{0.6717}} &
\textcolor{black}{\textbf{0.7639}} & \textcolor{black}{\textbf{0.6401}} &
\textcolor{black}{\textbf{0.6291}} & \textcolor{black}{\textbf{0.3706}} &
\textcolor{black}{\textbf{0.5208}} & \textcolor{black}{\textbf{0.3981}} &
\textcolor{black}{0.3969} & \textcolor{black}{0.3147} \\
\bottomrule
\end{tabular}
}
\end{table*}

\textcolor{black}{To ensure that Stage~1 pretraining does not unintentionally encode information about users who later appear in personalization, we conduct an additional experiment in which Stage~1 is trained on a 20\% subsample of users that does not overlap with any train/test users. This setup guarantees that none of the evaluation users contribute to Stage~1 and thus eliminates any possibility of user-level information leaking across stages. As shown in Table~\ref{tab: no overlap stage 1}, the performance of Aplaud and Aplaud+ remains virtually unchanged, confirming that Stage~1 captures only \textit{general task knowledge}, while Stage~2 is solely responsible for learning user-specific preferences. This further demonstrates that task-level learning and personalization are cleanly disentangled in our framework.}

\subsection{Pratical Running Time and Memory Usage}
In this section, we show practical running time and memory in table~\ref{tab: practical running time and mem}

\begin{table*}[t]
\centering
\caption{Pratical Running Time and Memory Usage}\label{tab: practical running time and mem}
\begin{tabular}{lcccc}
\toprule
Method & SVD Time & Time & GPU Mem & \# Per User Param \\
\midrule
LoRA            & –     & 29 min  & 49.18 GB & 100\%  \\
OPPU            & –     & 72 min  & 53.96 GB & 100\%  \\
APlaud+:Stage1      & 17 min & 38 min  & 67.09 GB & 0.54\% \\
APlaud+:Stage2(a)    & –     & 33 min  & 62.04 GB & 0.27\% \\
APlaud+:Stage2(b)   & 46 s  & 34 min  & 61.12 GB & 1.84\% \\
APlaud+:Stage2(c)   & 18 min & 105 min & 67.09 GB & 1.84\% \\
\bottomrule
\end{tabular}
\end{table*}

\subsection{Sensitivity analysis}
We conducted sensitivity analyses to assess the robustness of APlaud to imperfect initializations of the shared subspace and to examine potential error propagation during subsequent training stages.

To simulate noisy conditions, we added Gaussian noise \( \epsilon \cdot \mathcal{N}(0,1) \) to the global LoRA matrices \( A \) and \( B \) prior to performing SVD:  
\[
(U, \Sigma, V) = \mathrm{SVD}(AB + \epsilon \cdot \mathcal{N}(0,1)).
\]
This perturbation introduces stochasticity into the shared subspace and allows us to observe the impact of initialization noise on downstream personalization.

We specifically use \( \epsilon = 10^{-3} \) and \( 10^{-4} \), which introduce non-trivial yet controlled noise magnitudes. These values are chosen to reflect realistic perturbations relative to the typical scale of LoRA updates, where \( AB \) often has entries on the order of \( 10^{-2} \) or smaller. 

As shown in our experimental results (Table \ref{tab:sensitivity analysis}), APlaud demonstrates strong robustness to such perturbations, consistently exhibiting low performance variance across runs. This suggests that APlaud does not rely heavily on precise early-stage decompositions and can generalize effectively even in the presence of moderate noise during shared initialization.

\begin{table}[t]
\centering
\caption{Robustness study under random noise injection. We report mean $\pm$ std for ACC and Macro-F1.}
\label{tab:sensitivity analysis}
\begin{tabular}{lcc}
\toprule
Method & ACC & Macro-F1 \\
\midrule
Cu              & 0.6626 $\pm$ 0.0031 & 0.6455 $\pm$ 0.0070 \\
PuQu          & 0.6618 $\pm$ 0.0039 & 0.6411 $\pm$ 0.0038 \\
Aplaud+    & \textbf{0.6856 $\pm$ 0.0051} & \textbf{0.6686 $\pm$ 0.0062} \\
\midrule
Cu              & 0.6642 $\pm$ 0.0009 & 0.6435 $\pm$ 0.0011 \\
PuQu          & 0.6626 $\pm$ 0.0059 & 0.6424 $\pm$ 0.0058 \\
Aplaud+    & \textbf{0.6852 $\pm$ 0.0016} & \textbf{0.6677 $\pm$ 0.0020} \\
\midrule
Cu              & 0.6611 $\pm$ 0.0060 & 0.6401 $\pm$ 0.0058 \\
PuQu          & 0.6654 $\pm$ 0.0038 & 0.6432 $\pm$ 0.0048 \\
Aplaud+    & \textbf{0.6755 $\pm$ 0.0093} & \textbf{0.6593 $\pm$ 0.0050} \\
\bottomrule
\end{tabular}
\end{table}

\subsection{\textcolor{black}{Shared Subapace Similarity Analysis}}

\begin{figure}
    \centering
    \includegraphics[width=1.0\linewidth]{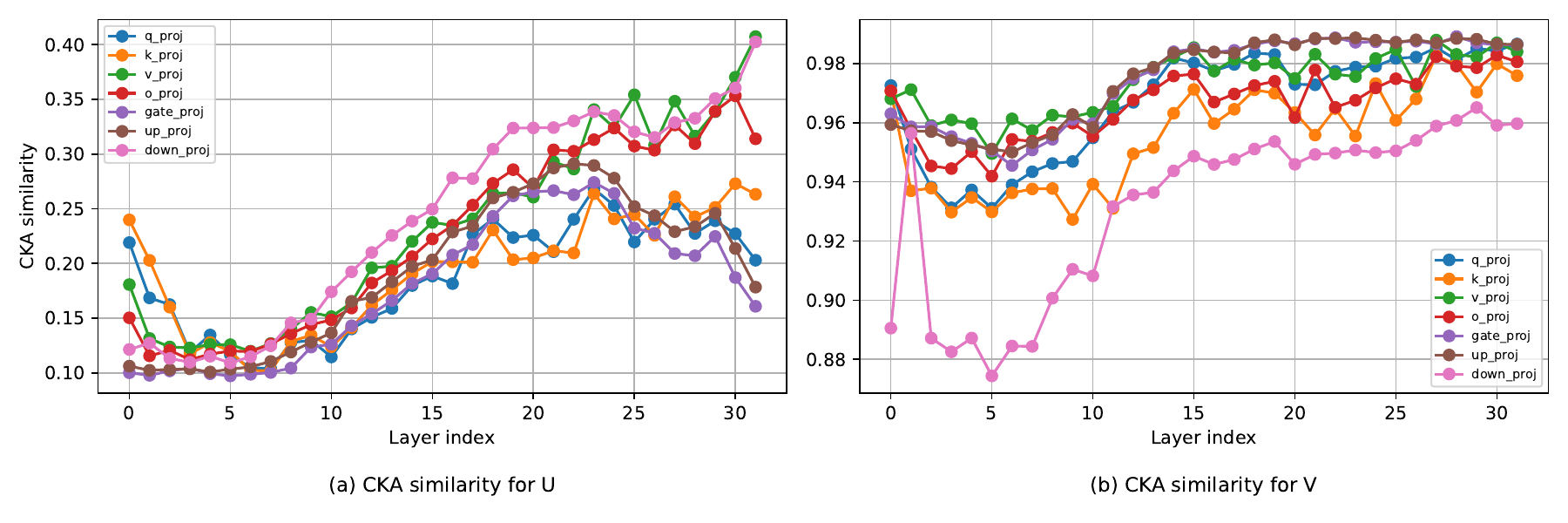}
    \caption{Average CKA similarity of U, V under different layers across different topic}
    \label{fig:cka similairy}
\end{figure}

\textcolor{black}{In this section, we analyze the similarity of different shared subpace $U$ and $V$ across different topics. Specifically,  we computed average (across different modules within the same layer) \emph{Centered Kernel Alignment} (CKA) ~\cite{kornblith2019similarity} similarities of the SVD components $U$ and $V$ between two different topics (G\&L vs.\ TS) as shown in Fig~\ref{fig:cka similairy}:
(i) $V$ is relatively stable across topics, while
(ii) $U$ varies more.
This confirms that $U$ and $V$ serve different purposes: $V$ acts as a coordinate generator which could be relevant across different topics, whereas $U$ serves as different semantic subspaces built on top of $V$. Despite this, both remain well-aligned across topics, supporting our shared-subspace design.}

\textcolor{black}{We would like to mention again that our method does not assume that a single pair of $(U,V)$ must generalize across all tasks or topics. In practice, the model learns different $(U,V)$ for different tasks/topics, as also supported by our CKA analysis. The personalized parameters $(C_u, \alpha_u, \beta_u)$ are then learned in Stage~2 within the subspace defined by that task/topic-specific $(U,V)$. This design ensures that personalization is performed inside an already aligned subspace, without requiring cross-topic invariance of $(U,V)$.}

\textcolor{black}{Regarding drift across waves and time, our current datasets mix users from different time points within each topic, so we cannot cleanly isolate purely temporal drift in this paper. We view a more fine-grained, time-indexed analysis as important future work. However, in real survey practice, questions within the same topic typically evolve slowly and remain within a relatively narrow semantic range. Combined with the high CKA stability of $V$ that we observe, this gives us good reason to believe that subspace drift across waves and time within a topic is gradual rather than catastrophic in the survey setting we target.}

\subsection{Significance}
We conducted 5 runs with different random seeds on two representative datasets with Llama2 backbone and compared against SOTA. The results, reported in \ref{tab:std error}, show that APlaud consistently and robustly outperforms baselines. Due to time constraints, we could not repeat all settings, but we will include the full repeated results in the camera-ready version.

\begin{table}[t]
\centering
\caption{Siginificance performance comparison across different methods (mean $\pm$ std over 5 runs).}
\label{tab:std error}
\begin{tabular}{lcccc}
\toprule
\multirow{2}{*}{Method} & \multicolumn{2}{c}{G\&L} & \multicolumn{2}{c}{LAMP Movie Tagging} \\
 & ACC & F1 & ACC & F1 \\
\midrule
\multicolumn{5}{l}{\textit{Non-Personalized}} \\
LoRA          & 0.6402 $\pm$ 0.0052 & 0.6174 $\pm$ 0.0083 & 0.6220 $\pm$ 0.0082 & 0.5037 $\pm$ 0.0074 \\
QLoRA         & 0.6315 $\pm$ 0.0149 & 0.6192 $\pm$ 0.0175 & 0.6257 $\pm$ 0.0097 & 0.5045 $\pm$ 0.0247 \\
PiSSA         & 0.6588 $\pm$ 0.0116 & 0.6501 $\pm$ 0.0127 & 0.6274 $\pm$ 0.0103 & 0.5172 $\pm$ 0.0074 \\
MiLoRA        & 0.6620 $\pm$ 0.0079 & 0.6581 $\pm$ 0.0130 & 0.6297 $\pm$ 0.0068 & 0.5298 $\pm$ 0.0063 \\
AdaLoRA       & 0.6637 $\pm$ 0.0113 & 0.6503 $\pm$ 0.0148 & 0.6201 $\pm$ 0.0139 & 0.5102 $\pm$ 0.0083 \\
\midrule
\multicolumn{5}{l}{\textit{Personalized}} \\
OPPU          & 0.6602 $\pm$ 0.0094 & 0.6423 $\pm$ 0.0187 & 0.6356 $\pm$ 0.0106 & 0.5174 $\pm$ 0.0115 \\
Cu            & 0.6627 $\pm$ 0.0071 & 0.6507 $\pm$ 0.0263 & 0.6410 $\pm$ 0.0103 & 0.5289 $\pm$ 0.0137 \\
%Cu + SVD      & 0.6631 $\pm$ 0.0021 & 0.6515 $\pm$ 0.0054 & 0.6418 $\pm$ 0.0047 & 0.5355 $\pm$ 0.0059 \\
Aplaud+ & \textbf{0.6785 $\pm$ 0.0040} & \textbf{0.6657 $\pm$ 0.0082} & \textbf{0.6602 $\pm$ 0.0079} & \textbf{0.5512 $\pm$ 0.0091} \\
\bottomrule
\end{tabular}
\end{table}

\subsection{\textcolor{black}{Relative of Improvement Performance over different baselines}} \label{sec: roi}

\begin{table*}[t]
\centering
\caption{\textcolor{black}{Relative of Improvements of Aplaud+ over different baselines across datasets with llama2-7B backbone.}}
\label{tab:main_results_llam2_roi}
\resizebox{\textwidth}{!}{%
\begin{tabular}{lcccccccccccccc}
\toprule
\multirow{2}{*}{\textcolor{black}{Method}} 
& \multicolumn{2}{c}{\textcolor{black}{G\&L}} 
& \multicolumn{2}{c}{\textcolor{black}{TS}} 
& \multicolumn{2}{c}{\textcolor{black}{F\&R}} 
& \multicolumn{2}{c}{\textcolor{black}{EI}} 
& \multicolumn{2}{c}{\textcolor{black}{GSS}} 
& \multicolumn{2}{c}{\textcolor{black}{LAMP MV}} 
& \multirow{2}{*}{\textcolor{black}{Avg ACC Gain}} 
& \multirow{2}{*}{\textcolor{black}{Avg F1 Gain}} \\
\cmidrule(lr){2-3} \cmidrule(lr){4-5} \cmidrule(lr){6-7}
\cmidrule(lr){8-9} \cmidrule(lr){10-11} \cmidrule(lr){12-13}
& \textcolor{black}{ACC} & \textcolor{black}{Macro-F1} 
& \textcolor{black}{ACC} & \textcolor{black}{Macro-F1} 
& \textcolor{black}{ACC} & \textcolor{black}{Macro-F1} 
& \textcolor{black}{ACC} & \textcolor{black}{Macro-F1} 
& \textcolor{black}{ACC} & \textcolor{black}{Macro-F1} 
& \textcolor{black}{ACC} & \textcolor{black}{Macro-F1} 
& & \\
\midrule
\multicolumn{15}{l}{\textcolor{black}{\textbf{Non-Personalized}}} \\
\textcolor{black}{LoRA}      
& \textcolor{black}{+6.8\%}  & \textcolor{black}{+8.2\%}
& \textcolor{black}{+13.1\%} & \textcolor{black}{+27.7\%}
& \textcolor{black}{+10.6\%} & \textcolor{black}{+10.6\%}
& \textcolor{black}{+12.3\%} & \textcolor{black}{+26.4\%}
& \textcolor{black}{+4.9\%}  & \textcolor{black}{+9.7\%}
& \textcolor{black}{+6.1\%}  & \textcolor{black}{+8.9\%}
& \textcolor{black}{+8.9\%}
& \textcolor{black}{+15.2\%} \\
\textcolor{black}{PiSSA}     
& \textcolor{black}{+8.4\%}  & \textcolor{black}{+8.1\%}
& \textcolor{black}{+8.0\%}  & \textcolor{black}{+21.4\%}
& \textcolor{black}{+16.6\%} & \textcolor{black}{+8.1\%}
& \textcolor{black}{+8.4\%}  & \textcolor{black}{+12.3\%}
& \textcolor{black}{+11.5\%} & \textcolor{black}{+16.5\%}
& \textcolor{black}{+6.3\%}  & \textcolor{black}{+4.7\%}
& \textcolor{black}{+9.9\%}
& \textcolor{black}{+11.9\%} \\
\textcolor{black}{MiLoRA}    
& \textcolor{black}{+1.7\%}  & \textcolor{black}{+0.1\%}
& \textcolor{black}{+7.7\%}  & \textcolor{black}{+18.3\%}
& \textcolor{black}{+15.0\%} & \textcolor{black}{+5.7\%}
& \textcolor{black}{+12.2\%} & \textcolor{black}{+18.7\%}
& \textcolor{black}{+3.5\%}  & \textcolor{black}{+1.4\%}
& \textcolor{black}{+4.5\%}  & \textcolor{black}{+3.5\%}
& \textcolor{black}{+7.4\%}
& \textcolor{black}{+8.0\%} \\
\textcolor{black}{AdaLoRA}  
& \textcolor{black}{+1.9\%}  & \textcolor{black}{-0.5\%}
& \textcolor{black}{+2.5\%}  & \textcolor{black}{+14.4\%}
& \textcolor{black}{+14.7\%} & \textcolor{black}{+2.8\%}
& \textcolor{black}{+9.9\%}  & \textcolor{black}{+13.3\%}
& \textcolor{black}{+1.6\%}  & \textcolor{black}{-4.1\%}
& \textcolor{black}{+7.3\%}  & \textcolor{black}{+9.3\%}
& \textcolor{black}{+6.3\%}
& \textcolor{black}{+5.9\%} \\
\textcolor{black}{QLoRA}     
& \textcolor{black}{+3.2\%}  & \textcolor{black}{+2.6\%}
& \textcolor{black}{+6.8\%}  & \textcolor{black}{+28.4\%}
& \textcolor{black}{+18.0\%} & \textcolor{black}{+8.6\%}
& \textcolor{black}{+10.7\%} & \textcolor{black}{+26.8\%}
& \textcolor{black}{+6.2\%}  & \textcolor{black}{+7.4\%}
& \textcolor{black}{+4.6\%}  & \textcolor{black}{+5.3\%}
& \textcolor{black}{+8.2\%}
& \textcolor{black}{+13.2\%} \\
\midrule
\multicolumn{15}{l}{\textcolor{black}{\textbf{Personalized}}} \\
\textcolor{black}{GPT5-profile} 
& \textcolor{black}{+26.6\%} & \textcolor{black}{+24.4\%}
& \textcolor{black}{+29.2\%} & \textcolor{black}{+97.9\%}
& \textcolor{black}{+28.8\%} & \textcolor{black}{+28.7\%}
& \textcolor{black}{+13.5\%} & \textcolor{black}{+19.7\%}
& \textcolor{black}{-15.4\%} & \textcolor{black}{+5.8\%}
& \textcolor{black}{+20.4\%} & \textcolor{black}{+22.7\%}
& \textcolor{black}{+17.2\%}
& \textcolor{black}{+33.2\%} \\
\textcolor{black}{GPT5-RAG}    
& \textcolor{black}{+7.1\%}  & \textcolor{black}{+5.3\%}
& \textcolor{black}{+13.9\%} & \textcolor{black}{+10.6\%}
& \textcolor{black}{+3.4\%}  & \textcolor{black}{+4.6\%}
& \textcolor{black}{-0.6\%}  & \textcolor{black}{+1.5\%}
& \textcolor{black}{-22.3\%} & \textcolor{black}{-22.8\%}
& \textcolor{black}{--}      & \textcolor{black}{--}
& \textcolor{black}{+0.3\%}
& \textcolor{black}{-0.1\%} \\
\textcolor{black}{OPPU}        
& \textcolor{black}{+2.7\%}  & \textcolor{black}{+1.3\%}
& \textcolor{black}{+5.6\%}  & \textcolor{black}{+8.7\%}
& \textcolor{black}{+4.7\%}  & \textcolor{black}{+2.5\%}
& \textcolor{black}{+3.5\%}  & \textcolor{black}{+2.2\%}
& \textcolor{black}{+7.2\%}  & \textcolor{black}{+5.1\%}
& \textcolor{black}{+4.1\%}  & \textcolor{black}{+7.4\%}
& \textcolor{black}{+4.6\%}
& \textcolor{black}{+4.5\%} \\
\bottomrule
\end{tabular}}
\end{table*}
\vspace{-2mm}

\begin{table*}[t]
\centering
\caption{\textcolor{black}{Relative of Improvements of Aplaud+ over different baselines with Mistral-7B backbone.}}
\label{tab:main_mistral7B_roi}
\resizebox{\textwidth}{!}{%
\begin{tabular}{lcccccccccccccc}
\toprule
\multirow{2}{*}{\textcolor{black}{Method}} 
& \multicolumn{2}{c}{\textcolor{black}{G\&L}} 
& \multicolumn{2}{c}{\textcolor{black}{TS}} 
& \multicolumn{2}{c}{\textcolor{black}{F\&R}} 
& \multicolumn{2}{c}{\textcolor{black}{EI}} 
& \multicolumn{2}{c}{\textcolor{black}{GSS}} 
& \multicolumn{2}{c}{\textcolor{black}{LAMP MV}} 
& \multirow{2}{*}{\textcolor{black}{Avg ACC Gain}} 
& \multirow{2}{*}{\textcolor{black}{Avg F1 Gain}}\\
\cmidrule(lr){2-3} \cmidrule(lr){4-5} \cmidrule(lr){6-7}
\cmidrule(lr){8-9} \cmidrule(lr){10-11} \cmidrule(lr){12-13}
& \textcolor{black}{ACC} & \textcolor{black}{Macro-F1}
& \textcolor{black}{ACC} & \textcolor{black}{Macro-F1}
& \textcolor{black}{ACC} & \textcolor{black}{Macro-F1}
& \textcolor{black}{ACC} & \textcolor{black}{Macro-F1}
& \textcolor{black}{ACC} & \textcolor{black}{Macro-F1}
& \textcolor{black}{ACC} & \textcolor{black}{Macro-F1}
& & \\
\midrule
\multicolumn{15}{l}{\textcolor{black}{\textbf{Non-Personalized}}} \\
\textcolor{black}{LoRA}
& \textcolor{black}{+4.9\%} & \textcolor{black}{+6.3\%}
& \textcolor{black}{+11.2\%} & \textcolor{black}{+72.1\%}
& \textcolor{black}{+15.1\%} & \textcolor{black}{+84.2\%}
& \textcolor{black}{+6.0\%} & \textcolor{black}{+14.9\%}
& \textcolor{black}{+6.2\%} & \textcolor{black}{+31.8\%}
& \textcolor{black}{+6.2\%} & \textcolor{black}{+3.4\%}
& \textcolor{black}{+8.3\%}
& \textcolor{black}{+35.5\%} \\
\textcolor{black}{PiSSA}
& \textcolor{black}{+5.4\%}  & \textcolor{black}{+4.3\%}
& \textcolor{black}{+6.6\%}  & \textcolor{black}{+39.4\%}
& \textcolor{black}{+20.9\%} & \textcolor{black}{+83.0\%}
& \textcolor{black}{+10.3\%} & \textcolor{black}{+23.8\%}
& \textcolor{black}{+12.0\%} & \textcolor{black}{+22.2\%}
& \textcolor{black}{+4.1\%}  & \textcolor{black}{+1.8\%}
& \textcolor{black}{+9.9\%}
& \textcolor{black}{+29.1\%} \\
\textcolor{black}{MiLoRA}
& \textcolor{black}{+4.4\%}  & \textcolor{black}{+3.7\%}
& \textcolor{black}{+5.6\%}  & \textcolor{black}{+42.9\%}
& \textcolor{black}{+18.0\%} & \textcolor{black}{+69.8\%}
& \textcolor{black}{+7.7\%}  & \textcolor{black}{+17.0\%}
& \textcolor{black}{+1.6\%}  & \textcolor{black}{+9.9\%}
& \textcolor{black}{+3.8\%}  & \textcolor{black}{+3.8\%}
& \textcolor{black}{+6.9\%}
& \textcolor{black}{+24.5\%} \\
\textcolor{black}{AdaLoRA}
& \textcolor{black}{+7.0\%}  & \textcolor{black}{+6.7\%}
& \textcolor{black}{+11.2\%} & \textcolor{black}{+51.1\%}
& \textcolor{black}{+8.9\%}  & \textcolor{black}{+19.6\%}
& \textcolor{black}{+6.0\%}  & \textcolor{black}{+23.9\%}
& \textcolor{black}{-18.5\%} & \textcolor{black}{-3.2\%}
& \textcolor{black}{+6.7\%}  & \textcolor{black}{-2.7\%}
& \textcolor{black}{+3.5\%}
& \textcolor{black}{+15.8\%} \\
\textcolor{black}{QLoRA}
& \textcolor{black}{+5.7\%}  & \textcolor{black}{+5.9\%}
& \textcolor{black}{+9.5\%}  & \textcolor{black}{+40.5\%}
& \textcolor{black}{+20.5\%} & \textcolor{black}{+75.7\%}
& \textcolor{black}{+6.2\%}  & \textcolor{black}{+16.7\%}
& \textcolor{black}{+3.8\%}  & \textcolor{black}{+19.3\%}
& \textcolor{black}{+0.5\%}  & \textcolor{black}{-5.7\%}
& \textcolor{black}{+7.7\%}
& \textcolor{black}{+25.4\%} \\
\midrule
\multicolumn{15}{l}{\textcolor{black}{\textbf{Personalized}}} \\
\textcolor{black}{GPT5-profile}
& \textcolor{black}{+27.4\%} & \textcolor{black}{+26.3\%}
& \textcolor{black}{+27.4\%} & \textcolor{black}{+108.9\%}
& \textcolor{black}{+31.9\%} & \textcolor{black}{+51.8\%}
& \textcolor{black}{+9.3\%}  & \textcolor{black}{+10.6\%}
& \textcolor{black}{-1.7\%}  & \textcolor{black}{+34.6\%}
& \textcolor{black}{+29.2\%} & \textcolor{black}{+15.4\%}
& \textcolor{black}{+20.6\%}
& \textcolor{black}{+41.3\%} \\
\textcolor{black}{GPT5-RAG}
& \textcolor{black}{+7.8\%}  & \textcolor{black}{+6.9\%}
& \textcolor{black}{+12.3\%} & \textcolor{black}{+16.8\%}
& \textcolor{black}{+5.9\%}  & \textcolor{black}{+23.3\%}
& \textcolor{black}{-4.2\%}  & \textcolor{black}{-6.3\%}
& \textcolor{black}{-9.8\%}  & \textcolor{black}{-1.7\%}
& \textcolor{black}{--}      & \textcolor{black}{--}
& \textcolor{black}{+2.4\%}
& \textcolor{black}{+7.8\%} \\
\textcolor{black}{OPPU}
& \textcolor{black}{+2.1\%}  & \textcolor{black}{+2.8\%}
& \textcolor{black}{+0.1\%}  & \textcolor{black}{+3.5\%}
& \textcolor{black}{+2.7\%}  & \textcolor{black}{+19.6\%}
& \textcolor{black}{+1.3\%}  & \textcolor{black}{-1.9\%}
& \textcolor{black}{+6.5\%}  & \textcolor{black}{+19.0\%}
& \textcolor{black}{+2.4\%}  & \textcolor{black}{+18.3\%}
& \textcolor{black}{+2.5\%}
& \textcolor{black}{+10.2\%} \\

\bottomrule
\end{tabular}}
\end{table*}

\textcolor{black}{In this section, we present the Relative of Improvement (ROI) of our method (we report results of Aplaud+ as a representative here) over all other baselines in Table~\ref{tab:main_results_llam2_roi} and Table~\ref {tab:main_mistral7B_roi}.} 

\textcolor{black}{Compared with non-personalized PEFT methods (LoRA, PiSSA, MiLoRA, AdaLoRA, QLoRA), Aplaud+ provides clear and stable gains. For example, on Llama2-7B it improves Macro-F1 by +27.7\% on TS and +10.6\% on F\&R over LoRA, and on Mistral-7B it further increases TS Macro-F1 by +51.1\% over AdaLoRA. These results indicate that generic finetuning cannot capture user-specific heterogeneity, while Aplaud+ effectively personalizes model behavior.}

\textcolor{black}{Compared with retrieval-based personalization (GPT5-profile / GPT5-RAG), Aplaud+ delivers significantly higher and more robust performance. On Llama2-7B, it surpasses GPT5-profile by +17.2\% ACC and +33.2\% Macro-F1 on average. Aplaud avoids dependence on prompt context quality and instead encodes stable user preferences in parameters.}

\textcolor{black}{Compared with OPPU, a strong personalized baseline, Aplaud still achieves consistent improvements. On Llama2-7B it yields +4.6\% ACC and +4.5\% Macro-F1 gains on average, and on Mistral-7B the gains reach +2.5\% ACC and +10.2\% Macro-F1. This demonstrates the superiority of our meticulously designed personalized modules to capture user preferences}.

\subsection{Qwen2.5-7B results}
We report additional results on the Qwen2.5-7B backbone in Table~\ref{tab:qwen_results}. The results show that APLaud/APLaud+ continue to outperform strong baselines, including LoRA and the personalized method OPPU, while remaining competitive with or superior to GPT-based methods. This confirms that our approach is robust across modern LLM backbones and is not limited to older architectures.

\begin{table*}[t]
\centering
\caption{Performance comparison on Qwen2.5-7B across six benchmarks.}
\label{tab:qwen_results}
\small
\setlength{\tabcolsep}{2.5pt}
\begin{tabular}{lcccccccccccc}
\toprule
\multirow{2}{*}{Method}
& \multicolumn{2}{c}{G\&L}
& \multicolumn{2}{c}{TS}
& \multicolumn{2}{c}{F\&R}
& \multicolumn{2}{c}{EI}
& \multicolumn{2}{c}{GSS}
& \multicolumn{2}{c}{LAMP MV} \\
\cmidrule(lr){2-3}
\cmidrule(lr){4-5}
\cmidrule(lr){6-7}
\cmidrule(lr){8-9}
\cmidrule(lr){10-11}
\cmidrule(lr){12-13}
& ACC & F1 & ACC & F1 & ACC & F1 & ACC & F1 & ACC & F1 & ACC & F1 \\
\midrule
LoRA
& 0.6522 & 0.6418
& 0.7214 & 0.4529
& 0.5979 & 0.3712
& 0.4700 & 0.3344
& 0.5028 & 0.3071
& 0.5999 & 0.4773 \\

GPT5-Profile
& 0.5394 & 0.5340
& 0.6170 & 0.3448
& 0.4954 & 0.2879
& 0.4566 & 0.3491
& 0.4689 & 0.2570
& -- & -- \\

GPT5-RAG
& 0.6377 & 0.6306
& 0.7001 & 0.6169
& 0.6174 & 0.3542
& 0.5213 & 0.4117
& 0.5106 & 0.3520
& -- & -- \\

OPPU
& 0.6634 & 0.6562
& 0.7700 & 0.6926
& 0.6304 & 0.4027
& 0.4875 & 0.3535
& 0.5042 & 0.3268
& 0.6396 & 0.5381 \\

Cu
& 0.6800 & 0.6743
& 0.7670 & 0.6587
& 0.6044 & 0.3677
& 0.4892 & 0.3542
& 0.5052 & 0.3248
& 0.6384 & 0.5370 \\

APLaud
& 0.6844 & 0.6764
& 0.7852 & 0.7205
& 0.6381 & 0.3974
& 0.5058 & 0.3787
& 0.5105 & 0.3311
& \textbf{0.6439} & \textbf{0.5397} \\

APLaud+
& \textbf{0.6860} & \textbf{0.6773}
& \textbf{0.7943} & \textbf{0.7253}
& \textbf{0.6524} & \textbf{0.4213}
& 0.5075 & \textbf{0.4269}
& \textbf{0.5141} & 0.3459
& 0.6420 & 0.5386 \\
\bottomrule
\end{tabular}
\end{table*}

\subsection{Performance Parameter Tradeoff Analsysis}
\label{app:performance_tradeoff}

To further examine whether APLaud provides a more parameter-efficient personalization mechanism, we conduct an additional performance--parameter tradeoff analysis on the TS dataset using Qwen2.5-7B as the backbone model. Specifically, we vary the LoRA rank in OPPU to control its per-user parameter budget and compare it with APLaud and APLaud+. The results are shown in Table~\ref{tab:tradeoff_ts_qwen}.

\begin{table}[t]
\centering
\small
\setlength{\tabcolsep}{5pt}
\begin{tabular}{lccc}
\toprule
\textbf{\% Params} & \textbf{Method} & \textbf{ACC} & \textbf{Macro-F1} \\
\midrule
4.240\% & OPPU ($r=128$) & 0.7700 & 0.6833 \\
2.120\% & OPPU ($r=64$)  & 0.7700 & 0.6926 \\
1.060\% & OPPU ($r=32$)  & 0.7693 & 0.6905 \\
0.530\% & OPPU ($r=16$)  & 0.7613 & 0.6615 \\
0.270\% & OPPU ($r=8$)   & 0.7579 & 0.5965 \\
0.130\% & OPPU ($r=4$)   & 0.7550 & 0.5358 \\
0.066\% & OPPU ($r=2$)   & 0.7406 & 0.5323 \\
0.033\% & OPPU ($r=1$)   & 0.7305 & 0.5282 \\
0.000\% & Non-personalized LoRA & 0.7214 & 0.4529 \\
\midrule
0.044\% & APLaud   & 0.7852 & 0.7205 \\
0.038\% & APLaud+  & \textbf{0.7943} & \textbf{0.7253} \\
\bottomrule
\end{tabular}
\caption{Performance--parameter tradeoff on the TS dataset with Qwen2.5-7B. Here, \% Params is computed as 
\(\frac{\#\text{ user-specific trainable parameters}}{\#\text{ total parameters of the 7B backbone}} \times 100\%\), 
i.e., it reflects the individual per-user trainable parameter ratio relative to the full model size.}
\label{tab:tradeoff_ts_qwen}
\end{table}

\paragraph{APLaud achieves strong performance with extremely small parameter budgets.}
APLaud and APLaud+ achieve the best overall results while using only \(0.044\%\) and \(0.038\%\) of the total model parameters as user-specific trainable parameters, respectively. These budgets are substantially smaller than most OPPU configurations. For example, compared with OPPU at rank \(64\), APLaud uses roughly \(48\times\) fewer per-user parameters while achieving higher ACC and Macro-F1.

\paragraph{APLaud outperforms OPPU under comparable or larger parameter budgets.}
Even when OPPU is assigned much larger per-user parameter budgets, it does not match the performance of APLaud or APLaud+. OPPU achieves its best Macro-F1 of \(0.6926\) at \(2.120\%\) parameters, while APLaud reaches \(0.7205\) Macro-F1 using only \(0.044\%\) parameters, and APLaud+ further improves to \(0.7253\) Macro-F1 with \(0.038\%\) parameters. This demonstrates that APLaud is not merely smaller, but also provides a more effective parameterization for personalization.

Overall, these results provide direct empirical evidence that APLaud achieves a stronger performance--parameter tradeoff than OPPU. By reusing a shared adaptation subspace and learning compact user-specific modulation, APLaud obtains better personalization performance while substantially reducing the per-user trainable parameter cost.

\subsection{Wasserstein Distance Result}
\label{appendix: wd distance}
As an addition to Table \ref{tab:main_results_llam2}, Table~\ref{table:wd} presents the results on approximating human responses for Pew Research Center surveys and the General Social Survey under the same setting. We report Accuracy (ACC), F1 Score (F1), and Wasserstein Distance (WD), with WD measuring the average distributional distance between real human subjects and simulated virtual subjects across test survey questions. A lower WD indicates better distributional alignment.
Across all datasets, the \textbf{Aplaud+} method consistently achieves the \textbf{lowest Wasserstein Distance (WD)}, indicating superior alignment between the distributions of human and simulated responses. 
\definecolor{lightgray}{gray}{0.95}
\begin{table*}[htbp]
\footnotesize
\centering
\caption{WD Performance across all survey datasets.}
\label{table:wd}
\vspace{0.5em}
\rowcolors{2}{white}{lightgray}
\resizebox{\textwidth}{!}{%
\begin{tabular}{l ccc ccc ccc ccc ccc}
\toprule
\multirow{2}{*}{\textbf{Method}} 
& \multicolumn{3}{c}{\textbf{G\&L}} 
& \multicolumn{3}{c}{\textbf{TS}} 
& \multicolumn{3}{c}{\textbf{F\&R}} 
& \multicolumn{3}{c}{\textbf{EI}} 
& \multicolumn{3}{c}{\textbf{GSS}} \\
\cmidrule(lr){2-16}
& ACC & F1 & WD & ACC & F1 & WD & ACC & F1 & WD & ACC & F1 & WD & ACC & F1 & WD \\
\midrule
LoRA         & 0.6554 & 0.6342 & 0.1111 & 0.7072 & 0.4186 & 0.0294 & 0.5681 & 0.2372 & 0.4462 & 0.4708 & 0.3356 & 0.3350 & 0.4336 & 0.2624 & 0.2782 \\
OPPU         & 0.6731 & 0.6560 & 0.0725 & 0.7852 & 0.6962 & 0.0324 & 0.6368 & 0.3654 & 0.2127 & 0.4925 & \textbf{0.3936} & \textbf{0.0933} & 0.4322 & 0.2906 & 0.1977 \\
Cu           & 0.6828 & 0.6691 & \textbf{0.0097} & 0.7781 & 0.6338 & \textbf{0.0203} & 0.5746 & 0.2796 & 0.4047 & 0.5008 & 0.3868 & 0.1625 & 0.4548 & 0.2917 & 0.2274 \\
%PuQu       & 0.6844 & 0.6705 & 0.0193 & 0.7791 & 0.6342 & 0.0253 & 0.5783 & 0.2783 & 0.4086 & \textbf{0.5017} & 0.3856 & 0.1708 & 0.4548 & 0.2897 & 0.2246 \\
Aplaud+  & \textbf{0.6876} & \textbf{0.6742} & \textbf{0.0097} & \textbf{0.7862} & \textbf{0.7204} & 0.0324 & \textbf{0.6537} & \textbf{0.4369} & \textbf{0.1362} & 0.4992 & 0.3859 & 0.1558 & \textbf{0.4605} & \textbf{0.3459} & \textbf{0.1949} \\
\bottomrule
\end{tabular}
}
\vspace{0.5em}
\end{table*}

For instance, on the \textbf{GSS} dataset, the WD achieved by Aplaud+ is \textbf{0.1949}, outperforming both the \textbf{LoRA} baseline (0.2782) and other personalized approaches. This trend holds across various survey domains:
\begin{itemize}
  \item In \textbf{Gender and Leadership (G\&L)}, our method achieves a WD of \textbf{0.0097}, which is substantially lower than LoRA (0.1111) and OPPU (0.0725).
  \item In \textbf{Trust in Science (TS)}, Cu records a WD of \textbf{0.0203}, outperforming LoRA (0.0294) and OPPU (0.0324).
  \item For \textbf{Friendship and Relationships (F\&R)}, the WD drops to \textbf{0.1362}, compared to \textbf{0.4462} with LoRA and \textbf{0.2127} with OPPU.
  \item In \textbf{Economic Inequality (EI)}, our model achieves a WD of \textbf{0.1558}, improving over LoRA (0.3350) and slightly increasing to OPPU (0.0933).
  \item In \textbf{Economic Inequality (EI)}, our model achieves a WD of \textbf{0.1558}, improving over LoRA (0.3350) and slightly increasing to OPPU (0.0933).
\end{itemize}
The results underscore the effectiveness of structure-aware personalization, particularly the bias-corrected matrix factorization \textbf{Aplaud+}, in accurately capturing subtle, user-specific behavioral patterns across diverse survey domains.

\subsection{Subgroup Comparisons Across Demographic Variables}
To assess how well each personalized model generalizes across diverse population subgroups, we conduct stratified evaluations along three key demographic dimensions: geographic region (CREGION), sex (SEX), and political affiliation (POLPARTY). This analysis spans four waves of the Pew American Trends Panel (ATP)—Waves 36, 42, 50, and 54—as well as the 2016–2020 General Social Survey (GSS) Panel.

For each dataset, we report model performance within each subgroup using two metrics: classification accuracy (ACC) and Wasserstein Distance (WD). We highlight the best-performing results in each subgroup to assess the consistency, fairness, and personalization quality of different adaptation methods.
\paragraph{Subgroup definitions:}
\begin{itemize}
    \item \textbf{CREGION} $\in$ \{Northeast, Midwest, South, West\}
    \item \textbf{SEX} $\in$ \{Male, Female\}
    \item \textbf{POLPARTY} $\in$ \{Republican, Democrat, Independent, Other\}
\end{itemize}

\subsubsection{ ATP Wave 36}
\paragraph{Subgroup Performance in ATP Wave 36 by Region (CREGION).}
As shown in Table~\ref{table:cregion_w36}, \textbf{Aplaud+} yields the highest accuracy in the West (0.7671) and consistently performs well across other regions. In contrast, OPPU achieved the lowest Wasserstein Distance in the Northeast (0.0323), indicating stronger distributional alignment in that specific subgroup.
Overall, these results reinforce the strength of structured personalization in APlaud, which combines shared matrix decomposition with user-specific adaptation. 

\begin{table*}[htbp]
\centering
\caption{ATP Wave 36: Performance by Region (CREGION). }
\label{table:cregion_w36}
\vspace{0.3em}
\begin{tabular}{lcccccccc}
\toprule
\textbf{Model} & \multicolumn{2}{c}{Midwest} & \multicolumn{2}{c}{Northeast} & \multicolumn{2}{c}{South} & \multicolumn{2}{c}{West} \\
 & Acc & WD & Acc & WD & Acc & WD & Acc & WD \\
\midrule
LoRA & 0.6552 & 0.1437 & 0.5376 & 0.1613 & \textbf{0.6683} & 0.0913 & 0.7123 & 0.0685 \\
OPPU & \textbf{0.6782} & 0.0920 & 0.5914 & \textbf{0.0323} & 0.6394 & 0.0817 & 0.7671 & 0.0753 \\
Cu & 0.6667 & 0.0345 & \textbf{0.6559} & 0.1075 & 0.6538 & \textbf{0.0529} & 0.7603 & 0.0753 \\
PuQu & 0.6724 & \textbf{0.0230} & 0.6452 & 0.0968 & 0.6538 & 0.0577 & 0.7671 & 0.0890 \\
Aplaud+ & \textbf{0.6782} & 0.0287 & \textbf{0.6559} & 0.1075 & 0.6538 & 0.0577 & \textbf{0.7671} & \textbf{0.0616} \\
\bottomrule
\end{tabular}
\end{table*}

\paragraph{Subgroup Performance in ATP Wave 36 by Gender (SEX).}
As shown in Table~\ref{table:sex_w36}, \textbf{Aplaud+} achieves the highest classification accuracy for both Female (0.6849) and Male (0.6914) respondents.
In terms of distributional alignment, our method showed significant gains. For Female users, \textbf{PuQu} achieved a Wasserstein Distance (WD) of 0.0055, representing a relative reduction of 92.8\% compared to LoRA (0.0767), and 90.4\% compared to OPPU (0.0575). For Male users, \textbf{Aplaud+} yields the lowest WD (0.0469), making a 70.7\% improvement over LoRA (0.1602), and 50.0\% better than OPPU (0.0938).

\begin{table}[htbp]
\centering
\caption{ATP Wave 36: Performance by Gender (SEX). Best values per column are \textbf{bolded}.}
\label{table:sex_w36}
\vspace{0.3em}
\begin{tabular}{lcccc}
\toprule
\textbf{Model} & \multicolumn{2}{c}{Female} & \multicolumn{2}{c}{Male} \\
 & Acc & WD & Acc & WD \\
\midrule
LoRA & 0.6521 & 0.0767 & 0.6602 & 0.1602 \\
OPPU & 0.6685 & 0.0575 & 0.6797 & 0.0938 \\
Cu & 0.6795 & 0.0219 & 0.6875 & 0.0547 \\
PuQu & 0.6795 & \textbf{0.0055} & \textbf{0.6914} & 0.0547 \\
Aplaud+ & \textbf{0.6849} & 0.0164 & \textbf{0.6914} & \textbf{0.0469} \\
\bottomrule
\end{tabular}
\end{table}
\paragraph{Subgroup Performance in ATP Wave 36 by Political Affiliation (POLPARTY).}
Table~\ref{table:polparty_w36} presents model performance stratified by political affiliation—Democrat, Republican, Independent, and Other—based on ATP Wave 36. \textbf{Aplaud+} achieved the highest classification accuracy for both Democrats (0.7095) and Independents (0.6968), while Cu and PuQu also demonstrated consistently strong generalization across subgroups. 

In terms of distributional alignment, OPPU achieved the lowest Wasserstein Distance for Democrats (0.0207) and Independents (0.0194), whereas \textbf{Aplaud+} performs best for Republicans (0.0733) and users classified as Other (0.1176). These findings suggest that structured personalization approaches, such as APlaud can effectively adapt to diverse political profiles, yielding both accurate and distributionally faithful response simulations.

\begin{table*}[htbp]
\centering
\caption{ATP Wave 36: Performance by Political Party (POLPARTY).}
\label{table:polparty_w36}
\vspace{0.3em}
\begin{tabular}{lcccccccc}
\toprule
\textbf{Model} & \multicolumn{2}{c}{Democrat} & \multicolumn{2}{c}{Republican} & \multicolumn{2}{c}{Independent} & \multicolumn{2}{c}{Other} \\
 & Acc & WD & Acc & WD & Acc & WD & Acc & WD \\
\midrule
LoRA & 0.6763 & 0.0830 & 0.6492 & 0.3455 & 0.6258 & 0.1161 & \textbf{0.6765} & 0.1471 \\
OPPU & 0.6805 & \textbf{0.0207} & \textbf{0.6649} & 0.1675 & 0.6839 & \textbf{0.0194} & 0.6176 & 0.1471 \\
Cu & 0.7054 & 0.0456 & 0.6545 & 0.0785 & 0.6903 & 0.0258 & 0.6471 & \textbf{0.1176} \\
PuQu & 0.7012 & 0.0373 & \textbf{0.6649} & 0.0995 & 0.6903 & 0.0258 & 0.6471 & \textbf{0.1176} \\
Aplaud+ & \textbf{0.7095} & 0.0498 & 0.6597 & \textbf{0.0733} & \textbf{0.6968} & 0.0258 & 0.6471 & \textbf{0.1176} \\
\bottomrule
\end{tabular}
\end{table*}

\subsubsection{ATP Wave 42}
\textit{Topic: Trust in Science}

\paragraph{Subgroup Performance in ATP Wave 42 by Region (CREGION).}
Table~\ref{table:cregion_w42} presents the subgroup performance across geographic regions. The \textbf{Cu} model achieved the highest accuracy in the Midwest (0.8008) and the lowest Wasserstein Distance (WD) of 0.0456, indicating strong performance in this region. In the Northeast, \textbf{PuQu} attains the best accuracy (0.7574) and the lowest WD (0.0221), reflecting excellent distributional alignment. While \textbf{Aplaud+} demonstrates the highest accuracy in both the South (0.8232) and West (0.7766), it did not achieve the lowest WD in these regions, suggesting that its distributional alignment was not optimal compared to other models.

\begin{table*}[htbp]
\centering
\caption{ATP Wave 42: Performance by Region (CREGION).}
\label{table:cregion_w42}
\vspace{0.3em}
\begin{tabular}{lcccccccc}
\toprule
\textbf{Model} & \multicolumn{2}{c}{Midwest} & \multicolumn{2}{c}{Northeast} & \multicolumn{2}{c}{South} & \multicolumn{2}{c}{West} \\
 & Acc & WD & Acc & WD & Acc & WD & Acc & WD \\
\midrule
LoRA & 0.7178 & 0.0539 & 0.6985 & 0.0368 & 0.7226 & \textbf{0.0244} & 0.6844 & 0.0390 \\
OPPU & 0.7759 & 0.0705 & 0.7574 & 0.0294 & 0.8171 & 0.0396 & 0.7695 & 0.0142 \\
Cu & \textbf{0.8008} & \textbf{0.0456} & 0.7426 & 0.0221 & 0.7866 & 0.0274 & 0.7660 & 0.0177 \\
PuQu & 0.7967 & 0.0498 & \textbf{0.7574} & \textbf{0.0221} & 0.7835 & 0.0305 & 0.7695 & \textbf{0.0142} \\
Aplaud+ & 0.7801 & 0.0664 & 0.7279 & 0.0809 & \textbf{0.8232} & 0.0396 & \textbf{0.7766} & 0.0213 \\
\bottomrule
\end{tabular}
\end{table*}

\paragraph{Subgroup Performance in ATP Wave 42 by Gender (SEX).}
Table~\ref{table:sex_w42} presents model performance by gender for ATP Wave 42, focusing on Trust in Science. \textbf{Aplaud+} achieves the highest accuracy among female respondents (0.7805), whereas OPPU yields the highest accuracy for male respondents (0.7955). In terms of Wasserstein Distance, reflecting distributional alignment, \textbf{Cu}, \textbf{PuQu}, and \textbf{Aplaud+} achieve equally strong alignment (WD = 0.0561) among female respondents. For male respondents, \textbf{PuQu} demonstrates the best distributional alignment, achieving the lowest WD (0.0035). These results underscore that structured personalization methods, particularly those incorporating low-rank decomposition and residual correction, effectively enhance prediction accuracy and response alignment across gender subgroups.
\begin{table}[htbp]
\centering
\caption{ATP Wave 42: Performance by Gender (SEX).}
\label{table:sex_w42}
\vspace{0.3em}
\begin{tabular}{lcccc}
\toprule
\textbf{Model} & \multicolumn{2}{c}{Female} & \multicolumn{2}{c}{Male} \\
 & Acc & WD & Acc & WD \\
\midrule
LoRA & 0.7098 & 0.0756 & 0.7054 & 0.0451 \\
OPPU & 0.7707 & 0.0585 & \textbf{0.7955} & 0.0139 \\
Cu & 0.7780 & \textbf{0.0561} & 0.7782 & 0.0052 \\
PuQu & 0.7780 & \textbf{0.0561} & 0.7799 & \textbf{0.0035} \\
Aplaud+ & \textbf{0.7805} & \textbf{0.0561} & 0.7903 & 0.0191 \\
\bottomrule
\end{tabular}
\end{table}
\paragraph{Subgroup Performance in ATP Wave 42 by Political Affiliation (POLPARTY).}
Table~\ref{table:polparty_w42} summarizes performance across political affiliation subgroups for ATP Wave 42, focused on Trust in Science. \textbf{Aplaud+} achieves the highest accuracy among Democrats (0.8293), while OPPU yielded the best accuracy for Republicans (0.8143) and Independents (0.7838). \textbf{Cu} provided the highest accuracy among respondents identifying as "Other" (0.7828). Regarding Wasserstein Distance (WD), OPPU has the lowest WD among Democrats (0.0585), \textbf{Aplaud+} achieved the lowest WD for Republicans (0.0643), \textbf{Cu} performed best for Independents (0.0113), and \textbf{LoRA} obtains the lowest WD for "Other" affiliations (0.0505). 

\begin{table*}[htbp]
\centering
\caption{ATP Wave 42: Performance by Political Party (POLPARTY).}
\label{table:polparty_w42}
\vspace{0.3em}
\begin{tabular}{lcccccccc}
\toprule
\textbf{Model} & \multicolumn{2}{c}{Democrat} & \multicolumn{2}{c}{Republican} & \multicolumn{2}{c}{Independent} & \multicolumn{2}{c}{Other} \\
 & Acc & WD & Acc & WD & Acc & WD & Acc & WD \\
\midrule
LoRA & 0.6976 & 0.0732 & 0.7429 & 0.1286 & 0.7095 & 0.0541 & 0.6869 & \textbf{0.0505} \\
OPPU & 0.8244 & \textbf{0.0585} & \textbf{0.8143} & 0.0714 & \textbf{0.7838} & 0.0405 & 0.7273 & 0.0707 \\
Cu & 0.7902 & 0.0829 & 0.7929 & 0.0786 & 0.7658 & \textbf{0.0113} & \textbf{0.7828} & 0.0556 \\
PuQu & 0.7854 & 0.0780 & 0.8000 & 0.0714 & 0.7725 & 0.0180 & 0.7727 & 0.0657 \\
Aplaud+ & \textbf{0.8293} & 0.0732 & 0.8071 & \textbf{0.0643} & 0.7748 & 0.0338 & 0.7525 & 0.0657 \\
\bottomrule
\end{tabular}
\end{table*}

\subsubsection{ATP Wave 50}
\textit{Topic: Family and Relationsihp}
\paragraph{Subgroup Performance in ATP Wave 50 by Region (CREGION).}
Table~\ref{table:cregion_w50} summarizes model performance across U.S. regions for ATP Wave 50, which focuses on Family and Relationship topics. The \textbf{Aplaud+} model achieved the highest accuracy in the Northeast (0.7023), South (0.6128) and West (0.6667) regions. Additionally, \textbf{Aplaud+} achieves the lowest Wasserstein Distance (WD) values in four regions, highlighting its superior alignment with real response distributions in these regions. The OPPU model attains the highest accuracy in the Midwest (0.6715), along with competitive WD performance. These results demonstrate that structured personalization, particularly through the \textbf{Aplaud+} method, significantly enhances both accuracy and distributional fidelity across geographic subpopulations.

\begin{table*}[htbp]
\centering
\caption{ATP Wave 50: Performance by Region (CREGION).}
\label{table:cregion_w50}
\vspace{0.3em}
\begin{tabular}{lcccccccc}
\toprule
\textbf{Model} & \multicolumn{2}{c}{Midwest} & \multicolumn{2}{c}{Northeast} & \multicolumn{2}{c}{South} & \multicolumn{2}{c}{West} \\
 & Acc & WD & Acc & WD & Acc & WD & Acc & WD \\
\midrule
LoRA & 0.5839 & 0.4088 & 0.5878 & 0.4504 & 0.6015 & 0.4286 & 0.5105 & 0.4852 \\
OPPU & \textbf{0.6715} & 0.2409 & 0.6794 & 0.2137 & 0.5940 & 0.1316 & 0.6414 & 0.2869 \\
Cu & 0.6277 & 0.4015 & 0.6031 & 0.4351 & 0.5677 & 0.3383 & 0.5359 & 0.4641 \\
PuQu & 0.6204 & 0.4088 & 0.6031 & 0.4351 & 0.5639 & 0.3459 & 0.5359 & 0.4641 \\
Aplaud+ & 0.6642 & \textbf{0.1898} & \textbf{0.7023} & \textbf{0.0992} & \textbf{0.6128} & \textbf{0.0977} & \textbf{0.6667} & \textbf{0.2025} \\
\bottomrule
\end{tabular}
\end{table*}

\paragraph{Subgroup Performance in ATP Wave 50 by Gender (SEX).}
As reported in Table~\ref{table:sex_w50}, \textbf{Aplaud+} achieves the best accuracy for both Female (0.6510) and Male (0.6590) subgroups. In addition, it showed the strongest distributional alignment, with WDs of 0.1000 and 0.2069 respectively. This supports the robustness of our personalized decomposition strategy between genders.

\begin{table}[htbp]
\centering
\caption{ATP Wave 50: Performance by Gender (SEX).}
\label{table:sex_w50}
\vspace{0.3em}
\begin{tabular}{lcccc}
\toprule
\textbf{Model} & \multicolumn{2}{c}{Female} & \multicolumn{2}{c}{Male} \\
 & Acc & WD & Acc & WD \\
\midrule
LoRA & 0.5471 & 0.4333 & 0.6092 & 0.4713 \\
OPPU & 0.6392 & 0.1706 & 0.6322 & 0.2950 \\
Cu & 0.5451 & 0.3765 & 0.6322 & 0.4598 \\
PuQu & 0.5431 & 0.3804 & 0.6284 & 0.4636 \\
Aplaud+ & \textbf{0.6510} & \textbf{0.1000} & \textbf{0.6590} & \textbf{0.2069} \\
\bottomrule
\end{tabular}
\end{table}

\paragraph{Subgroup Performance in ATP Wave 50 by Political Affiliation (POLPARTY).}
Table~\ref{table:polparty_w50} illustrates the performance across political affiliation subgroups in ATP Wave 50, focusing on family and relationship issues. The \textbf{Aplaud+} model demonstrates superior accuracy among Democrats (0.6349) and Republicans (0.7137), as well as competitive performance for Independents (0.5644) and Others (0.6442). Additionally, \textbf{Aplaud+} achieves the lowest Wasserstein Distance (WD) for Democrats (0.1905) and Republicans (0.0745). The OPPU method also shows strong performance, especially among Independents, attaining both the highest accuracy (0.5743) and lowest WD (0.0990). These findings emphasize that structured personalization, particularly the \textbf{Aplaud+} approach, effectively captures nuanced subgroup differences and improves alignment with real-world response distributions across political affiliations.

\begin{table*}[htbp]
\centering
\caption{ATP Wave 50: Performance by Political Party (POLPARTY).}
\label{table:polparty_w50}
\vspace{0.3em}
\begin{tabular}{lcccccccc}
\toprule
\textbf{Model} & \multicolumn{2}{c}{Democrat} & \multicolumn{2}{c}{Republican} & \multicolumn{2}{c}{Independent} & \multicolumn{2}{c}{Other} \\
 & Acc & WD & Acc & WD & Acc & WD & Acc & WD \\
\midrule
LoRA & 0.5516 & 0.4405 & 0.6380 & 0.5460 & 0.5686 & 0.4118 & 0.4950 & 0.3861 \\
OPPU & 0.5952 & 0.2460 & 0.6980 & 0.2078 & \textbf{0.5743} & \textbf{0.0990} & \textbf{0.6442} & 0.4356 \\
Cu & 0.5595 & 0.4286 & 0.5922 & 0.3725 & 0.5050 & 0.3168 & 0.6135 & 0.4969 \\
PuQu & 0.5556 & 0.4325 & 0.5922 & 0.3765 & 0.4950 & 0.3267 & 0.6135 & 0.4969 \\
Aplaud+ & \textbf{0.6349} & \textbf{0.1905} & \textbf{0.7137} & \textbf{0.0745} & 0.5644 & \textbf{0.0990} & \textbf{0.6442} & \textbf{0.2822} \\
\bottomrule
\end{tabular}
\end{table*}

\subsubsection{ATP Wave 54}
\textit{Topic: Economic Inequality}

\paragraph{Subgroup Performance in ATP Wave 54 by Region (CREGION).}
As shown in Table ~\ref{table:cregion_w54}, \textbf{Aplaud+} achieves the highest accuracy in the Northeast (0.6176). Additionally, the model OPPU attaining the lowest Wasserstein Distance (WD) in all regions: Midwest (0.1518), Northeast (0.0490), South (0.0889), and West (0.0810). 

\begin{table*}[htbp]
\centering
\caption{ATP Wave 54: Performance by Region (CREGION).}
\label{table:cregion_w54}
\vspace{0.3em}
\begin{tabular}{lcccccccc}
\toprule
\textbf{Model} & \multicolumn{2}{c}{Midwest} & \multicolumn{2}{c}{Northeast} & \multicolumn{2}{c}{South} & \multicolumn{2}{c}{West} \\
& Acc & WD & Acc & WD & Acc & WD & Acc & WD \\
\midrule
LoRA & 0.4167 & 0.3690 & 0.5539 & 0.3431 & 0.4578 & 0.3333 & \textbf{0.5048} & 0.2762 \\
OPPU & \textbf{0.4821} & \textbf{0.1518} & 0.5539 & \textbf{0.0490} & 0.4822 & \textbf{0.0889} & 0.4714 & \textbf{0.0810} \\
Cu & 0.4613 & 0.1994 & 0.6078 & 0.1618 & 0.4844 & 0.1667 & 0.4952 & 0.1333 \\
PuQu & 0.4613 & 0.1935 & 0.6029 & 0.1716 & \textbf{0.4867} & 0.1733 & 0.5000 & 0.1381 \\
Aplaud+ & 0.4583 & 0.1964 & \textbf{0.6176} & 0.1422 & 0.4778 & 0.1578 & 0.4952 & 0.1381 \\
\bottomrule
\end{tabular}
\end{table*}

\paragraph{Subgroup Performance in ATP Wave 54 by Gender (SEX).}
Table~\ref{table:sex_w54} presents the subgroup performance by gender for ATP Wave 54. \textbf{Aplaud+} achieves the highest accuracy for both Female (0.5000) and Male (0.4980) respondents. In terms of distributional alignment measured by Wasserstein Distance (WD), OPPU performs best, achieving the lowest WD values for both Female (0.1092) and Male (0.0714) groups. This indicates that while \textbf{Aplaud+} was effective in maximizing predictive accuracy, OPPU better captures the nuanced distributional patterns across gender groups.

\begin{table}[htbp]
\centering
\caption{ATP Wave 54: Performance by Gender (SEX).}
\label{table:sex_w54}
\vspace{0.3em}
\begin{tabular}{lcccc}
\toprule
\textbf{Model} & \multicolumn{2}{c}{Female} & \multicolumn{2}{c}{Male} \\
 & Acc & WD & Acc & WD \\
\midrule
LoRA & \textbf{0.5000} & 0.2974 & 0.4306 & 0.3869 \\
OPPU & 0.5086 & \textbf{0.1092} & 0.4702 & \textbf{0.0714} \\
Cu & 0.5043 & 0.1494 & 0.4960 & 0.1806 \\
PuQu & 0.5057 & 0.1566 & 0.4960 & 0.1905 \\
Aplaud+ & \textbf{0.5000} & 0.1408 & \textbf{0.4980} & 0.1766 \\
\bottomrule
\end{tabular}
\end{table}

\paragraph{Subgroup Performance in ATP Wave 54 by Political Party (POLPARTY).}
Table~\ref{table:polparty_w54} summarizes model performance across political party subgroups for ATP Wave 54. \textbf{Aplaud+} achieved the highest accuracy among Democrats (0.5317), while \textbf{PuQu} had the best accuracy for Republicans (0.5072) and \textbf{Cu} had the best accuracy for Independents (0.5072). \textbf{LoRA} performed best in terms of accuracy for the "Other" category (0.4833). Regarding Wasserstein Distance (WD), OPPU showed distributional alignment, yielding the lowest WD values for Democrats (0.0952), Independents (0.0560), and "Other" affiliations (0.2611). For Republicans, \textbf{Aplaud+} achieves the lowest WD (0.1307).

\begin{table*}[htbp]
\centering
\caption{ATP Wave 54: Performance by Political Party (POLPARTY).}
\label{table:polparty_w54}
\vspace{0.3em}
\begin{tabular}{lcccccccc}
\toprule
\textbf{Model} & \multicolumn{2}{c}{Democrat} & \multicolumn{2}{c}{Republican} & \multicolumn{2}{c}{Independent} & \multicolumn{2}{c}{Other} \\
 & Acc & WD & Acc & WD & Acc & WD & Acc & WD \\
\midrule
LoRA & 0.5119 & 0.3492 & 0.3750 & 0.4028 & 0.4626 & 0.3376 & \textbf{0.4833} & 0.3444 \\
OPPU & 0.5198 & \textbf{0.0952} & 0.4583 & 0.1389 & 0.4957 & \textbf{0.0560} & 0.4556 & \textbf{0.2611} \\
Cu & 0.5317 & 0.1865 & 0.3889 & 0.2778 & \textbf{0.5072} & 0.1394 & 0.4778 & 0.3056 \\
PuQu & 0.5317 & 0.2024 & \textbf{0.5072} & 0.1466 & 0.4833 & 0.2778 & 0.3889 & 0.2778 \\
Aplaud+ & \textbf{0.5317} & 0.1786 & 0.5057 & \textbf{0.1307} & 0.4778 & 0.3000 & 0.3750 & 0.2917 \\
\bottomrule
\end{tabular}
\end{table*}
\subsubsection{GSS Panel (2016–2020)}
\textit{Topic: General Social Trends}\\
We applied the same stratified evaluation to the General Social Survey panel dataset, using Wave 1a variables (2016). The table below summarizes model performance by subgroup. Unlike ATP, GSS includes different varialbes for region.

\paragraph{Subgroup Performance in GSS by Region (CREGION).}
As shown in Table~\ref{table:gss_cregion}, the \textbf{Aplaud+} model (APlaud) achieved the highest accuracy and lowest Wasserstein Distance in both SoNew England and Pacific regions, indicating particularly strong performance in these areas. In contrast, simpler models such as LoRA and OPPU perform best in different regions, with LoRA attaining its highest accuracy in the Middle Atlantic (Acc = 0.4171) and OPPU in East North Central (Acc = 0.3596), while OPPU’s lowest WD is observed in the Middle Atlantic (0.2362). 
\begin{table*}[htbp]
    \centering
    \caption{GSS: Performance by Sub-Region (CREGION)}
    \label{table:gss_cregion}
    \renewcommand\arraystretch{1.15}
    \setlength{\tabcolsep}{5pt}
    \footnotesize
    \begin{tabularx}{0.95\textwidth}{l *{4}{>{\centering\arraybackslash}X >{\centering\arraybackslash}X}}
        \toprule
        \multirow{2}{*}{Model} 
        & \multicolumn{2}{c}{East North Central} 
        & \multicolumn{2}{c}{Middle Atlantic} 
        & \multicolumn{2}{c}{SoNew England} 
        & \multicolumn{2}{c}{Pacific} \\
        \cmidrule(lr){2-3} \cmidrule(lr){4-5} \cmidrule(lr){6-7} \cmidrule(lr){8-9}
        & Acc & WD & Acc & WD & Acc & WD & Acc & WD \\
        \midrule
        LoRA
        & 0.3483 & \textbf{0.1910}
        & \textbf{0.4171} & 0.3266
        & 0.4340 & 0.3113
        & 0.4545 & 0.3455 \\
        OPPU
        & \textbf{0.3596} & \textbf{0.1910}
        & 0.4070 & \textbf{0.2362}
        & 0.5094 & 0.2453
        & 0.4364 & 0.2424 \\
        Cu
        & 0.3371 & 0.2584
        & 0.3970 & 0.2714
        & 0.5283 & 0.2264
        & 0.4788 & 0.2364 \\
        PuQu
        & 0.3371 & 0.2584
        & 0.3970 & 0.2714
        & 0.5377 & 0.2170
        & 0.4788 & 0.2364 \\
        Aplaud+
        & 0.3146 & 0.2360
        & 0.3970 & \textbf{0.2362}
        & \textbf{0.5566} & \textbf{0.1887}
        & \textbf{0.5091} & \textbf{0.2061} \\
        \bottomrule
    \end{tabularx}
\end{table*}

\paragraph{Subgroup Performance in GSS by Gender (SEX).}
Table~\ref{table:gss_sex} presents performance by gender for the GSS data. \textbf{PuQu} achieved the highest accuracy for females (0.4936), while \textbf{Cu} attained the highest accuracy for males (0.4873). In terms of Wasserstein Distance (WD), \textbf{PuQu} yields the lowest WD for females (0.1911), and OPPU achieved the lowest WD for males (0.1656). 
\begin{table}[htbp]
\centering
\caption{GSS: Performance by Gender (SEX).}
\label{table:gss_sex}
\vspace{0.3em}
\begin{tabular}{lcccc}
\toprule
\textbf{Model} & \multicolumn{2}{c}{Female} & \multicolumn{2}{c}{Male} \\
 & Acc & WD & Acc & WD \\
\midrule
LoRA        & 0.4061 & 0.3198 & 0.4682 & 0.2580 \\
OPPU        & 0.4162 & 0.2234 & 0.4522 & \textbf{0.1656} \\
Cu          & 0.4289 & 0.2487 & \textbf{0.4873} & 0.2006 \\
PuQu      & \textbf{0.4936} & \textbf{0.1911} & 0.4676 & 0.1877 \\
Aplaud+ & 0.4442 & 0.2081 & 0.4809 & 0.1847 \\
\bottomrule
\end{tabular}
\end{table}

\paragraph{Subgroup Performance in GSS by Political Party (POLPARTY).}
Table~\ref{table:gss_polparty} presents model performance by political party affiliation. For Democrats, our APlaud approach (\textbf{Aplaud+}) achieves the highest accuracy (0.4778). Among Independents and respondents identifying as Other, several models—including OPPU, Cu, PuQu, and Aplaud+—reached the maximum possible accuracy (0.5000). In terms of Wasserstein Distance (WD), both APlaud and OPPU achieve the lowest value (0.1667) for the Other group, while APlaud yielded the lowest WD for Independents (0.2038).

\begin{table*}[htbp]
    \centering
    \caption{GSS: Performance by Political Party (POLPARTY)}
    \label{table:gss_polparty}
    \renewcommand\arraystretch{1.15}
    \setlength{\tabcolsep}{7pt}
    \footnotesize
    \begin{tabularx}{0.95\textwidth}{l *{4}{>{\centering\arraybackslash}X >{\centering\arraybackslash}X}}
        \toprule
        \multirow{2}{*}{Model}
        & \multicolumn{2}{c}{Democrat}
        & \multicolumn{2}{c}{Republican}
        & \multicolumn{2}{c}{Independent}
        & \multicolumn{2}{c}{Other} \\
        \cmidrule(lr){2-3} \cmidrule(lr){4-5} \cmidrule(lr){6-7} \cmidrule(lr){8-9}
        & Acc & WD & Acc & WD & Acc & WD & Acc & WD \\
        \midrule
        LoRA        
            & 0.4608 & \textbf{0.1877}
            & 0.4088 & 0.4380
            & 0.4189 & 0.4038
            & \textbf{0.5000} & 0.1667 \\
        OPPU        
            & 0.4164 & 0.2287
            & \textbf{0.4891} & \textbf{0.2336}
            & 0.4189 & 0.2226
            & \textbf{0.5000} & \textbf{0.1667} \\
        Cu          
            & 0.4710 & 0.2014
            & 0.4526 & 0.3285
            & \textbf{0.4340} & 0.2792
            & \textbf{0.5000} & 0.5000 \\
        PuQu      
            & 0.4676 & \textbf{0.1877}
            & 0.4599 & 0.3358
            & \textbf{0.4340} & 0.2830
            & \textbf{0.5000} & 0.5000 \\
        Aplaud+ 
            & \textbf{0.4778} & 0.2048
            & 0.4672 & 0.2628
            & 0.4377 & \textbf{0.2038}
            & \textbf{0.5000} & \textbf{0.1667} \\
        \bottomrule
    \end{tabularx}
\end{table*}

\section{More Experiments Details}
In this section, we describe the experimental framework used to simulate responses to human surveys using large language models (LLM). Specifically, we present the prompt designs, user profile extraction strategies, and training procedures adopted in our method. Our goal is to enable LLMs to approximate individual-level human responses through structured personalization, achieved via APlaud, a parameter-efficient method that requires orders of magnitude fewer parameters per user.
 
\subsection{Prompts for User Profile}

To simulate natural-language user profiles for survey response modeling, we construct textual summaries that integrate demographic metadata and selected survey responses from ATP data. These profiles serve as personalized inputs for downstream machine learning tasks, such as response generation or classification. Each summary captures a user’s background, financial stressors, and attitudes toward government responsibility. This profile-based approach allows language models to produce outputs that are grounded in realistic user context, improving both personalization and interpretability.

\vspace{0.5em}
\begin{tcolorbox}[enhanced, colback=gray!5, colframe=black!70, boxrule=0.5pt,
  width=\textwidth, arc=6mm, sharp corners=south, breakable,
  title=Prompt Template for Simulated User Profile]

\textbf{You are a professional assistant tasked with summarizing a user's demographic characteristics and their economic attitudes, financial stressors, and beliefs about inequality and government responsibility based on W54 survey data. Your output should be a single, coherent paragraph suitable for input into a machine learning model.}

\vspace{0.5em}
\textbf{Instructions:}
\begin{itemize}
  \item Write in complete, natural English sentences.
  \item Begin by summarizing demographic information: age, sex, race, education, marital status, religion, religious attendance, political party, political ideology, income, and region.
  \item Then summarize the user’s reported financial well-being, including current household finances, experiences growing up, and ability to meet basic needs.
  \item Include financial worries such as debt, retirement savings, or healthcare expenses.
  \item Describe the user's access to financial resources and assets, such as savings accounts, investments, or loans.
  \item Capture beliefs about economic fairness, hard work, and the role of government in providing housing, healthcare, education, or other forms of support.
  \item Summarize attitudes toward economic inequality—its perceived causes, who is responsible for fixing it, and which policy proposals are seen as effective.
  \item Include how the user thinks current economic conditions impact various groups (e.g., middle class, wealthy, poor).
  \item If available, mention expected future economic conditions and views on powerful actors like corporations or wealthy individuals.
  \item Skip any questions answered with “Refused” or missing responses.
  \item Do not add interpretation or sentiment beyond what is explicitly stated.
\end{itemize}

\vspace{0.5em}
\textbf{User demographic metadata:} \texttt{{\{\{metadata\}\}}} \\
\textbf{Survey responses:} \texttt{{\{\{profile\_questions\}\}}}
\vspace{0.5em}
\textbf{Generate a concise, fluent paragraph summarizing the user:}

\end{tcolorbox}

\subsection{Prompts for User Profile and History Q\&A}
We extend the simulated user profiling approach by merging each generated profile with a subset of previously answered survey questions and responses. This combined context served as an input prompt to simulate responses to new, unseen test questions. The prompt includes three key components: (1) a natural-language user background summary generated from structured metadata and survey answers, (2) the new survey question to be predicted, and (3) a multiple-choice format with clear answer options. The prompt is explicitly designed to constrain the model’s output to a single valid choice (e.g., A, B, C), allowing for consistent evaluation and comparison across users and items. This approach allowed the language model to condition its predictions on both the inferred user profile and their past answer behavior, enhancing personalization and response coherence.
\begin{tcolorbox}[enhanced, colback=gray!3, colframe=black, width=\textwidth,
  arc=6mm, boxrule=0.4pt, sharp corners=south, breakable,
  title=Prompt Template: Simulated Survey Response with Profile + History]

\textbf{You are user \{\{user\_id\}\}, with the following background summary:}

\{\{user\_profile\_paragraph\}\}

\vspace{0.5em}

\textbf{Here is the question:}

\{\{test\_question\}\}

\vspace{0.5em}

\textbf{This is a single-answer multiple choice question. Here are the options:}

\{\{A. ..., B. ..., C. ..., etc.\}\}

\vspace{0.5em}

\textbf{Please select the most appropriate answer based on your background.}

Respond with only the corresponding uppercase letter (e.g., A, B, C), and format your answer exactly like this: \texttt{A} \\
Do not include any explanation, reasoning, or repeat the question.
\end{tcolorbox}

\subsection{\textcolor{black}{Data and Prompts for generating User Profile}}\label{appdix: profile generation}

{\color{black}

We generate each user profile using \textbf{ChatGPT-4}. The profile is constructed from two sources of information:

\begin{itemize}
    \item \textbf{Survey-provided demographic metadata.} These metadata fields come directly from the original survey and include:
    \begin{itemize}
        \item Region (CREGION): Northeast, Midwest, South, West
        \item Sex (SEX): Male, Female
        \item Age group (AGE): 18--29, 30--49, 50--64, 65+
        \item Education level (EDUCATION): Less than high school, High school graduate, Some college, Associate's degree, College graduate, Postgraduate
        \item Citizenship (CITIZEN): Yes, No
        \item Marital status (MARITAL): Married, Divorced, Separated, Widowed, Never married
        \item Religion (RELIG): Protestant, Catholic, Jewish, Muslim, Buddhist, Hindu, Atheist, Agnostic, Other, Nothing in particular
        \item Religious attendance (RELIGATTEND): More than once a week, Weekly, Monthly, Few times/year, Seldom, Never
        \item Political party (POLPARTY): Republican, Democrat, Independent, Other
        \item Political ideology (POLIDEOLOGY): Very conservative, Conservative, Moderate, Liberal, Very liberal
        \item Race/ethnicity (RACE): White, Black, Asian, Hispanic, Other
        \item Income (INCOME): $<30k$, $30-50k$, $50-75k$, $75-100k$, $>100k$
    \end{itemize}

    \item \textbf{10 survey questions most relevant to user characterization}. These are selected among the user's answered items and reflect personal attitudes, preferences, or values. To avoid information leakage, we \textbf{remove} these 10 profile-related questions prior to constructing the train/validation/test split.
\end{itemize}

The generated profile is a neutral paragraph rewriting the demographic metadata and the selected 10 questions. No additional information is inferred.

The exact prompt used to generate the profile is shown below.
}

\begin{tcolorbox}[enhanced, colback=gray!3, colframe=black, width=\textwidth,
  arc=6mm, boxrule=0.4pt, sharp corners=south, breakable,
  title=Prompt for Generating the User Profile]

\textbf{You are a professional assistant tasked with summarizing a user's demographic
information and survey response profile in a clean, coherent paragraph for
input into a machine learning model.}

\vspace{0.5em}

\textbf{Instructions:}

\begin{itemize}
    \item Use complete, natural English sentences.
    \item Start by summarizing demographic information (age, sex, race, education,
    marital status, religion, political ideology, income, device type, language).
    \item Then summarize the user’s self-reported life satisfaction.
    \item Then summarize their leadership values and views about business or political
    leadership, based only on answered questions.
    \item Then summarize their beliefs about gender and leadership, if any.
    \item Skip any survey questions where the user answered ``No answer.''
    \item Be neutral and descriptive, without adding interpretation.
\end{itemize}

\vspace{0.5em}

\textbf{User demographic metadata:}\\
\texttt{\{metadata\}}

\vspace{0.5em}

\textbf{Survey responses (10 most profile-relevant items):}\\
\texttt{\{profile\_text\}}

\vspace{0.5em}

\textbf{Generate a concise, fluent paragraph summarizing the user.}

\end{tcolorbox}

\section{Details on Human Studies Data: Pew ATP and General Society Survey}
% In your document:
\subsection{Pew Research ATP}
The American Trends Panel (ATP) is a nationally representative panel of U.S. adults conducted by the Pew Research Center. ATP is designed to study a wide variety of topics, including politics, religion, internet usage, and family life. We analyze sampled questions from four waves, selecting only \textit{ASK ALL} questions—that is, questions posed to all respondents regardless of subgroup membership or branching logic.
In the original ATP design, many questions include randomized Likert-scale options (e.g., positive-to-negative or vice versa). To align with this, we also randomize the presentation order of answer choices in our LLM prompts.
\subsubsection{ATP Wave 36}
Wave 36 (fielded June 19 – July 2, 2018) explores public attitudes toward gender representation in leadership roles. While a majority of Americans express support for having more women in top leadership positions, many remain skeptical that gender parity will be achieved. Views vary notably by political affiliation and gender, reflecting broader social divides.

\begin{tcolorbox}[colback=gray!5!white, colframe=black!80, title=Sample Questions from ATP Wave 36, fonttitle=\bfseries]
\small
\begin{enumerate}[leftmargin=1em, label=\textbf{Q\arabic*.}, itemsep=0.8em]
    \item In general, how important, if at all, is it to you for someone in a top executive business position to provide guidance or mentorship to young employees?\\
    \textit{Options:} (A) Essential \quad (B) Important, but not essential \quad (C) Not important \quad (D) Refused

    \item Do you think that men and women in leadership roles are...\\
    \textit{Options:} (A) Basically similar \quad (B) Basically different \quad (C) Refused

    \item Who generally has a better approach to leadership?\\
    \textit{Options:} (A) Women \quad (B) Men \quad (C) Neither \quad (D) Refused

    \item What is the ideal situation for the number of women in high political office?\\
    \textit{Options:} (A) More, but still fewer than men \quad (B) Equal \quad (C) More than men \quad (D) Refused

    \item What is the ideal number of women in top executive business positions?\\
    \textit{Options:} (A) More, but still fewer than men \quad (B) Equal \quad (C) More than men \quad (D) Refused

    \item As more women run for office...\\
    \textit{Options:} (A) Gender parity is inevitable \quad (B) Men will still dominate \quad (C) Refused

    \item As more women enter management...\\
    \textit{Options:} (A) Gender parity is inevitable \quad (B) Men will still dominate \quad (C) Refused

    \item How much would more women in leadership improve life for women?\\
    \textit{Options:} (A) A lot \quad (B) Some \quad (C) Not much \quad (D) Nothing \quad (E) Refused

    \item How much would more women in leadership improve life for men?\\
    \textit{Options:} (A) A lot \quad (B) Some \quad (C) Not much \quad (D) Nothing \quad (E) Refused

    \item How much would more women in leadership improve life for all Americans?\\
    \textit{Options:} (A) A lot \quad (B) Some \quad (C) Not much \quad (D) Nothing \quad (E) Refused
\end{enumerate}
\end{tcolorbox}
\subsubsection{ATP Wave 42}

Wave 42 of the American Trends Panel, conducted from January 7 to January 21, 2019, focuses on public attitudes toward scientists, trust in science, and perceptions of the scientific method. The survey explores how Americans view the role of scientists in public policy, their confidence in scientific experts, and whether science is seen as a force for societal good. Respondents were also asked about the objectivity and integrity of scientists, as well as how much trust they place in scientists from different institutional backgrounds (e.g., industry, government, academia). The data provide insight into partisan and demographic divisions in trust toward scientific information and decision-making processes.

\vspace{1em}

\begin{tcolorbox}[
  enhanced,
  colback=gray!5!white,
  colframe=black!80,
  title=Sample Questions from ATP Wave 42,
  fonttitle=\bfseries,
  arc=5mm,
  boxrule=0.7pt,
  top=3mm,
  bottom=3mm,
  breakable
]
\small
\begin{enumerate}[leftmargin=1em, label=\textbf{Q\arabic*.}, itemsep=0.8em]
    \item Compared with twenty years ago, do you think developments in science have made people’s lives...\\
    \textit{Options:} (A) Better \quad (B) Worse \quad (C) About the same

    \item Looking ahead to the next twenty years, do you think developments in science will make people’s lives...\\
    \textit{Options:} (A) Better \quad (B) Worse \quad (C) About the same

    \item Overall, would you say science has had a mostly positive effect on our society or a mostly negative effect on our society?\\
    \textit{Options:} (A) Mostly positive \quad (B) Mostly negative \quad (C) Equal positive and negative effects

    \item How much confidence, if any, do you have in scientists to act in the best interests of the public?\\
    \textit{Options:} (A) A great deal \quad (B) A fair amount \quad (C) Not too much \quad (D) No confidence at all

    \item Which of these statements comes closer to your own view?\\
    \textit{Options:} (A) Scientists should take an active role in public policy debates\\
    \phantom{\textit{Options:}} (B) Scientists should stay out of public policy debates

    \item Which of these statements comes closer to your own view?\\
    \textit{Options:} (A) Public opinion should guide scientific policy\\
    \phantom{\textit{Options:}} (B) Issues are too complex for public opinion to guide

    \item In general, would you say scientific experts are...\\
    \textit{Options:} (A) Usually better \quad (B) Usually worse \quad (C) Neither

    \item When you hear research is reviewed by an independent committee, does this make you...\\
    \textit{Options:} (A) Trust more \quad (B) Less \quad (C) No difference

    \item Which best describes what you think about the scientific method?\\
    \textit{Options:} (A) Accurate conclusions \quad (B) Can produce any desired conclusion

    \item Which of these statements comes closer to your view?\\
    \textit{Options:} (A) Judgments based solely on facts \quad (B) Judgments as biased as others’
\end{enumerate}
\end{tcolorbox}
\subsubsection{ATP Wave 50}

Wave 50 of the American Trends Panel was conducted from June 25 to July 8, 2019, with responses from 9,834 U.S. adults. This wave focused on family life, romantic relationships, parenting, cohabitation, marriage expectations, and household dynamics. The survey included split-form designs to compare attitudes toward men and women across different relationship and parenting roles. Questions also explored satisfaction with family life, financial situations, and perceived social support. Responses were collected online, with weighting applied to ensure national representativeness across demographics such as age, gender, race, education, political affiliation, and internet access.

\vspace{1em}

\begin{tcolorbox}[
  enhanced,
  colback=gray!5!white,
  colframe=black!80,
  title=Sample Questions from ATP Wave 50,
  fonttitle=\bfseries,
  arc=5mm,
  boxrule=0.7pt,
  top=3mm,
  bottom=3mm,
  breakable
]
\small
\begin{enumerate}[leftmargin=1em, label=\textbf{Q\arabic*.}, itemsep=0.8em]
    \item In general, how important is it for a \textbf{man} to have a job or career he enjoys in order to live a fulfilling life?\\
    \textit{Options:} (A) Essential \quad (B) Important, but not essential \quad (C) Not important

    \item In general, how important is it for a \textbf{woman} to have a job or career she enjoys in order to live a fulfilling life?\\
    \textit{Options:} (A) Essential \quad (B) Important, but not essential \quad (C) Not important

    \item What do you think is the ideal situation for \textbf{women with young children}?\\
    \textit{Options:} (A) Working full-time \quad (B) Working part-time \quad (C) Not working for pay

    \item What do you think is the ideal situation for \textbf{men with young children}?\\
    \textit{Options:} (A) Working full-time \quad (B) Working part-time \quad (C) Not working for pay

    \item Do you think \textbf{couples who live together before marriage} have a...\\
    \textit{Options:} (A) Better chance at a successful marriage \quad (B) Worse chance \quad (C) Doesn’t make much difference

    \item How much pressure, if any, do you feel from \textbf{society} to marry your partner?\\
    \textit{Options:} (A) A lot \quad (B) Some \quad (C) Not too much \quad (D) No pressure at all

    \item How do you feel about the way \textbf{household chores} are divided between you and your partner?\\
    \textit{Options:} (A) Very satisfied \quad (B) Somewhat satisfied \quad (C) Somewhat dissatisfied \quad (D) Very dissatisfied

    \item Have you ever \textbf{reduced your work hours} due to balancing parenting and career?\\
    \textit{Options:} (A) Yes \quad (B) No

    \item Do you think couples who are \textbf{not married but living together} can raise children as well as married couples?\\
    \textit{Options:} (A) Yes \quad (B) No

    \item Do you trust your partner to \textbf{handle money responsibly}?\\
    \textit{Options:} (A) A great deal \quad (B) A fair amount \quad (C) Not much \quad (D) Not at all
\end{enumerate}
\end{tcolorbox}

\subsubsection{ATP Wave 54}

Wave 54 of the American Trends Panel was conducted from September 16 to 29, 2019, with responses from 6,878 U.S. adults. This wave focused on attitudes toward gender roles, parenting, household responsibilities, and societal expectations. Respondents were sampled across five strata to improve representation of underrepresented groups. The survey was administered online, with weights applied to correct for demographic and behavioral differences. The margin of error for the weighted sample is ±1.59 percentage points.

\vspace{1em}

\begin{tcolorbox}[
  enhanced,
  colback=gray!5!white,
  colframe=black!80,
  title=Sample Questions from ATP Wave 54,
  fonttitle=\bfseries,
  arc=5mm,
  boxrule=0.7pt,
  top=3mm,
  bottom=3mm,
  breakable
]
\small
\begin{enumerate}[leftmargin=1em, label=\textbf{Q\arabic*.}, itemsep=0.8em]
    \item Would you say there is...\\
    \textit{Options:} (A) Too much economic inequality \quad (B) Too little economic inequality \quad (C) About the right amount

    \item Do you think the U.S. economic system...\\
    \textit{Options:} (A) Requires only minor changes \quad (B) Requires major changes \quad (C) Needs to be completely rebuilt

    \item How much responsibility should the federal government have in reducing economic inequality?\\
    \textit{Options:} (A) A lot \quad (B) Some \quad (C) Only a little \quad (D) None

    \item How much does the current tax system contribute to economic inequality?\\
    \textit{Options:} (A) A great deal \quad (B) A fair amount \quad (C) Not too much \quad (D) Not at all

    \item Do you think some people start out with more opportunities than others?\\
    \textit{Options:} (A) Contributes a great deal to inequality \quad (B) A fair amount \quad (C) Not too much \quad (D) Not at all

    \item How much would raising the federal minimum wage reduce economic inequality?\\
    \textit{Options:} (A) A great deal \quad (B) A fair amount \quad (C) Not too much \quad (D) Nothing at all

    \item How much would expanding Medicare to cover all Americans reduce economic inequality?\\
    \textit{Options:} (A) A great deal \quad (B) A fair amount \quad (C) Not too much \quad (D) Nothing at all

    \item Should the government invest in education and job training, or give direct financial assistance?\\
    \textit{Options:} (A) Invest in education and job training \quad (B) Give direct assistance

    \item Do you think filling out the U.S. census will...\\
    \textit{Options:} (A) Benefit you personally \quad (B) Harm you personally \quad (C) Neither benefit nor harm

    \item How important is it for the government to provide a high-quality K–12 education?\\
    \textit{Options:} (A) Yes, it's the government’s responsibility \quad (B) No, it's not

    \item Thinking about your household's financial situation, how much are you affected by job availability in your area?\\
    \textit{Options:} (A) A great deal \quad (B) A fair amount \quad (C) Not too much \quad (D) Not at all

    \item How often do you worry about the cost of health care?\\
    \textit{Options:} (A) Every day \quad (B) Almost every day \quad (C) Sometimes \quad (D) Rarely \quad (E) Never

    \item Have you received government assistance such as SNAP, Medicaid, or unemployment benefits in the past 12 months?\\
    \textit{Options:} (A) Yes \quad (B) No

    \item How much does your family’s financial situation affect your children’s ability to succeed in life?\\
    \textit{Options:} (A) A great deal \quad (B) A fair amount \quad (C) Not too much \quad (D) Not at all

    \item How do you rate current U.S. economic conditions?\\
    \textit{Options:} (A) Excellent \quad (B) Good \quad (C) Only fair \quad (D) Poor
\end{enumerate}
\end{tcolorbox}
\subsection{General Social Survey (GSS)}

The General Social Survey (GSS) 2016–2020 Panel is a longitudinal dataset that re-interviewed respondents from the 2016 and 2018 GSS cross-sectional samples to measure social and attitudinal change over time. Participants from these earlier waves were invited to complete a follow-up survey in 2020. The resulting three-wave panel study includes responses from 2016 (Wave 1a), 2018 (Wave 1b), and 2020 (Wave 2).

\vspace{1em}

\begin{tcolorbox}[
  enhanced,
  colback=gray!5!white,
  colframe=black!80,
  title=Sample Questions from the GSS Panel,
  fonttitle=\bfseries,
  arc=5mm,
  boxrule=0.7pt,
  top=3mm,
  bottom=3mm,
  breakable
]
\small
\begin{enumerate}[leftmargin=1em, label=\textbf{Q\arabic*.}, itemsep=0.8em]
    \item What do you think the chances are these days that a white person won't get a job or promotion while an equally or less qualified Black person gets one instead?\\
    \textit{Options:} 1 = Very likely; 2 = Somewhat likely; 3 = Not very likely

    \item In general, do you think the courts in this area deal too harshly or not harshly enough with criminals?\\
    \textit{Options:} 1 = Too harshly; 2 = Not harshly enough; 3 = About right

    \item Should divorce in this country be easier or more difficult to obtain than it is now?\\
    \textit{Options:} 1 = Easier; 2 = More difficult; 3 = Stay as is

    \item Do you feel that the demands of your family life interfere with your job?\\
    \textit{Options:} 1 = Always; 2 = Often; 3 = Sometimes; 4 = Hardly ever; 5 = Never

    \item Have you ever given up or would you give up good job opportunities for the benefit of your family life?\\
    \textit{Options:} 1 = Yes, and would again; 2 = Yes, but wouldn’t again; 3 = No, but would; 4 = No, and wouldn’t

    \item A working mother can establish just as warm and secure a relationship with her children as a mother who does not work.\\
    \textit{Options:} 1 = Strongly agree; 2 = Agree; 3 = Disagree; 4 = Strongly disagree

    \item It is much better for everyone involved if the man is the achiever outside the home and the woman takes care of the home and family.\\
    \textit{Options:} 1 = Strongly agree; 2 = Agree; 3 = Disagree; 4 = Strongly disagree

    \item Because of past discrimination, employers should make special efforts to hire and promote qualified women.\\
    \textit{Options:} 1 = Strongly agree; 2 = Agree; 3 = Neither; 4 = Disagree; 5 = Strongly disagree

    \item Do you favor or oppose preferential hiring and promotion of women?\\
    \textit{Options:} 1 = Strongly favor; 2 = Not strongly favor; 3 = Not strongly oppose; 4 = Strongly oppose

    \item Most men are better suited emotionally for politics than are most women.\\
    \textit{Options:} 1 = Agree; 2 = Disagree

    \item A preschool child is likely to suffer if his or her mother works.\\
    \textit{Options:} 1 = Strongly agree; 2 = Agree; 3 = Disagree; 4 = Strongly disagree

    \item Should the government promote equality between men and women?\\
    \textit{Options:} 1 = Definitely should; 2 = Probably should; 3 = Probably should not; 4 = Definitely should not

    \item Compared with American families in general, how would you rate your family income?\\
    \textit{Options:} 1 = Far below average; 2 = Below average; 3 = Average; 4 = Above average; 5 = Far above average

    \item Are you satisfied with your present financial situation?\\
    \textit{Options:} 1 = Pretty well satisfied; 2 = More or less satisfied; 3 = Not satisfied at all

    \item Do you think there is any area near here where you would be afraid to walk alone at night?\\
    \textit{Options:} 1 = Yes; 2 = No
\end{enumerate}
\end{tcolorbox}

\section{\textcolor{black}{Relation with Recommender System}}

\textcolor{black}{Personalization has long been a central theme in recommender systems, where models infer user preferences from historical interactions and estimate item relevance over a large catalog. Conceptually, survey QA prediction has a distant parallel to collaborative filtering. When responses are binarized (yes/no or agree/disagree), a survey can be represented as a $Respondents \times Items$ matrix where each entry reflects a respondent’s position.  For multi-level items such as Likert scales, each question–response option can be expanded into a set of binary indicators--for example, mapping a 5-point item into five item-specific binary variables -- yielding a uniform binary representation across all items. Alternatively, when responses reflect ordered categories, these items may be encoded using a single ordinal score (e.g., 1–5), which preserves the inherent ordering of the response levels. Together, these encoding strategies allow heterogeneous survey instruments to be transformed into a structured matrix format that is compatible with downstream modeling.
From this perspective, predicting a respondent’s answer to a new item resembles preference completion in recommender systems. }

\textcolor{black}{
Despite this superficial similarity, the underlying formulation is fundamentally different. Modern recommendation is typically framed as a learning-to-rank problem, whereas personalized survey response prediction does not involve ranking. A second key distinction concerns the observation regime: recommender systems operate under extreme sparsity, where each user interacts with only a tiny fraction of the item space, and models must infer preferences from partial interactions across many users. In contrast, survey datasets provide complete responses over a shared set of questions, and user preference is inferred directly from how individuals semantically interpret and answer natural-language survey items. As a result, our problem is closer to modeling user-specific semantic judgments than to reconstructing latent preference structures from sparse interactions.}

\textcolor{black}{Classical recommendation methods such as matrix factorization\cite{he2016fast}, NeuMF\cite{he2017neural}, or LightGCN\cite{he2020lightgcn} rely exclusively on latent collaborative filtering signals without any item semantics. These approaches treat items as no-smenantic indices and assume user–item interactions follow a modeled structure. Such assumptions do not hold in our setting: survey questions have explicit wording and domain meaning, and since every user observes the same question set, collaborative filtering cannot exploit sparsity or cross-user item co-occurrence patterns, nor can it model the semantic structure shared across questions. As a result, traditional recommendation techniques are not directly applicable to personalized survey response generation.}

\textcolor{black}{More recently, large language models have been incorporated into recommender systems.  Early studies ~\cite{peng2024ecellm, dai2023uncovering, wang2023rethinking}
explore LLM’s zero-shot/few-shot potential via in-context learning. The mismatch between LLMs' general-purpose training and the specific demands of recommendation tasks results in inadequate performance. To better align LLM with the recommendation domain, one research line would formulate recommendation as a sequential generation task and methods such as P5~\cite{geng2022recommendation}, M6-Rec~\cite{cui2022m6} and serialize user--item histories into natural-language prompts and train an LLM to generate the next item or a ranked list. Another research line would use LLMs as auxiliary modules to enrich representations which leverages LLMs to augment item/user embeddings or to support re-ranking such as GPT4Rec~\cite{li2023gpt4rec}, LaMAR~\cite{valizadeh2025language}, LLM4Rec~\cite{gao2025llm4rerank}. To the best of our knowledge, existing LLM-based recommenders generally adopt a one-size-fits-all design and compress the personalized information into input tokens either by hard or soft prompts. Thus, while conceptually related through the lens of personalization, our framework achieves personalization model-wise and introduces a parameter-efficient personalization not present in current recommender system literature.}

\textcolor{black}{Overall, our contribution is orthogonal to the development of recommendation systems.  
We introduce (i) an SVD-based \emph{shared subspace} $(U,V)$ tailored specifically for structured survey QA, and (ii) a \emph{per-user lightweight personalization layer} $(C_u, \alpha_u, \beta_u)$ learned within that subspace. This design directly captures individual answer preferences under a shared question structure.  
To the best of our knowledge, no existing recommendation method performs LLM-powered low-rank shared-subspace personalization learning, making our approach distinct in both problem setting and technical design.}

\section{Broder Impacts}
% \clearpage
This paper presents \textbf{APlaud}, a parameter-efficient framework for personalized large language models, with a primary application to individualized survey response prediction. The goal of this work is to advance the field of machine learning by enabling scalable, data-efficient personalization while explicitly addressing overfitting, deployment cost, and shared structure induced by common survey questions across users.

\paragraph{Positive Societal Impact.}
APlaud has the potential to reduce the cost, time, and logistical burden of traditional survey-based data collection in both public and private sectors. By enabling personalized synthetic respondents models calibrated to an individual’s historical responses, the proposed approach may help organizations prototype and validate new survey instruments, conduct preliminary analyses, and extend the useful lifetime of existing datasets without repeatedly re-contacting participants. This can help lower respondent fatigue, reduce incentive pressure, and improve accessibility to research insights for smaller organizations or under-resourced settings. More broadly, the framework contributes to the development of personalized LLMs and digital twin technologies, which have applications in recommendation systems, human-centered AI, and decision support.

\paragraph{Ethical Considerations and Risks.}
The use of synthetic respondents raises important ethical considerations, particularly around misrepresentation, over-reliance on generated data, and the potential erosion of distinctions between real and synthetic human input. If used improperly, personalized synthetic responses could be mistaken for genuine human feedback or be applied outside the scope for which they were calibrated. Additionally, as with all personalization methods, there is a risk that models may encode or amplify biases present in historical survey responses.

% \paragraph{Mitigations and Scope.}
% APlaud is designed to model individual response patterns \emph{conditional on existing survey data} and is not intended to replace real human respondents in high-stakes decision-making, policy formation, or sensitive social inference. We emphasize that personalized synthetic responses should be used as a complementary analytical tool rather than a substitute for direct human engagement. Moreover, the method operates within the constraints of parameter-efficient fine-tuning and shared representations, which naturally limit memorization and reduce the risk of exposing sensitive information. Responsible deployment should include transparency about synthetic data usage, appropriate human oversight, and adherence to existing ethical guidelines for survey research and data governance.

% Overall, we believe that APlaud represents a measured and responsible advance toward scalable personalization in machine learning. While it introduces new capabilities for modeling individual-level preferences, the framework is designed to support exploratory analysis and system development, rather than to replace human judgment or participation in real world data collection.

% \section{Example Appendix}
% \label{sec:appendix}

% This is an appendix.

\end{document}